\documentclass[10pt,twocolumn,letterpaper]{article}

\usepackage[pagenumbers]{cvpr} %

\usepackage{booktabs}
\usepackage{multirow}
\usepackage[utf8]{inputenc}
\usepackage{booktabs}
\usepackage{amsmath,amssymb}
\usepackage{placeins}

\definecolor{cvprblue}{rgb}{0.21,0.49,0.74}
\usepackage[pagebackref,breaklinks,colorlinks,allcolors=cvprblue]{hyperref}
\usepackage[table]{xcolor}
\usepackage{booktabs}
\def\confName{CVPR}
\def\confYear{2027}

\title{Evidence-Driven Failure Diagnosis and Recovery for Video Object Segmentation}

\author{Valay Bundele\qquad Susmit Agrawal\qquad Mehran Hosseinzadeh\qquad Hendrik P.\ A.\ Lensch\\
University of T\"ubingen, Germany\\
{\tt\small \{valay.bundele,susmit.agrawal,mehran.hosseinzadeh,hendrik.lensch\}@uni-tuebingen.de}\\
}

\begin{document}
\twocolumn[{%
  \renewcommand\twocolumn[1][]{#1}%
  \maketitle
  \begin{center}
    \centering
    \captionsetup{type=figure}
    \includegraphics[width=0.85\textwidth]{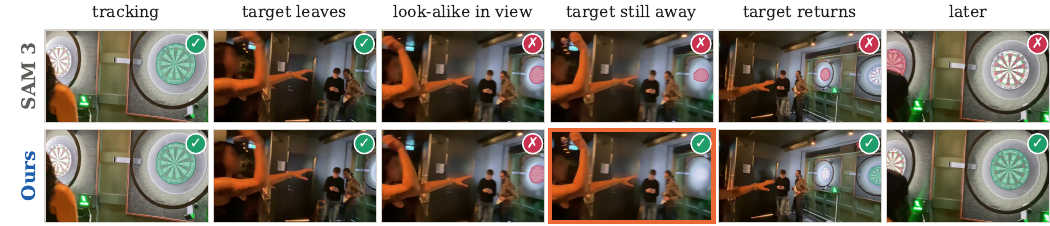}
    \captionof{figure}{When the target leaves, SAM~3 switches to a look-alike and fails to recover. Ours detects the error, erases the incorrect mask (orange), and re-identifies the returning target. Masks are green for IoU $>0.5$ with ground truth, red otherwise.}
    \label{fig:teaser}
  \end{center}%
  }]
\begin{abstract}
Despite advances in visual object tracking, occlusions, appearance changes, and distractors continue to cause persistent failures. Existing methods improve propagation robustness, while recent VLM-based approaches enable explicit verification and correction. However, reliable self-correction requires more than detecting errors and predicting corrections: it demands principled decisions about when to intervene, whether a correction is reliable, and when additional evidence is necessary. We address these challenges with an evidence-driven self-correcting framework comprising three components: (i) a training-free sequential failure detector that accumulates evidence from temporal dynamics,  (ii) a unified VLM that jointly assesses tracking correctness and localization, trained with identity-conditioned contrasts for recovery and restraint, and (iii) active zooming with confidence-aware commitment to resolve identity ambiguity and prevent unreliable memory updates. Our framework improves SAM~3 on all four VOS benchmarks while running 3.3$\times$ faster than Dual-Agent, detects 86\% of failures on SA-V with nearly half of Dual-Agent's inspections, and transfers to SAM~2.1 and SeC without adaptation, gaining up to 15.4 $J\&F$. Project page: \href{https://valaybundele.github.io/VIGIL}{valaybundele.github.io/VIGIL}.
\end{abstract}

\section{Introduction}
\label{sec:intro}
Video object segmentation (VOS) is fundamental to video understanding, enabling applications from video editing and interactive segmentation to robotics and autonomous systems. Recent advances in memory-based tracking, from XMem~\cite{cheng2022xmem} and Cutie~\cite{cheng2024putting} to foundation models such as SAM~2~\cite{ravi2025sam} and SAM~3~\cite{carion2026sam}, have substantially improved object propagation across videos. Nevertheless, occlusions, appearance changes, and visually similar distractors continue to cause target loss and identity switches. In memory-based trackers, such errors can contaminate memory and reinforce subsequent failures. Reliable tracking therefore requires not only accurate propagation, but also the ability to recognize and recover from its own failures.

Recent efforts to improve tracking robustness include alternative memory hypotheses, motion-aware memory selection, and distractor-aware representations~\cite{ding2025sam2long,yang2024samurai,videnovic2025distractor}. However, these methods strengthen propagation rather than explicitly assessing its correctness. A complementary direction uses vision-language models (VLMs) for visual reasoning~\cite{wu2026vos, bundele2026trackers, cai2026models}. In particular, Dual-Agent~\cite{bundele2026trackers} showed that VLM-based verification and correction can recover from failures that conventional propagation struggles to resolve.

Yet, reliable self-correction requires more than detecting errors and predicting corrections. Recovery operates in a closed loop, where missed failures allow erroneous propagation to persist, while unnecessary interventions can corrupt tracker memory and compromise future predictions. Moreover, tracking confidence may not reflect correctness, and recovery can remain ambiguous for small targets or visually similar distractors. Dual-Agent relies on heuristic verification schedules and predefined visual observations, limiting its ability to exploit evolving tracking evidence and resolve ambiguity before intervention. This raises a central question: \emph{how can a tracker recognize failures, acquire sufficient evidence for recovery, and determine whether a correction is trustworthy?}

To this end, we introduce VIGIL, an evidence-driven self-correcting framework that enables a tracker to recognize and reliably recover from its failures. Our key observation is that, although individual tracking signals may not reliably reflect correctness, they provide complementary evidence of failures.
We exploit this through a training-free \emph{sequential failure detector} that integrates evidence from propagation dynamics and memory consistency over time, distinguishing persistent failures from transient uncertainty.

Given a suspected failure, we introduce a unified VLM corrector that \emph{jointly reasons} about tracking correctness, target visibility, and localization, \emph{actively acquires} additional visual evidence, and \emph{selectively commits} corrections to tracker memory. This formulation shifts recovery from predicting a plausible correction to establishing whether a correction is warranted and sufficiently reliable.

We develop two complementary training strategies to support these decisions. First, to ensure that recovery is guided by the intended target rather than visual shortcuts, we introduce \emph{identity-conditioned contrasts}: identical query frames and tracker predictions receive different supervision depending solely on the target exemplar. Second, recognizing that the initial observations may lack sufficient detail to resolve identity ambiguity, we formulate recovery as a \emph{multi-turn reasoning} process. Through supervised dialogues and reward-based refinement, the corrector learns to selectively acquire high-resolution crops to resolve ambiguity before committing to a recovery decision.

Crucially, improving the ability to recover does not necessarily improve the ability to make a reliable correction. We therefore decouple these learning objectives: reinforcement learning refines evidence acquisition and localization, while correctness judgments remain directly supervised to preserve the reliability of verdict confidence. This enables \emph{confidence-aware commitment}, where the corrector's own verdict confidence governs memory updates. Together, these design choices enable recovery that is explicitly conditioned on target identity, actively seeks evidence, and selectively intervenes to avoid new failures.

Experiments across four VOS benchmarks show improvements in both failure detection and closed-loop recovery. On SA-V, our detector identifies 86\% of persistent failures with nearly half the inspections of Dual-Agent, transfers across trackers without retuning, and remains effective after corrective interventions. VIGIL consistently outperforms SAM 3 and Dual-Agent, with particularly strong gains following scene changes, while running 3.3× faster than Dual-Agent. Recovery transfers without adaptation to SeC and SAM 2.1, improving J\&F by up to 15.4 points.

\section{Related Work}
\label{sec:related}

\paragraph{Memory-based Video Object Segmentation.} Memory-based trackers such as XMem~\cite{cheng2022xmem} and Cutie~\cite{cheng2024putting}, and foundation models such as SAM~2~\cite{ravi2025sam} and SAM~3~\cite{carion2026sam}, propagate object identities using temporal memory. Recent methods improve robustness through motion-guided memory selection~\cite{yang2024samurai}, distractor-aware memory~\cite{videnovic2025distractor}, alternative memory hypotheses~\cite{ding2025sam2long}, and semantic target representations~\cite{zhang2025advancing}. However, these approaches primarily strengthen propagation rather than explicitly diagnosing and correcting failures. Our framework complements them by combining tracker-internal failure detection with VLM-based reasoning for selective recovery.
\paragraph{VLMs in the tracking loop} VLMs have been used for target identification in reasoning VOS~\cite{yan2024visa,bai2024one} and agentic referring-VOS~\cite{jiang2026refer,jin2026agentrvos}, while leaving propagation to conventional trackers. Recently, VOS-Agent~\cite{wu2026vos} and ACTrack~\cite{cai2026models} re-prompt SAM 3 when its prediction conflicts with a second tracker, with a VLM deciding which to trust. Dual-Agent~\cite{bundele2026trackers} introduced VLM-based recovery through heuristic triggers and separate verification and correction models, where errors can compound and recovery is constrained by fixed evidence. We instead formulate recovery as \emph{evidence-driven, selective self-correction}, where a sequential detector accumulates tracker-internal evidence, while a unified VLM actively acquires visual evidence and jointly reasons about tracking correctness and localization.

\paragraph{Active perception and zoom for VLMs.}
VLMs process images at a fixed, often reduced resolution, which can remove fine details. Training-free zooming~\cite{shen2025zoomeye} and crop-and-answer models trained with SFT and RL~\cite{shao2024visual,zheng2026deepeyes,zhang2025chain,lai2026mini,su2026pixel} let VLMs request magnified regions to answer questions about image content. In object tracking, VOS-Agent~\cite{wu2026vos} uses a 397B-parameter VLM only as an arbiter: when SAM~3 disagrees with an auxiliary tracker, the VLM picks one of the two boxes from system-built crops and never decides where to look. We bring active perception into the tracking loop. A single 4B VLM judges the track itself, zooms where it needs detail and localizes the target from what it sees. It needs no auxiliary tracker, and it corrects only when the evidence is clear.
\section{Methodology}
\label{sec:method}

\paragraph{Problem formulation.} Given a video $I_{1:T}$ and target mask $M_{t_0}^{\star}$, our objective is to segment the same target in subsequent frames. At each frame $t$, a tracker with fixed parameters $\phi$ predicts a mask and produces a provisional state, $(\widehat M_t,\widetilde S_t)
=\mathcal{T}_{\phi}(I_t,S_{t-1})$. Here, $S_{t-1}$ is the tracking state from the preceding frame, while $\widetilde S_t$ is the state resulting from ordinary propagation, before any intervention.

\begin{figure*}[t]
  \centering
  \includegraphics[width=0.9\textwidth]{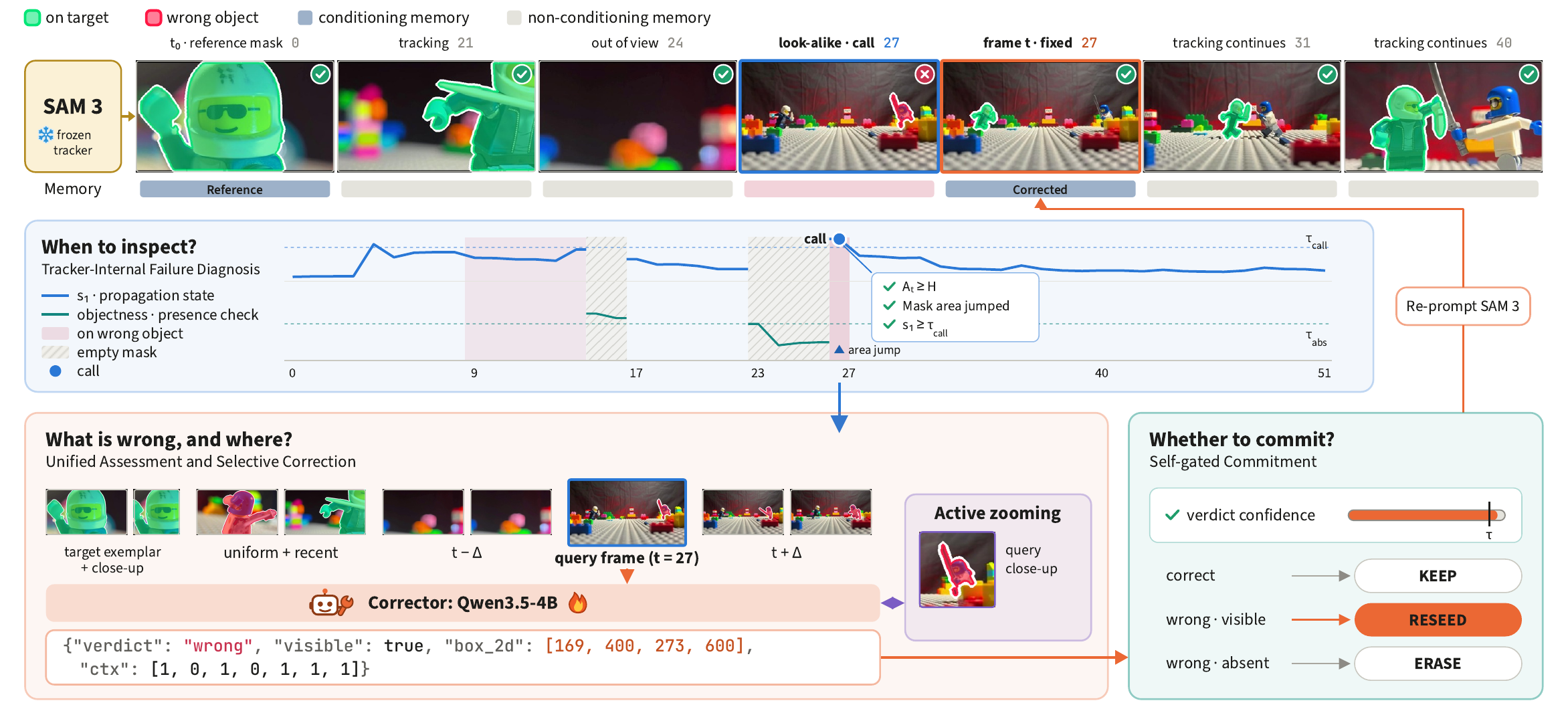}
  \caption{Overview of our evidence-driven self-correcting tracking framework. SAM 3 loses the target and switches to a look-alike. Tracker-internal failure diagnosis accumulates temporal evidence to trigger inspection. A unified VLM assesses tracking correctness, visibility, and localization, using active zooming to resolve identity ambiguity. Confidence-aware commitment accepts the proposed correction and re-prompts the frozen tracker, restoring the target identity for subsequent propagation.}
  \label{fig:pipeline}
\end{figure*}
Our framework determines \emph{when} to inspect this prediction and \emph{whether} to modify tracking state. Let $q_t\in\{0,1\}$ denote an inspection request. Upon inspection, the corrector proposes an action $a_t$: \textsc{keep} retains the prediction, $\textsc{reseed}(b_t)$ supplies a corrected target box, and \textsc{erase} supplies an empty-mask prompt. A commitment gate $g_t\in\{0,1\}$ accepts corrective actions only when there is sufficient confidence, and is zero for \textsc{keep} or rejected proposals.
The final prediction and state are
\begin{equation}
(\overline M_t,S_t)=
\begin{cases}
\mathcal{U}_{\phi,t}(\widetilde S_t,a_t),
    & q_tg_t=1,\\
(\widehat M_t,\widetilde S_t),
    & \text{otherwise}.
\end{cases}
\label{eq:state_update}
\end{equation}
Operator $\mathcal{U}_{\phi,t}$ re-runs the tracker on frame $t$ with the corrective prompt while retaining earlier memory. 
\paragraph{Preliminaries:} SAM~3~\cite{carion2026sam} combines an image-level detector with a memory-based video tracker, which propagates object masks using stored representations of the target. These comprise \emph{conditioning memory} $\mathcal C_t$, derived from explicitly prompted frames, and \emph{non-conditioning memory} $\mathcal B_t$, accumulated from previous predictions. At each frame, the image encoder extracts spatial features, which are conditioned on these memory representations through memory attention.
The resulting features are processed by a mask decoder that predicts three candidate masks $\{\widehat M_t^{(a)}\}_{a=1}^{3}$, associated mask-quality estimates, and an object-presence logit $o_t$. The selected prediction $\widehat M_t$ has effective confidence $u_t^{\mathrm{eff}}$, with presence probability $p_t^{\mathrm{obj}}=\sigma(o_t)$. The memory encoder then combines the predicted mask with image features to produce a memory representation $\mu_t$ for subsequent propagation. We use these internal signals and memory representations for failure detection.

\subsection {VIGIL}
Our framework enables a frozen tracker to detect and recover from propagation failures by separating \emph{when} the tracking state warrants inspection from \emph{whether} a proposed correction should modify it (\cref{fig:pipeline}). 
We address these decisions through three components. First, a training-free \emph{sequential failure detector}, accumulates temporal evidence to identify potential failures. Second, a \emph{unified VLM corrector} jointly assesses tracking correctness, target visibility, and localization to determine the appropriate recovery action. Third, \emph{active visual zooming} enables the corrector to acquire high-resolution evidence and \emph{confidence-aware commitment} selectively applies reliable corrections, preventing erroneous updates from compromising propagation.

\subsubsection{Tracker-Internal Failure Diagnosis}
\label{sec:trigger}

Memory-based trackers propagate object identities through an evolving memory, where erroneous predictions can reinforce identity switches over later frames. Such failures are difficult to detect from confidence alone: incorrect tracks may remain confident, while correct tracks may become uncertain. The challenge is therefore to recognize when propagation has lost correspondence with the initial target.

We address this through a training-free sequential diagnosis mechanism that examines two complementary aspects of propagation: \emph{how the tracking state evolves} and \emph{whether that state remains consistent with target specification}. The resulting measurements are accumulated within candidate failure episodes, allowing inspection to be triggered by sustained evidence of unreliable propagation.
\paragraph{Propagation-State Evidence.} We begin by characterizing the state maintained during propagation. The tracker's effective confidence \(u_t^{\mathrm{eff}}\) and object-presence probability \(p_t^{\mathrm{obj}}\) provide estimates of prediction quality and target presence. However, these quantities can remain high after an identity switch. We therefore also monitor changes in representation of the tracked object relative to a fixed reference.

Let \(\mathbf f_t\) denote spatial memory features pooled over the current predicted mask \(\widehat M_t\). Using the exemplar pooled representation as a fixed reference \(\mathbf f_{\mathrm{ref}}\), we compute $c_t=\cos(\mathbf f_t,\mathbf f_{\mathrm{ref}})$.
This provides a stable reference for detecting representation drift that may not be reflected in confidence. However, appearance changes can also reduce similarity, so $c_t$ is treated as diagnostic evidence rather than a direct measure of identity correctness.

Since propagation also depends on memory accumulated from earlier predictions, we use mean confidence of the non-conditioning entries:
$m_t=\frac{1}{|\mathcal B_t|}
\sum_{j\in\mathcal B_t}u_j^{\mathrm{eff}}$,
where $u_j^{\mathrm{eff}}$ is the confidence recorded when entry $j$ was generated. Together, $\mathcal V_{1,t}=\{u_t^{\mathrm{eff}},p_t^{\mathrm{obj}},c_t,m_t\}$ characterizes prediction reliability, representation change, and memory quality. 

\paragraph{Target-consistency evidence.}
The second evidence group,
$\mathcal V_{2,t}=\{r_t^{\mathrm{cond}},r_t^{\mathrm{back}},r_t^{\mathrm{hyp}}\}$,
assesses whether the current prediction is consistent with target identity and whether its localization is ambiguous. Let $D_\phi(I;\mathcal M)$ denote mask prediction on image $I$ using memory entries $\mathcal M$. We evaluate consistency in two complementary directions: first, comparing current prediction $\widehat M_t$ with one obtained using only conditioning memory $\mathcal C_t$, and second, by measuring how well a prediction on the initial frame using the current memory feature $\mu_t$ matches the ground-truth mask $M_{t_0}^{\star}$:
\begin{equation}
\begin{aligned}
r_t^{\mathrm{cond}}
&=\operatorname{IoU}\!\left(
D_\phi(I_t;\mathcal C_t),\widehat M_t
\right),\\
r_t^{\mathrm{back}}
&=\operatorname{IoU}\!\left(
D_\phi(I_{t_0};\{\mu_t\}),M_{t_0}^{\star}
\right).
\end{aligned}
\label{eq:memory_checks}
\end{equation}
Low $r_t^{\mathrm{cond}}$ indicates disagreement with the conditioning memory, whereas low $r_t^{\mathrm{back}}$ suggests that the current representation is inconsistent with the original target.

Memory consistency alone may not reveal ambiguous localization. We therefore measure agreement among SAM~3's three candidate masks $\{\widehat M_t^{(a)}\}_{a=1}^{3}$:
\begin{equation}
r_t^{\mathrm{hyp}}=
\frac{1}{3}\sum_{1\le a<b\le3}
\operatorname{IoU}\!\left(
\widehat M_t^{(a)},\widehat M_t^{(b)}
\right).
\label{eq:hypothesis_consistency}
\end{equation}
Low agreement indicates competing localization hypotheses. Together, these signals expose target inconsistency and ambiguity that may not be reflected in SAM 3's confidence.

\paragraph{Sequential Accumulation and Inspection.} We now use these observations to identify the temporal development of potential failures. A single discrepancy does not establish that propagation has lost the target: representation changes, uncertain masks, and disagreement between memory-conditioned predictions can also occur during successful tracking. We therefore follow their evolution, summarizing each group as a discrepancy score:
\begin{equation}
s_{k,t}=\frac{1}{|\mathcal V_{k,t}|}
\sum_{v\in\mathcal V_{k,t}}(1-v),
k\in\{1,2\}.
\label{eq:evidence_scores}
\end{equation}
Larger scores indicate stronger deviations, rather than calibrated failure probabilities. When \(s_{1,t}\ge\theta_{\mathrm{open}}\), we initiate a candidate failure episode and accumulate evidence:%
\begin{equation}
A_t=A_{t^-}+\delta_t
\left[\max(s_{1,t},s_{2,t})-\kappa\right]_+,
\label{eq:accumulator}
\end{equation}
where $[x]_+=\max(x,0)$, $t^-$ denotes preceding non-empty observation, and $\delta_t=\max(1,t-t^-)$ accounts for gaps between processed frames. $A_t$ starts from current contribution at episode onset. The offset $\kappa$ suppresses weak deviations. %

As evidence accumulates, we also monitor changes in the predicted segmentation. A substantial change in segmentation mask area in an episode can indicate target loss, incorrect localization, or a switch to a distractor:
\begin{equation}
J_t=J_{t^-}\lor
\mathbf{1}\!\left[
\left|\log|\widehat M_t|-\log|\widehat M_{t^-}|\right|
\ge\eta
\right],
\label{eq:area_jump}
\end{equation}

where $J_t=0$ at onset and $\eta$ controls the minimum relative area change. 
An inspection is requested when the accumulated evidence reaches $H$, an area change has occurred, and the current propagation-state discrepancy remains elevated:
 \begin{equation}
q_t^{\mathrm{mask}}=
\mathbf{1}[A_t\ge H]\,
\mathbf{1}[J_t=1]\,
\mathbf{1}[s_{1,t}\ge\tau_{\mathrm{call}}].
\label{eq:nonempty_trigger}
\end{equation}

\paragraph{Presence Monitoring.} Empty predictions require separate handling because mask-dependent diagnostic measurements are unavailable. Moreover, an empty mask may indicate either genuine target absence or a failure to localize a visible target.
A periodic presence check uses the tracker's objectness estimate to identify potential missed detections.
Let $n_t$ count consecutive empty predictions, resetting on a non-empty output. Every $P$ empty frames, we request inspection when the objectness logit exceeds $\tau_{\mathrm{abs}}$:
\begin{equation}
q_t^{\mathrm{abs}}=
\mathbf{1}[n_t>0]\,
\mathbf{1}[n_t\bmod P=0]\,
\mathbf{1}[o_t\ge\tau_{\mathrm{abs}}].
\label{eq:absence_trigger}
\end{equation}
The final inspection request is $q_t=q_t^{\mathrm{mask}}\lor q_t^{\mathrm{abs}}$. Both channels reset after an applied correction, ensuring that subsequent evidence is accumulated relative to the revised state.

\subsubsection{Unified Assessment and Selective Correction}
\label{sec:corrector}

The detector identifies potentially unreliable propagation but cannot establish \emph{whether a prediction is incorrect} or \emph{how to recover the target}. We address this with a unified VLM corrector that jointly assesses tracking correctness, visibility, and localization, acquires high-resolution crops when needed, and commits only confident corrections.

\paragraph{Target-conditioned temporal assessment.}
Reliable recovery requires establishing the intended target identity and assessing how tracking has evolved over time. We construct an evidence set $\mathcal{E}_t^{(0)}$ containing a target exemplar and close-up, the query frame, and temporal context at three scales: (i) $U$ uniformly sampled historical frames for long-term identity, (ii) $R$ recent frames sampled at fixed intervals for tracking evolution, and (iii) densely sampled frames within $[t-\Delta,t+\Delta]$ for local propagation dynamics. All frames are overlaid with predicted contours. Future frames are obtained by propagating a separate copy of the tracker state.

Conditioned on $\mathcal{E}_t^{(0)}$, the corrector jointly predicts
$y_t=(d_t,v_t,b_t,h_t)$, where $d_t\in\{\textsc{correct},\textsc{wrong}\}$ assesses whether the current prediction identifies the intended target, $v_t\in\{0,1\}$ indicates target visibility, and $b_t$ specifies its bounding box in normalized coordinates. Auxiliary judgments $h_t$ assess correctness across context frames, supervising temporal error reasoning. The resulting predictions define three recovery actions:
\begin{equation}
a_t=
\begin{cases}
\textsc{keep}, & d_t=\textsc{correct},\\
\textsc{reseed}(b_t), & d_t=\textsc{wrong},\ v_t=1,\\
\textsc{erase}, & d_t=\textsc{wrong},\ v_t=0.
\end{cases}
\label{eq:action_mapping}
\end{equation}
Thus, the corrector jointly determines whether intervention is necessary and how to recover the target.

\paragraph{Active zooming.}
\label{sec:zoom}
Fixed-resolution observations may be insufficient to distinguish small targets or visually similar distractors. We therefore use a bounded multi-turn zoom dialogue: the corrector selects a query-frame crop, examines the returned detail, and either produces its assessment or requests one further crop. The target reference and temporal context remain available to establish identity.
The close-up $C_t^{(k)}$ extracts a rectangle $r_t^{(k)}\in[0,1000]^4$ from the source image and resizes it keeping the aspect ratio. We append the view and its full-frame coordinates to the dialogue:
$\mathcal E_t^{(k)}
=\mathcal E_t^{(k-1)}\oplus(r_t^{(k)},C_t^{(k)})$.
Here $\oplus$ denotes dialogue concatenation. Magnification is capped at $6\times$. Explicit crop coordinates keep the correspondence needed to predict the bounding box $b_t$ in the full query frame.

\paragraph{Self-gated commitment.}
As incorrect interventions can corrupt tracker memory and compromise subsequent propagation, self-gated commitment accepts corrections only when sufficiently confident. %
We estimate confidence in the \textsc{wrong} verdict using the mean conditional log-probability of its generated tokens at positions $\mathcal K_t$:
\begin{equation}
\ell_t=
\frac{1}{|\mathcal K_t|}
\sum_{j\in\mathcal K_t}
\log p_\theta
\left(y_{t,j}\mid
\mathcal E_t^{\mathrm{final}},y_{t,<j}\right).
\label{eq:verdict_confidence}
\end{equation}
A correction is eligible for commitment only if $\ell_t\ge\log\tau$. Accepted corrections re-run the frozen tracker on the current frame with the corrective prompt, otherwise, the original prediction and state are preserved.%

\paragraph{Inference.}
During propagation, the failure detector accumulates temporal evidence and triggers inspection when a potential failure is identified. The unified corrector assesses the tracking state, selectively acquires high-resolution crops, and proposes a recovery action, which is committed only when sufficiently confident.

\begin{figure}[t]
    \centering
    \includegraphics[width=\linewidth]{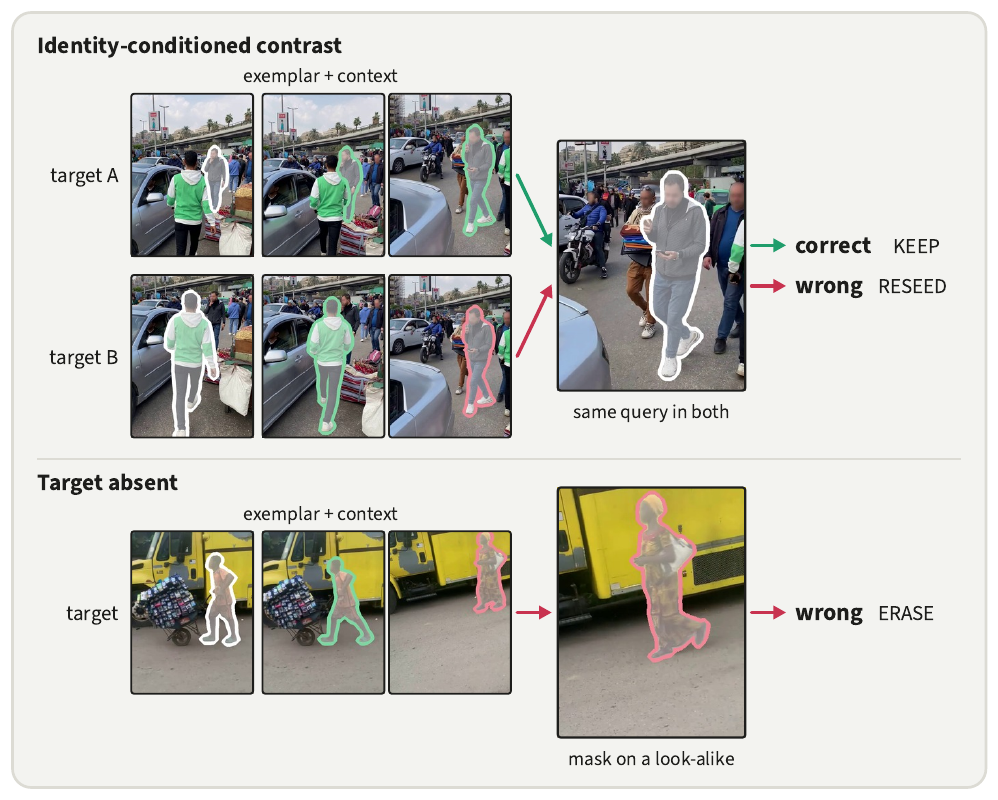}
    \caption{\emph{Top:} an identity-conditioned contrast. The same query and
    prediction are correct for target~A (\textsc{keep}) but wrong for
    target~B (\textsc{reseed}). \emph{Bottom:} the target is absent and the
    tracker's mask lies on a look-alike, so the correct action is
    \textsc{erase}.}
    \label{fig:training_data}
  \end{figure}

\subsubsection{Training the Unified Corrector}
\label{sec:method_training}

We fine-tune Qwen3.5-4B~\cite{qwen3.5} to jointly assess tracking correctness, target visibility, and localization, while actively acquiring additional visual evidence when needed. Training proceeds in three stages: identity-aware supervision, multi-turn zoom cold-start, and reward-based refinement.

\paragraph{Identity-aware recovery and restraint.}
We construct a challenging dataset of 90,379 examples from SAM 3 trajectories, %
providing structured supervision for correctness verdict \(d_t\), target visibility \(v_t\), bounding box \(b_t\), and historical correctness judgments \(h_t\). The dataset covers five tracking states of SAM3: correct visible tracks, incorrect masks, missed visible targets, spurious predictions during target absence, and correct absence. This trains the corrector to recover from failures while preserving valid predictions.

To strengthen identity discrimination, around 10\% of the examples contain identity-conditioned contrasts built from look-alike object pairs $(T,L)$ with JEV similarity above 0.5. Each pair yields examples with identical query images and tracker predictions but different target references (\cref{fig:training_data}, top): a prediction covering $L$ is correct when tracking $L$ but incorrect when tracking $T$. We construct these contrasts bidirectionally using tracker predictions. Complementary empty-mask contrasts require localizing the reference-specified target, while absent-target controls require preserving an empty prediction despite visible distractors (\cref{fig:training_data}, bottom). Together, these examples supervise target-conditioned correctness verdicts, and localization rather than decisions based on the query alone.

We optimize a masked autoregressive next-token objective over structured responses, training language and vision LoRA adapters together with the visual merger.

\paragraph{Active zoom training.}

We next train the corrector to request crops through supervised learning followed by reward-based refinement. The supervised cold start uses 7,582 dialogues, each with one or two crop requests, the returned crops, and a final assessment. To improve robustness, we include perturbed crops and visually similar distractors. Uninformative crop requests stay in the context but receive zero loss weight, so the model learns to answer after a poor crop without learning to request one. We use the same masked next-token objective and trainable components as in the first stage.

Since supervised demonstrations cannot cover the range of zooming choices encountered at inference, 
we refine the zoom policy using GRPO~\cite{shao2024deepseekmath} on 2,375 queries spanning the five tracking states. For each query, we sample eight dialogues and reward target visibility $v_t$ and bounding box $b_t$ against annotations $v_t^\star$ and $b_t^\star$. Valid responses receive:
\begin{equation}
R_t=
\begin{cases}
\operatorname{IoU}(b_t,b_t^\star), & v_t=v_t^\star=1,\\
1, & v_t=v_t^\star=0,\\
0, & \text{otherwise}.
\end{cases}
\label{eq:zoom_reward}
\end{equation}
Invalid dialogues receive zero reward. We use group-centered advantages $A_i=R_i-\frac{1}{8}\sum_{j=1}^{8}R_j$ without variance normalization~\cite{liu2025understanding}, avoiding amplification of small reward differences in low-variance groups.

\paragraph{Keeping the verdict out of RL.} The commitment gate relies on the verdict confidence, so refinement must keep this confidence calibrated. The GRPO reward, however, scores only visibility and the box (\cref{eq:zoom_reward}). If the verdict tokens were trained by the policy gradient, their probability would follow the reward of the box rather than the correctness of the verdict, and the confidence would no longer indicate whether the tracker is wrong. We therefore keep the verdict out of the policy gradient and supervise it directly:
\begin{equation}
\mathcal L_{\mathrm{refine}}
=
\mathcal L_{\mathrm{PG}}^{\neg d}
+\beta\mathcal L_{\mathrm{KL}}^{\neg d}
+\lambda_{\mathrm v}\mathcal L_{\mathrm{CE}}(d_t^\star),
\qquad \lambda_{\mathrm v}=0.02.
\label{eq:zoom_refinement}
\end{equation}
Here, $\neg d$ denotes exclusion of verdict tokens from policy-gradient and KL losses, while $\beta$ controls regularization against frozen model. The verdict tokens instead receive cross-entropy supervision toward the annotated verdict $d_t^\star$. Because this loss is computed inside each sampled dialogue, the verdict is learned conditioned on the crops the corrector itself chose, as at test time. During refinement, we freeze vision encoder and update only language LoRA (App.~\ref{sec:data_details}). %

\section{Experiments}
\label{sec:exp}
\paragraph{Datasets.}We evaluate recovery on four video object segmentation benchmarks: SA-V test (150 sequences, 278 objects), LVOS~v2 validation~\cite{hong2025lvos} (140 sequences, 239 objects), SeCVOS~\cite{zhang2025advancing} (160 sequences, 182 objects), and Cut-VOS~\cite{hu2026segment} (100 sequences, 173 objects), using their official metrics. SeCVOS and Cut-VOS contain scene changes, which interrupt temporal continuity. Failure-detection analyses use SA-V test, LVOS, and SeCVOS datasets.

\paragraph{Implementation details.}
We use the proposed failure detector and the LoRA-adapted Qwen3.5-4B corrector. %
Unless evaluating transfer, we use frozen SAM~3 initialized with a ground-truth mask at each target's first appearance. The corrector uses greedy decoding, a commitment threshold of \(\ell_t\ge\log 0.9\), and temporal context with \(U=4\), \(R=12\) (1\,s intervals), and \(\Delta=0.5\)\,s. Detector thresholds are tuned once on SA-V validation (App.~\ref{sec:method_details}).%

\paragraph{Evaluation metrics.}
We report tracking accuracy using mask IoU ($\mathcal{J}$), boundary F-score ($\mathcal{F}$), and their mean ($\mathcal{J}\&\mathcal{F}$). Failure detection is measured by episode recall and inspection precision. A failure occurs when a visible target has IoU below $0.5$, or a mask is predicted for an absent target. Consecutive failing annotated frames form an episode, and each inspection is assigned to the next annotated frame. Precision is the fraction assigned to failures, while recall is the fraction of episodes lasting at least one second (persistent) covered by an inspection. As a baseline, we place the same number of inspections per object at uniformly random frames, repeat this 1,000 times, and report the mean and the 2.5–97.5th percentile range.

\subsection{Trigger Analysis}
\label{sec:exp_trigger}
The trigger decides when the corrector inspects a frame. A good trigger inspects failing frames rather than correct ones, because an inspection of a correct frame can only keep the track or damage it. It inspects soon after a failure begins, and it works for other trackers without redesign. We evaluate our detector, C1A2, against these requirements.

\paragraph{Selective failure localization.}
C1A2 consistently localizes tracking failures more effectively than count-matched random inspection, detecting more failure episodes while achieving $1.4$--$2.0\times$ higher inspection precision across all three benchmarks (App.~\ref{sec:trigger_analysis}). Compared to Dual-Agent, C1A2 achieves higher inspection precision while substantially reducing unnecessary inspections. On SA-V, for instance, C1A2 detects 86\% of persistent failure episodes compared to 78\% for Dual-Agent, despite making nearly half as many requests. Consequently, 44\% of its requests coincide with tracking failures, compared to 25\% for random inspection and 16\% for Dual-Agent.

To understand this improved selectivity, we additionally compare against periodic inspection and confidence thresholding (App.~\ref{sec:trigger_baselines}). The former disregards tracking reliability, while the latter misses confident identity errors and reacts to transient uncertainty, requiring $1.7$--$2.7\times$ as many requests to match the failure coverage of C1A2's mask-based trigger. These results highlight the benefit of accumulating complementary tracker-internal evidence over time to selectively localize persistent failures.

\begin{figure}[t]
\centering
\includegraphics[width=0.9\linewidth]{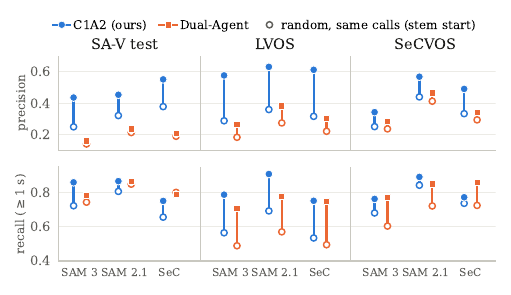}
\caption{Failure localization and tracker transfer. Stems connect each trigger's random baseline (circle) to its precision and recall.}
\label{fig:trigger_transfer}
\end{figure}

\paragraph{Generalization to other trackers.}
We assess the generality of C1A2 by transferring it to SAM~2.1 and SeC without refitting. Across both trackers and all three benchmarks, C1A2 achieves inspection precision 13--29\% above count-matched random inspection, exceeding the 97.5th percentile of 1,000 random runs (\cref{fig:trigger_transfer}, App.~\ref{sec:trigger_analysis}). This generalization is particularly notable given that the reliability of individual diagnostic signals varies across trackers. On SA-V, native-confidence AUROC drops from 0.83 for SAM~3 to 0.75 for SeC, whereas conditioning-memory consistency remains strongly discriminative (0.82 and 0.84, respectively). These results highlight how complementary diagnostic signals enable reliable failure detection across trackers despite differences in the reliability of their native confidence estimates.

\begin{figure}[t]
\centering
\includegraphics[width=0.9\linewidth]{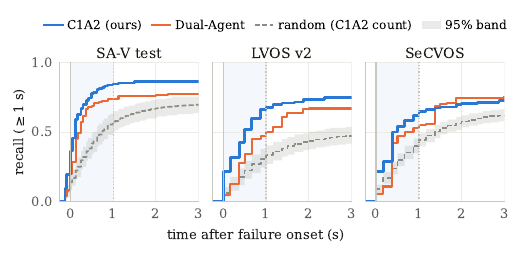}
\caption{Cumulative recall after failure onset for episodes $\geq 1$s. Gray: count-matched random mean and 95\% randomization band.}
\label{fig:trigger_onset}
\end{figure}

\paragraph{Detection near failure onset.}
Episode recall alone does not distinguish a prompt inspection from one made near the end of a failure.
We therefore measure the fraction of persistent episodes detected within a given time of their annotated onset (\cref{fig:trigger_onset}).
Within 1\,s, C1A2 detects 84\%, 68\%, and 65\% of episodes on SA-V, LVOS, and SeCVOS, exceeding count-matched random inspection at the same horizon by 21--34\%.
With the same number of requests per object, C1A2 detects more episodes within 1 s than random inspection detects over their full duration: 84\% versus 72\% on SA-V, and 68\% versus 57\% on LVOS. These results show that C1A2 mostly inspects near failure onsets.

\paragraph{Complementary evidence.}
Our ablations show that propagation-state and target-consistency signals provide complementary failure evidence (\cref{tab:signal_ablation}). Removing $s_1$ reduces precision by 3.4 points, confirming its discriminative value. Removing $s_2$ reduces recall by 14.0 points together with 20.6\% fewer inspection requests, indicating that it triggers a large share of the inspections that cover failures. 
The area-change criterion balances coverage and inspection cost: disabling it increases recall by 7.1 points but raises requests by 39.8\%. Lastly, presence monitoring addresses failures with empty predicted masks, where mask-dependent diagnostics are unavailable. It improves episode recall by 19--27 points across three benchmarks at 1.7--3.9$\times$ the requests, and 58--83\% of the persistent failures it detects are missed by the primary inspection channel. These findings show the value of combining complementary evidence with dedicated presence monitoring to detect diverse failures while controlling inspection overhead.

\paragraph{Failure detection during closed-loop recovery.}

\begin{figure}[t]
\centering
\includegraphics[width=0.8\linewidth]{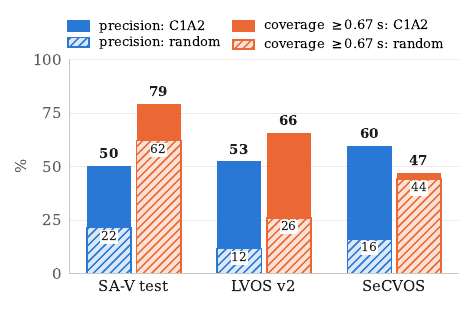}
\caption{Detection during oracle recovery. Blue: precision among applied updates (hatched: count-matched random requests). Orange: correction coverage of episodes lasting at least 0.67\,s.}
\label{fig:closed_loop_trigger}
\end{figure}

A trigger must remain informative as corrections change the memory used for subsequent propagation. We evaluate this feedback loop by using an oracle corrector that supplies an annotated target whenever the trigger fires, and compare C1A2 with random requests: the same number of requests per object, placed at uniformly sampled frames and answered by the same oracle. Across the three benchmarks, 50--60\% of C1A2's applied updates occur on failing predictions, compared with 12--22\% for random requests (\cref{fig:closed_loop_trigger}). The precision advantage shows that the trigger continues to direct updates toward failures as the tracking state changes.

Beyond inspection precision, C1A2 corrects more failures than random requests at every failure length, and longer failures more often. Tracker-internal evidence also remains discriminative between interventions: excluding the frames on which a correction is applied, $s_1$ reaches a failure AUROC of 0.78--0.86, about as high as on uncorrected SAM~3 predictions (App.~\ref{sec:trigger_analysis}). Together, these findings show that the detector retains both failure coverage and discriminative power throughout closed-loop recovery.

\begin{table}[t]
\centering
\small
\setlength{\tabcolsep}{5pt}
\caption{Component ablation pooled over three VOS benchmarks.}
\resizebox{0.9\columnwidth}{!}{%
\begin{tabular}{@{}lrrr@{}}
\toprule
Removed & $\Delta$Calls & $\Delta$R & $\Delta$P \\
\midrule
Target-consistency evidence $s_2$ & $-20.6\%$ & $-13.97$ & $-1.09$ \\
Propagation-state evidence $s_1$ & $-0.4\%$ & $-7.76$ & $-3.44$ \\
Area-change criterion & $+39.8\%$ & $+7.07$ & $-2.61$ \\
Presence monitoring & $-61.2\%$ & $-24.48$ & $+9.87$ \\
\bottomrule
\end{tabular}%
}
\label{tab:signal_ablation}
\end{table}

\begin{table*}[t]
\small
\newcommand{\na}{--}
\centering
\begin{minipage}[c]{0.85\textwidth}
\centering
\caption{Tracking performance on four VOS benchmarks under official protocols. SeCVOS separates temporal segments by the number of preceding scene changes.}
\label{tab:main}
\setlength{\tabcolsep}{3.4pt}
\resizebox{\linewidth}{!}{%
\begin{tabular}{@{}l ccc c ccc c cccc c ccc@{}}
\toprule
& \multicolumn{3}{c}{SA-V test} && \multicolumn{3}{c}{LVOS v2 val} && \multicolumn{4}{c}{SeCVOS (J\&F)} && \multicolumn{3}{c}{Cut-VOS} \\
\cmidrule{2-4}\cmidrule{6-8}\cmidrule{10-13}\cmidrule{15-17}
Method & J\&F & J & F && J\&F & J & F && 0 changes & 1 change & $\geq$2 changes & overall && J\&F & J & F \\
\midrule
\multicolumn{17}{@{}l}{\emph{SAM-based trackers}} \\
SAM~2.1~\cite{ravi2025sam} & 79.6 & 76.1 & 83.2 && 84.1 & 80.7 & 87.6 && 79.4 & 58.5 & 52.4 & 58.2 && 59.4 & 58.3 & 60.6 \\
SAMURAI~\cite{yang2024samurai} & 80.0 & \na & \na && 84.2 & \na & \na && 81.8 & 60.6 & 59.3 & 62.2 && \na & \na & \na \\
SAM2Long~\cite{ding2025sam2long} & 81.2 & 77.6 & 84.9 && 85.9 & 82.4 & 89.6 && 81.3 & 61.8 & 58.5 & 62.3 && \na & \na & \na \\
SeC~\cite{zhang2025advancing} & 81.7 & \na & \na && 86.5 & \na & \na && 84.2 & 69.6 & 67.5 & 70.0 && \na & \na & \na \\
SAAS~\cite{hu2026segment} & \na & \na & \na && \na & \na & \na && \na & \na & \na & \na && 62.0 & 60.5 & 63.6 \\
SAM~3~\cite{carion2026sam} & 84.4 & 80.4 & 88.4 && 88.2 & 84.4 & 92.2 && 87.1 & 74.6 & 69.3 & 74.2 && 72.9 & 71.1 & 74.8 \\
\midrule
\multicolumn{17}{@{}l}{\emph{Self-correcting trackers}} \\
Dual-Agent~\cite{bundele2026trackers} & 81.7 & 77.8 & 85.7 && 89.4 & 85.4 & 93.4 && \textbf{88.1} & 76.2 & 73.4 & 76.9 && 72.7 & 70.8 & 74.7 \\
\rowcolor{gray!10}
VIGIL (Ours) & \textbf{85.6} & \textbf{81.5} & \textbf{89.6} && \textbf{89.8} & \textbf{85.7} & \textbf{93.8} && 87.4 & \textbf{80.7} & \textbf{77.1} & \textbf{79.5} && \textbf{75.8} & \textbf{73.9} & \textbf{77.6} \\
\bottomrule
\end{tabular}}
\end{minipage}
\end{table*}

\begin{table}[t]
\centering
\small
\caption{Corrector ablations (J\&F). $\tau$: commitment threshold (Off: every proposed correction is applied). Middle: the three training stages (\cref{sec:method}); identity contrasts are removed from the first stage. Bottom: one component removed from the final model.}
\label{tab:corrector_ablation}
\setlength{\tabcolsep}{4pt}
\begin{tabular}{@{}lccc@{}}
\toprule
Corrector & $\tau$ & SA-V & LVOS \\
\midrule
SAM~3 alone & -- & 84.4 & 88.2 \\
\midrule
Base (identity-aware SFT) & Off & 83.5 & 88.8 \\
Base (identity-aware SFT) & 0.9 & 85.1 & 89.3 \\
\quad without identity contrasts & 0.9 & 84.9 & 88.4 \\
+ Zoom SFT & 0.9 & 84.8 & 88.7 \\
+ GRPO (final) & 0.9 & \textbf{85.6} & \textbf{89.8} \\
\midrule
Final, without the gate & Off & 81.3 & 87.7 \\
Final, without $\mathcal L_{\mathrm{CE}}(d_t^\star)$ & 0.9 & 84.8 & 89.2 \\
Final, without zoom & 0.9 & 84.8 & 89.7 \\
\bottomrule
\end{tabular}
\end{table}

\begin{table}[t]
\centering
\small
\caption{Quality of the verdict confidence $p$, measured on runs that apply every proposed correction. ECE: expected calibration error; Help: AUROC of $p$ for corrections that help vs.\ hurt.}
\label{tab:main_calibration}
\setlength{\tabcolsep}{4pt}
\begin{tabular}{@{}l cc cc@{}}
\toprule
& \multicolumn{2}{c}{SA-V test} & \multicolumn{2}{c}{LVOS v2 val} \\
\cmidrule(lr){2-3}\cmidrule(l){4-5}
Corrector & ECE$\downarrow$ & Help$\uparrow$ & ECE$\downarrow$ & Help$\uparrow$ \\
\midrule
Final model & 0.053 & \textbf{0.727} & 0.107 & \textbf{0.774} \\
Without $\mathcal L_{\mathrm{CE}}(d_t^\star)$ & 0.243 & 0.621 & 0.409 & 0.481 \\
Without zoom & 0.061 & 0.666 & 0.090 & 0.677 \\
\bottomrule
\end{tabular}
\end{table}

\begin{figure}[t]
\centering
\includegraphics[width=\linewidth]{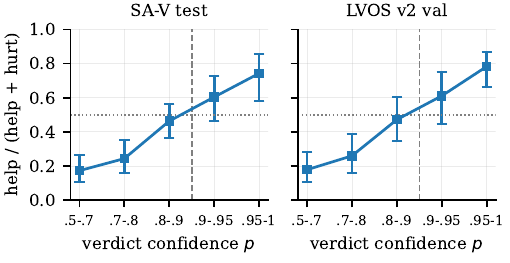}
\caption{Share of helpful corrections by verdict confidence $p$, among corrections that help or hurt, when every proposed correction is applied (Wilson 95\% intervals). Dashed: commitment threshold $\tau=0.9$.}
\label{fig:main_gate_help}
\end{figure}

\begin{table}[t]
\centering
\small
\caption{System cost on SA-V test, one RTX PRO 6000 per system. TFLOP: nominal, per video. GPU GiB: peak memory with all models loaded in PyTorch.}
\label{tab:main_cost}
\setlength{\tabcolsep}{4pt}
\begin{tabular}{@{}l rrrr@{}}
\toprule
System & s/video & TFLOP & GPU GiB & J\&F \\
\midrule
SAM~3 alone & 78.0 & 5,410 & 6.2 & 84.4 \\
\midrule
Dual-Agent~\cite{bundele2026trackers} & 410.1 & 8,641 & 47.1 & 81.7 \\
VIGIL (Ours) & \textbf{122.7} & \textbf{7,146} & \textbf{18.4} & \textbf{85.6} \\
\bottomrule
\end{tabular}
\end{table}

\begin{figure*}[t]
\centering
\includegraphics[width=\textwidth]{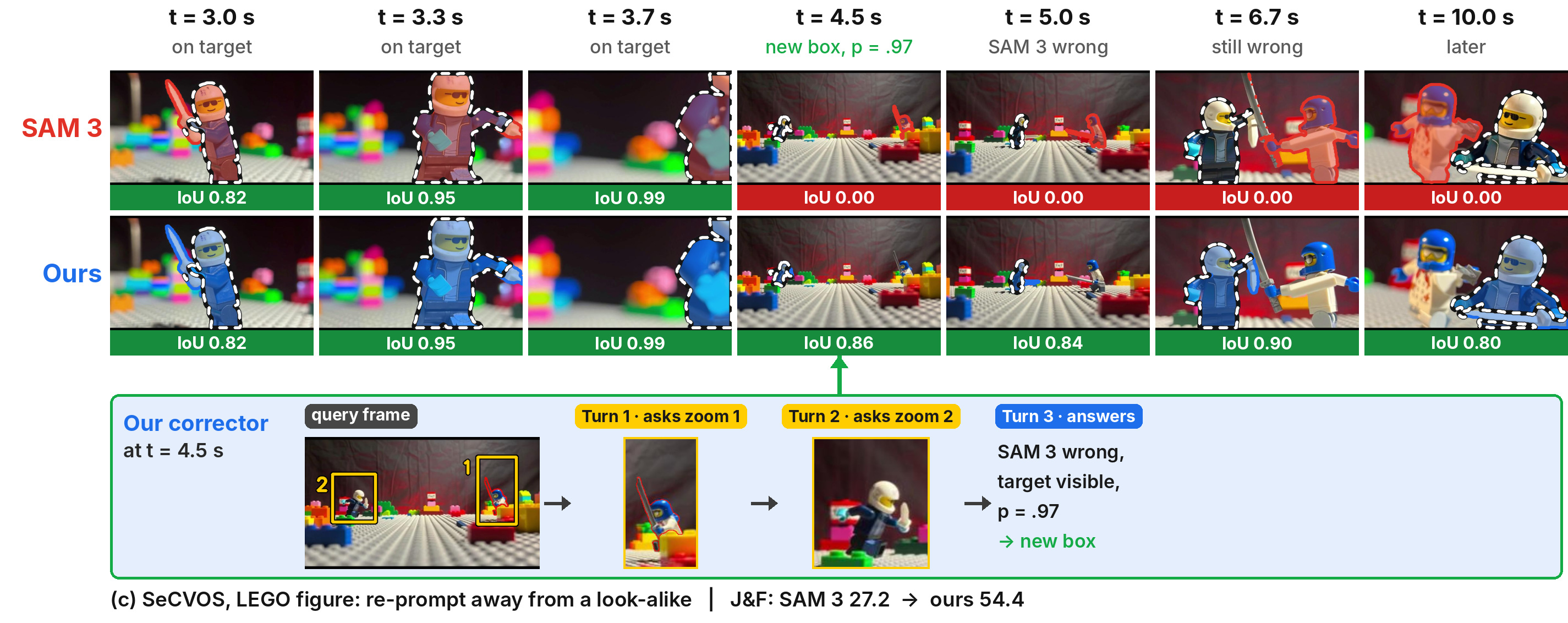}\\[6pt]
\includegraphics[width=\textwidth]{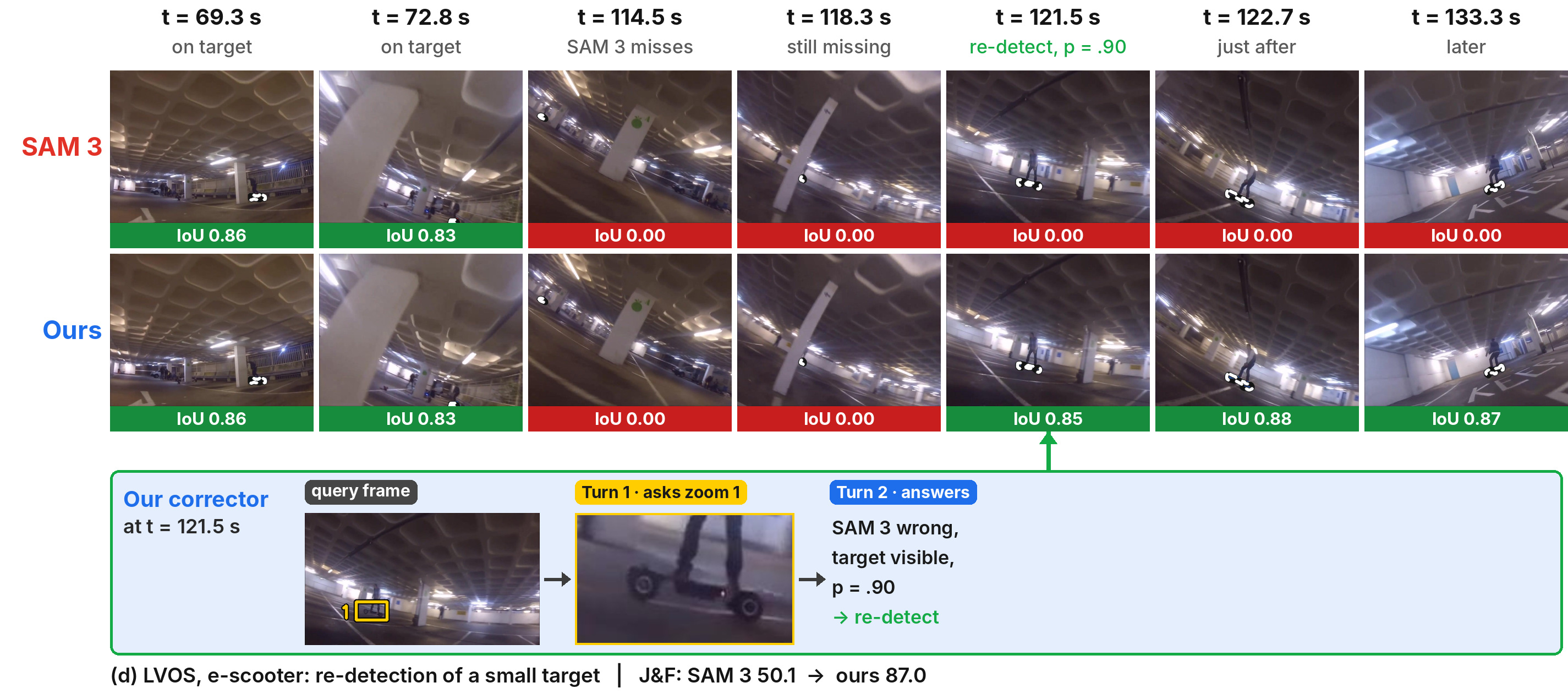}
\caption{SAM~3 alone (top row, red) vs.\ VIGIL (bottom row, blue). Dashed white line: annotated target; bar under each tile: mask IoU (green $\ge0.5$, red $<0.5$). Green header and arrow: the frame at which the correction is applied, with the action and the verdict confidence $p$. Blue strip: the corrector's turns: the zoom regions it requests (yellow, numbered), the crops it receives, and its answer. (c)~After SAM~3 switches to a look-alike, two crops separate the target from the look-alike, and the corrector re-seeds on the target. (d)~The target, an e-scooter, covers only 0.2\% of the frame and is lost by SAM~3; a crop resolves it, and the corrector re-detects it.}
\label{fig:main_qual}
\end{figure*}

\subsection{Recovery performance}
\label{sec:exp_corrector_main}
The complete framework, combining C1A2 with the unified 4B VLM corrector, achieves the highest overall $J\&F$ across all four benchmarks, outperforming both SAM~3 and Dual-Agent (\cref{tab:main}). On SeCVOS, the benefits of recovery become increasingly pronounced as temporal continuity is disrupted: the improvement over SAM~3 grows from 0.3 points before the first scene change to 7.8 points after multiple changes. These gains are achieved with substantially lower computational cost than Dual-Agent. By jointly assessing tracking correctness and recovery within a single 4B VLM, rather than using separate 8B verifier and 4B corrector models, our framework runs $3.3\times$ faster and reduces GPU memory requirements by a factor of 2.6, fitting on a single 24\,GB GPU (\cref{tab:main_cost}, App.~\ref{sec:cost}).

Importantly, while Dual-Agent degrades SAM~3's performance on SA-V and Cut-VOS, our framework consistently improves it across all benchmarks. This contrast highlights a fundamental challenge of closed-loop recovery: successful corrections must outweigh the errors introduced by incorrect interventions. Our framework addresses this through active visual inspection to resolve identity ambiguity and confidence-gated commitment to avoid unreliable memory updates, whose contributions we analyze below. Recovery nevertheless remains challenging under prolonged occlusions and near-identical distractors, where even additional visual evidence may be insufficient to establish target identity.


\paragraph{Transfer to other trackers.}
We transfer the trigger and corrector to SeC and SAM~2.1 without any adaptation (\cref{tab:closed_loop_transfer}, App.~\ref{sec:transfer_lvos}). Our method improves $J\&F$ by 6.5 points on SeC and 15.4 points on SAM~2.1, outperforming Dual-Agent on both trackers, and both trackers also improve on LVOS. The gain is largest on the weaker SAM~2.1: although it trails SeC by 11.8 points, the two recovered trackers perform comparably, so the corrector restores target identity even when the propagator is much weaker. SeC benefits as well, although it already uses vision-language reasoning, showing that explicit failure diagnosis and correction add capabilities beyond the tracker's own. The corrector was trained only on SAM~3 trajectories but needs no retraining, because it never accesses tracker internals: it sees the tracker's predicted masks only as outlines drawn on the frames, and it acts only through a box prompt or an empty-mask prompt, which every SAM~2-family tracker accepts. %

\begin{table}[t]
\centering
\small
\caption{Closed-loop transfer on SeCVOS and LVOS v2 val (J\&F).}
\setlength{\tabcolsep}{3pt}
\resizebox{0.85\columnwidth}{!}{%
\begin{tabular}{@{}lccccc@{}}
\toprule
& \multicolumn{4}{c}{SeCVOS} & LVOS \\
\cmidrule(lr){2-5}\cmidrule(l){6-6}
Tracker & 0 changes & 1 change & $\geq$2 changes & all & J\&F \\
\midrule
SeC & 84.2 & 69.6 & 67.5 & 70.0 & 86.5 \\
\quad + Dual-Agent & 84.6 & 72.2 & 70.5 & 72.0 & -- \\
\quad + VIGIL & \textbf{86.5} & \textbf{76.0} & \textbf{74.0} & \textbf{76.5} & \textbf{87.4} \\
\midrule
SAM~2.1 & 79.4 & 58.5 & 52.4 & 58.2 & 84.1 \\
\quad + Dual-Agent & 83.4 & 61.6 & 61.3 & 65.1 & -- \\
\quad + VIGIL & \textbf{85.2} & \textbf{73.5} & \textbf{70.5} & \textbf{73.6} & \textbf{86.3} \\
\bottomrule
\end{tabular}%
}
\label{tab:closed_loop_transfer}
\end{table}

\subsection{What makes recovery work}
\label{sec:exp_corrector_ablation}
\Cref{tab:corrector_ablation} removes one component at a time and measures closed-loop J\&F on SA-V and LVOS. To analyze the commitment gate, we also run the corrector without the gate, so that every proposed correction is applied, and label each correction \emph{helpful} if the target's J rises after it and \emph{harmful} if J falls (App.~\ref{sec:gate_calibration}). On these runs we measure two properties of the verdict confidence $p=\exp(\ell_t)$ (\cref{tab:main_calibration}): whether it is \emph{calibrated}, i.e., whether verdicts given with confidence $p$ are correct with probability close to $p$ (ECE), and whether it ranks helpful corrections above harmful ones (Help, an AUROC). Four findings emerge:

\begin{figure*}[t]
\centering
\includegraphics[width=0.72\textwidth]{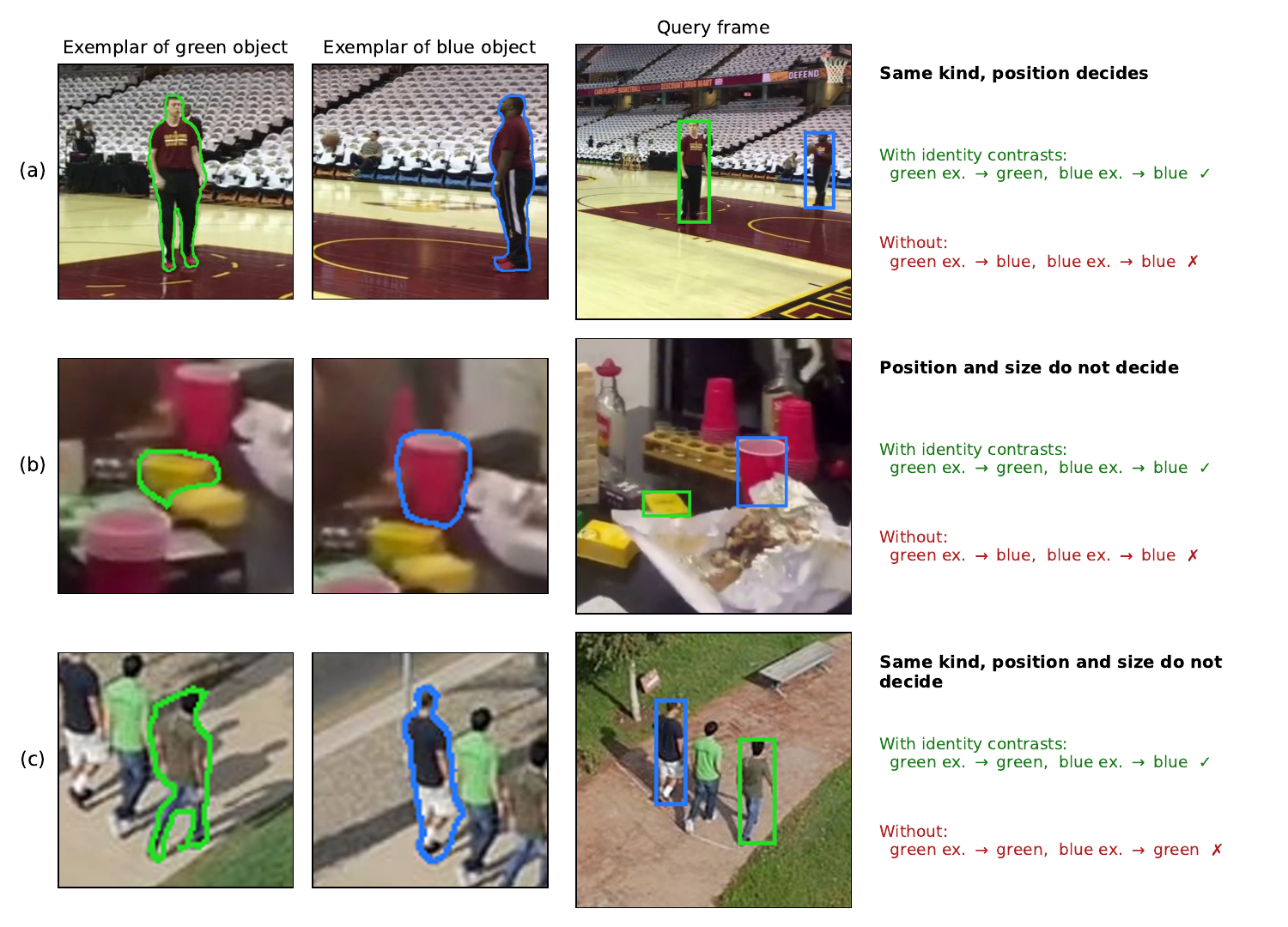}
\caption{Identity tests. The corrector receives the same frames with the exemplar of one of two visible objects; it passes if its choice follows the exemplar. Outlines mark the two annotated objects for illustration only: the corrector's input contains no mask. With identity contrasts, the choice switches with the exemplar; without them, the corrector picks the same object for both exemplars.}
\label{fig:main_identity_test}
\end{figure*}

\paragraph{Identity contrasts improve appearance-based target discrimination.} Removing the identity contrasts from training lowers closed-loop J\&F on both benchmarks, most on LVOS (\cref{tab:corrector_ablation}). J\&F, however, mixes identity errors with all other kinds of failure, so we also measure directly whether the corrector can tell which object is the target (App.~\ref{sec:calibration}, \cref{fig:main_identity_test}). Each test uses frames in which the target and a second annotated object are both visible. The corrector receives these frames twice: once with the exemplar of the first object, and once with the exemplar of the second. No mask is drawn, so SAM~3's prediction cannot reveal the answer. In each case, we compare the probabilities that the corrector assigns to the boxes of the two objects and take the more likely one as its choice. The corrector passes the test if its choice follows the exemplar, i.e., it chooses the first object when given the first exemplar and the second object when given the second.

Identity contrasts help most on the hardest tests. In these tests, the two objects are of the same kind, and the object closer to the exemplar in position or size is the wrong one, so only appearance can decide. On these tests, identity contrasts raise the pass rate from 46.8\% to 58.4\%. Without them, the corrector often chooses the same object for both exemplars, i.e., it ignores the exemplar (\cref{fig:main_identity_test}).

\paragraph{Effective self-correction requires the ability to abstain.}
Since corrections modify tracker memory, erroneous interventions can introduce persistent failures. Without gating, our corrector performs below SAM~3; with gating, it achieves the best performance.  The same effect explains why Dual-Agent falls below SAM~3 on SA-V and Cut-VOS: its harmful corrections outweigh its helpful ones. 

The effectiveness of gating stems from verdict confidence being predictive of correction utility: the proportion of helpful corrections increases steadily with confidence (\cref{fig:main_gate_help}), with helpful corrections outnumbering harmful ones by at least two to one at $p\geq0.9$. Verdict confidence thus enables selective recovery by filtering out unreliable interventions before they contaminate tracker memory.

\paragraph{Verdict supervision preserves reliable confidence estimates.}
The GRPO reward optimizes target visibility and localization but does not supervise verdict correctness (\cref{eq:zoom_reward}). Training with this reward alone leads to overconfident verdicts, with confidence becoming nearly uninformative in distinguishing helpful from harmful corrections on LVOS (Help AUROC approaches chance; \cref{tab:main_calibration}). This compromises selective commitment and degrades closed-loop J\&F (\cref{tab:corrector_ablation}). Adding a cross-entropy loss on the verdict, $\mathcal L_{\mathrm{CE}}(d_t^\star)$, during GRPO refinement (\cref{eq:zoom_refinement}) preserves both confidence calibration and its ability to discriminate beneficial from detrimental interventions, maintaining the effectiveness of the commitment gate.


\paragraph{Active zoom pays off after reward-based refinement.} The corrector sees each frame at a reduced resolution, which can leave a small target only a few pixels wide; zoom lets it request crops at the source resolution. Supervised demonstrations teach scripted crop choices, and GRPO then refines where to zoom based on the outcomes of the model's own dialogues. After this refinement, the corrector with zoom outperforms one trained identically without crops, and its confidence better separates helpful from harmful corrections, so the gate admits more reliable corrections (\cref{tab:main_calibration}). The learned crops land where the decision is made (App.~\ref{sec:errors}): on LVOS, when the target is present, the crop covers at least half of it in most calls (79\%). The actions taken from these crops are reliable. When a crop shows the target, the corrector re-detects it, and the re-detection helps in 39 of 45 cases across LVOS and SA-V. \Cref{fig:main_qual} shows two examples: a look-alike switch resolved by two crops, and a target too small to judge.

\section{Conclusion}
\label{sec:conclusion}
We present VIGIL, an evidence-driven self-correcting tracking framework combining training-free sequential failure diagnosis with unified VLM-based recovery. Active visual zoom resolves identity ambiguity, while confidence-aware commitment prevents unreliable memory updates. VIGIL consistently improves tracking across four VOS benchmarks. It also transfers to other video trackers without adaptation. Our analysis identifies what makes this possible: evidence combined across complementary signals for detection, a calibrated confidence that tells the corrector when to abstain, and identity-conditioned training that grounds its decisions in the target exemplar. These findings demonstrate that reliable failure diagnosis and selective correction can substantially improve tracking robustness.


{
    \small
    \bibliographystyle{ieeenat_fullname}
    \bibliography{main,refs}
}

\clearpage
\setcounter{page}{1}
\twocolumn[{%
\centering\Large\textbf{\thetitle}\\[0.5em]Supplementary Material\\[1.2em]
\normalsize\raggedright
\setlength{\fboxsep}{6pt}\noindent\colorbox{cvprblue!7}{\parbox{\dimexpr\textwidth-2\fboxsep\relax}{%
{\centering\small\textsc{Contents}\par}\vspace{4pt}
\begin{minipage}[t]{0.235\linewidth}\small\raggedright
\textbf{\ref{sec:data_details}\hspace{0.4em}Training the corrector}
\footnotesize\begin{itemize}\itemsep0pt\parsep0pt\topsep2pt\leftmargin6pt
\item[] \ref{sec:supervision_details}\hspace{0.4em}Examples and labels
\item[] \ref{sec:data_stage1}\hspace{0.4em}Stage 1: recovery corpus
\item[] \ref{sec:zoom_training_data}\hspace{0.4em}Stage 2: crop dialogues
\item[] \ref{sec:data_refinement}\hspace{0.4em}Stage 3: GRPO refinement
\item[] \ref{sec:shortcuts}\hspace{0.4em}Shortcut checks
\end{itemize}
\end{minipage}\hspace{3pt}{\color{cvprblue!40}\vrule width 0.6pt}\hspace{5pt}%
\begin{minipage}[t]{0.21\linewidth}\small\raggedright
\textbf{\ref{sec:method_details}\hspace{0.4em}Implementation details}
\footnotesize\begin{itemize}\itemsep0pt\parsep0pt\topsep2pt\leftmargin6pt
\item[] \ref{sec:supp_detector}\hspace{0.4em}Failure detector
\item[] \ref{sec:supp_inference}\hspace{0.4em}Corrector inference
\end{itemize}
\end{minipage}\hspace{3pt}{\color{cvprblue!40}\vrule width 0.6pt}\hspace{5pt}%
\begin{minipage}[t]{0.245\linewidth}\small\raggedright
\textbf{\ref{sec:trigger_analysis}\hspace{0.4em}Failure-detection analysis}
\footnotesize\begin{itemize}\itemsep0pt\parsep0pt\topsep2pt\leftmargin6pt
\item[] \ref{sec:measure}\hspace{0.4em}Evaluation protocol
\item[] \ref{sec:trigger_timing}\hspace{0.4em}Detection timing
\item[] \ref{sec:trigger_baselines}\hspace{0.4em}Simpler detectors
\item[] \ref{sec:trigger_ablation}\hspace{0.4em}Component ablation
\item[] \ref{sec:trigger_sensitivity}\hspace{0.4em}Sensitivity to the constants
\item[] \ref{sec:tracker_transfer}\hspace{0.4em}Transfer to SAM~2.1 and SeC
\item[] \ref{sec:oracle_trigger_details}\hspace{0.4em}Closed loop with an oracle
\end{itemize}
\end{minipage}\hspace{3pt}{\color{cvprblue!40}\vrule width 0.6pt}\hspace{5pt}%
\begin{minipage}[t]{0.245\linewidth}\small\raggedright
\textbf{\ref{sec:recovery_results}\hspace{0.4em}Additional recovery results}
\footnotesize\begin{itemize}\itemsep0pt\parsep0pt\topsep2pt\leftmargin6pt
\item[] \ref{sec:cost}\hspace{0.4em}Efficiency
\item[] \ref{sec:gate_calibration}\hspace{0.4em}Self-gated commitment
\item[] \ref{sec:supp_ablation}\hspace{0.4em}Training ablations
\item[] \ref{sec:errors}\hspace{0.4em}Active zoom
\item[] \ref{sec:transfer_lvos}\hspace{0.4em}Generalization
\item[] \ref{sec:qual}\hspace{0.4em}Qualitative examples
\end{itemize}
\end{minipage}
}}
\vspace{1.2em}
}]
\appendix
\renewcommand{\thesection}{\Alph{section}}
\setcounter{figure}{0}
\setcounter{table}{0}
\setcounter{equation}{0}
\renewcommand{\thefigure}{S\arabic{figure}}
\renewcommand{\thetable}{S\arabic{table}}
\renewcommand{\theequation}{S\arabic{equation}}
\renewcommand{\floatpagefraction}{0.75}
\renewcommand{\dblfloatpagefraction}{0.75}
\section{Training the corrector}
\label{sec:data_details}
\label{sec:supp_training}

The corrector is Qwen3.5-4B~\cite{qwen3.5} with LoRA adapters. It is trained in three stages, each on its own set of training data (\cref{tab:data_overview}):
\begin{itemize}\itemsep0pt
 \item \emph{Stage~1} (\cref{sec:data_stage1}) teaches the task: from the exemplar and the frames, decide whether SAM~3 is right and, if not, where the object is.
 \item \emph{Stage~2} (\cref{sec:zoom_training_data}) teaches the crop protocol: ask for zoomed crops of the query frame before answering.
 \item \emph{Stage~3} (\cref{sec:data_refinement}) refines the whole dialogue with reinforcement learning, so that the corrector learns which crops lead to correct answers.
\end{itemize}
\Cref{tab:supp_train} lists the training settings.
All stages use bf16, seed 42, the same chat template without thinking, and the same image budget as inference.
All three training sets come only from SA-V \emph{training} videos.

\begin{table}[!ht]
 \centering
 \caption{\textbf{Training sets} from SA-V train videos.}
 \label{tab:data_overview}
 \small
 \setlength{\tabcolsep}{2.5pt}
 \begin{tabular}{@{}lrrcl@{}}
  \toprule
  Set & Units & Videos & Shards & Stage \\
  \midrule
  Recovery corpus & 90,379 & 22,834 & 000--039 & 1: recovery SFT \\
  Crop dialogues & 7,582 & 5,391 & 000--054 & 2: crop SFT \\
  Refinement queries & 2,375 & 2,032 & 000--054 & 3: GRPO \\
  \bottomrule
 \end{tabular}
\end{table}

\begin{table*}[t]
\centering
\small
\caption{\textbf{Training settings.}}
\label{tab:supp_train}
\setlength{\tabcolsep}{4pt}
\resizebox{\textwidth}{!}{%
\begin{tabular}{@{}l lll@{}}
\toprule
& Stage 1: recovery SFT & Stage 2: crop-dialogue SFT & Stage 3: GRPO refinement \\
\midrule
Initialization & Qwen3.5-4B & stage-1 model & stage-2 adapters merged into the base \\
Data & 90,379 training examples, 1 epoch & 7,582 dialogues, 1 epoch & 2,375 queries \\
Trainable & LM LoRA, vision LoRA, merger & LM LoRA, vision LoRA, merger & new LM LoRA; vision tower and merger frozen \\
LoRA rank / scale / dropout & 16 / 32 / 0.05 & 16 / 32 / 0.05 & 16 / 32 / 0 \\
Learning rate & LM, vision $10^{-4}$; merger $2\times10^{-5}$ & LM, vision $10^{-4}$; merger $2\times10^{-5}$ & $10^{-5}$ \\
Schedule & cosine, 5\% warmup & cosine, 5\% warmup & 20 warmup steps, then constant \\
Weight decay & 0.01 & 0.01 & -- \\
Effective batch & 32 & 32 & 64 (8 queries $\times$ 8 dialogues) \\
Steps & 2,825 & 237 & 130 \\
Max.\ sequence & 24,576 tokens & 24,576 tokens & 28,672 tokens; 192 per completion \\
Loss & CE on responses & CE; unhelpful crop requests masked & GRPO + verdict CE ($\lambda_{\mathrm v}=0.02$) \\
Framework & LLaMA-Factory, DeepSpeed ZeRO-2 & LLaMA-Factory, DeepSpeed ZeRO-2 & ms-swift 4.2.0.dev0, TRL 0.24.0, vLLM \\
\bottomrule
\end{tabular}}
\end{table*}

\subsection{Training examples and labels}
\label{sec:supervision_details}

\begin{figure*}[t]
\centering
\includegraphics[width=\textwidth]{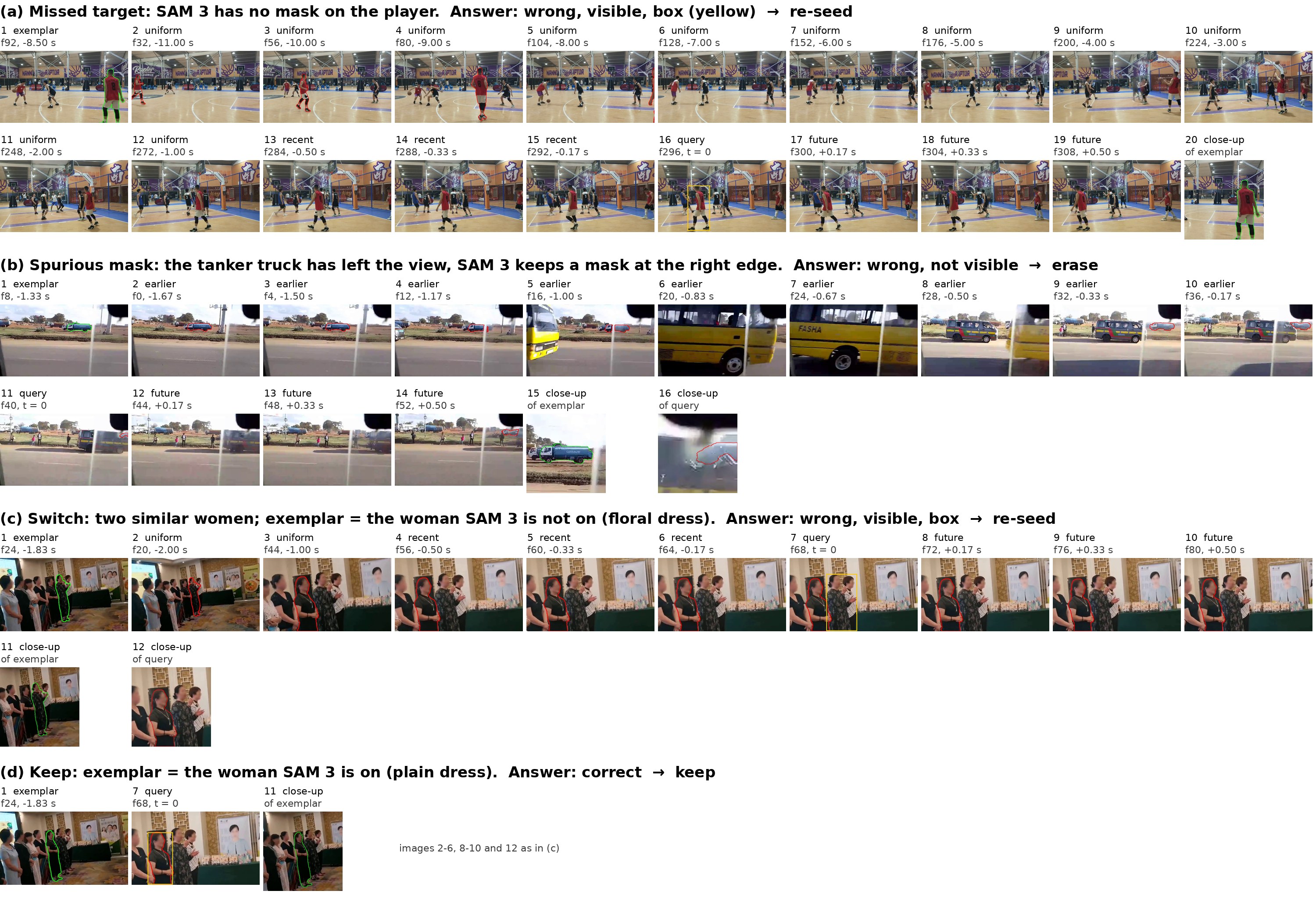}
\caption{\textbf{Four stage-1 training examples}: every image the corrector receives, numbered in input order, and the target output. Earlier frames are spread over the tracked history (\emph{uniform}) or just before the query (\emph{recent}); Green: annotated mask; red: SAM~3's mask; yellow: target box; labels give the SA-V frame index and the time relative to the query (for illustration only). (a)~SAM~3 has lost the basketball player. (b)~The tanker truck has left the view, but SAM~3 keeps a small mask at the right edge. (c,d)~Two women in similar dark dresses, one floral and one plain. SAM~3 is on the woman in the floral dress in image~2 and on the woman in the plain dress from image~3 on. Both examples have the same frames and only the exemplar differs, so (c) needs a re-seed to the woman in the floral dress and (d) a keep.}
\label{fig:train_examples}
\end{figure*}

\paragraph{Training examples.}
We run frozen SAM~3 on SAV-train videos, starting each object at its first annotated frame, and store its masks.
A training example is one annotated frame of an object, the \emph{query frame}, together with the input the corrector receives at that frame: the \emph{exemplar} (an annotated view of the object), earlier frames, and three later frames, each with SAM~3's predicted mask drawn as an outline (\cref{sec:supp_inference}; \cref{fig:train_examples}).
The target output states whether SAM~3's mask on the query frame is correct and, if not, where the object is.
All masks on the earlier, query and later frames are SAM~3's predictions.

\paragraph{Annotation states.}
SA-V annotates every fourth frame.
On an annotated frame the object is either \emph{present} (non-empty annotated mask) or \emph{absent} (empty annotated mask).
On unannotated frames its state is unknown, so only annotated frames are used as query frames.

\paragraph{Situations and actions.}
Comparing the annotation with SAM~3's predicted mask $\widehat M$ on the query frame gives five situations (\cref{tab:training_cells}). IoU below is the mask IoU between $\widehat M$ and the annotated mask.
\begin{itemize}\itemsep0pt
 \item \emph{Correct track}: the object is present and SAM~3's mask covers it well ($\widehat M$ non-empty, IoU $\ge0.6$). Action: \textsc{keep}.
 \item \emph{Incorrect mask}: the object is present, but SAM~3's mask has a poor shape or lies on another object ($\widehat M$ non-empty, IoU $<0.4$). Action: \textsc{reseed} with a new box.
 \item \emph{Missed target}: the object is present, but SAM~3 has lost it and outputs no mask ($\widehat M$ empty). Action: \textsc{reseed}.
 \item \emph{Spurious mask}: the object is absent (e.g.\ occluded or out of view), but SAM~3 still outputs a mask ($\widehat M$ non-empty). Action: \textsc{erase}.
 \item \emph{Correct absence}: the object is absent and SAM~3 correctly outputs no mask ($\widehat M$ empty). Action: \textsc{keep}.
\end{itemize}
Frames in between ($0.4\le\text{IoU}<0.6$) are not used as query frames, so every verdict label is clear-cut.

\paragraph{What the corrector is trained to output.}
For the query frame, the corrector outputs three things:
the \emph{verdict}, whether SAM~3 is right (\texttt{correct} in the two keep situations, \texttt{wrong} otherwise);
the \emph{visibility}, whether the object is present;
and the \emph{box} of the object, the bounding box of its annotated mask in normalized $[0,1000]$ full-frame coordinates, or zero when it is absent.
In addition, it judges SAM~3 on each earlier frame it is shown: a history vector $h$ holds one binary flag per earlier frame, 1 if SAM~3's mask on that frame was correct and 0 otherwise.
Here ``correct'' means both absent, or both present with mask IoU $\ge0.5$.
\Cref{fig:train_examples} shows four examples.
\newpage

\subsection{Stage 1: recovery corpus}
\label{sec:data_stage1}

\subsubsection{Exemplar and context frames}
Each training example follows the test-time input layout (\cref{sec:supp_inference}):
\begin{itemize}
 \item \emph{Exemplar}: the annotated mask on the object's first annotated frame. It is replaced by the largest annotated view of the object if that view has at least three times the area and lies at least 1\,s before the query, so that the reference is easier to read; some examples keep a small first view, so that training also contains small references. At test time the exemplar is always the given initial mask.
 \item \emph{Earlier frames}: 12 recent frames, of which up to three lie within the last 0.5\,s and the rest are spaced about 1\,s apart, and up to four evenly spaced older frames, each with SAM~3's mask drawn in red.
 \item \emph{Later frames}: the frames 4, 8 and 12 frames after the query on SA-V, with the masks that a separate copy of SAM~3 predicts for them.
 \item \emph{Close-ups}: of the exemplar, and of SAM~3's query mask whenever it is non-empty. In stage~2 the query close-up is replaced by crops that the model requests itself.
 \item \emph{Time stamps}: each frame is labeled $t_0+i/24$ seconds for frame index $i$, with a per-example offset $t_0\sim\mathcal U[0,300]$, so that absolute time does not identify a video.
\end{itemize}

\subsubsection{Mix of situations}
The stage-1 corpus has 90,379 training examples, and every situation has at least 9,580 of them (\cref{tab:training_cells}).
Summed over situations, 49,635 training examples ($55\%$; correct track plus correct absence) have the verdict \texttt{correct} and 40,744 ($45\%$; the other three situations) the verdict \texttt{wrong}; the object is present in 55,153 ($61\%$) and absent in 35,226 ($39\%$).

\begin{table}[!ht]
 \centering
 \caption{\textbf{Stage-1 corpus}, with the action each situation requires.}
 \label{tab:training_cells}
 \small
 \setlength{\tabcolsep}{4pt}
 \begin{tabular}{lcrr}
  \toprule
  Situation & Action & Count & \% of corpus \\
  \midrule
  Correct track & Keep & 33,463 & $37.0\%$ \\
  Incorrect mask & Re-seed & 9,580 & $10.6\%$ \\
  Missed target & Re-seed & 12,110 & $13.4\%$ \\
  Spurious mask & Erase & 19,054 & $21.1\%$ \\
  Correct absence & Keep & 16,172 & $17.9\%$ \\
  \midrule
  Total & & 90,379 & $100\%$ \\
  \bottomrule
 \end{tabular}
\end{table}

\subsubsection{How query frames are chosen}
The candidates are the annotated frames of every object in the SAM~3 runs on the training videos.
Each candidate falls into one of the five situations, by comparing SAM~3's mask on that frame with the annotation.
A sampler then picks query frames from the candidates with four rules.
\begin{itemize}
 \item \emph{Remove objects and frames that cannot be learned from.} An object is removed if SAM~3 is already wrong when tracking starts (mask IoU $<0.5$ with the annotation on the initialization frame), because its later errors are not tracking failures. A frame is removed if the target is too small: annotated box shorter side below 40 pixels at 640-pixel image width, or area below $0.6\%$ of the frame.
 \item \emph{Prefer frames where the failure detector fires.} We run our failure detector (\cref{sec:trigger}) over the SAM~3 rollouts and prefer the frames where it would call the corrector, because these resemble the frames the corrector sees at test time. If the detector fires on an unannotated frame, we use the next annotated frame.
 \item \emph{Spread the frames in time.} For the same object and situation, chosen frames are at least 1\,s apart for missed targets and spurious masks and 2\,s apart otherwise, so that one long failure does not fill the corpus.
 \item \emph{Keep every failure, thin out correct frames.} Every candidate where SAM~3 is wrong is kept. Correct tracks and correct absences are more common, so they are subsampled, and each object contributes only a few query frames.
\end{itemize}

\begin{figure*}[t]
\centering
\begin{minipage}[t]{0.24\textwidth}\centering\includegraphics[width=\linewidth]{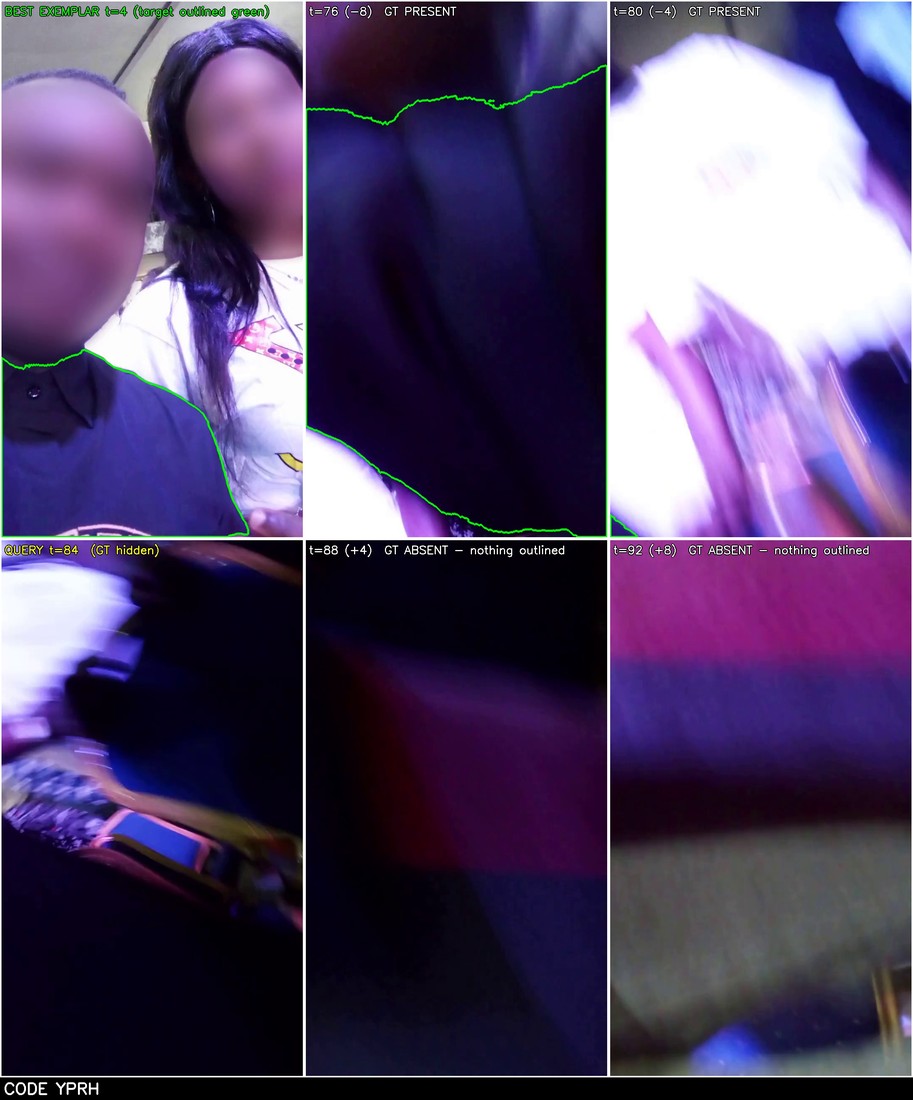}\\[2pt]\raggedright\scriptsize \textbf{(a) Question 1: removed}\\Annotation: present. Answer: \emph{cannot tell}\\``The query is very dark and motion-blurred.''\end{minipage}\hfill
\begin{minipage}[t]{0.24\textwidth}\centering\includegraphics[width=\linewidth]{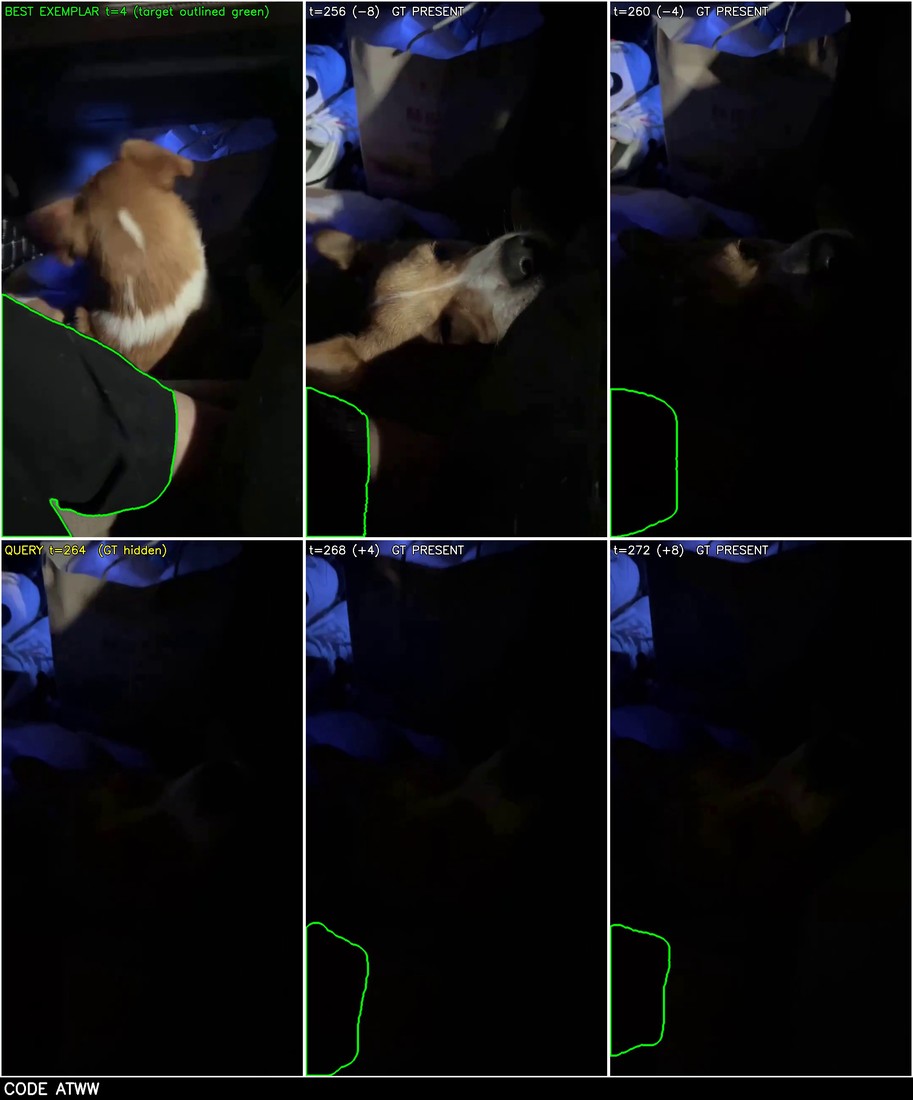}\\[2pt]\raggedright\scriptsize \textbf{(b) Question 1: removed}\\Annotation: present. Answer: \emph{cannot tell}\\``The query frame is too dark there to tell whether any of it is present.''\end{minipage}\hfill
\begin{minipage}[t]{0.24\textwidth}\centering\includegraphics[width=\linewidth]{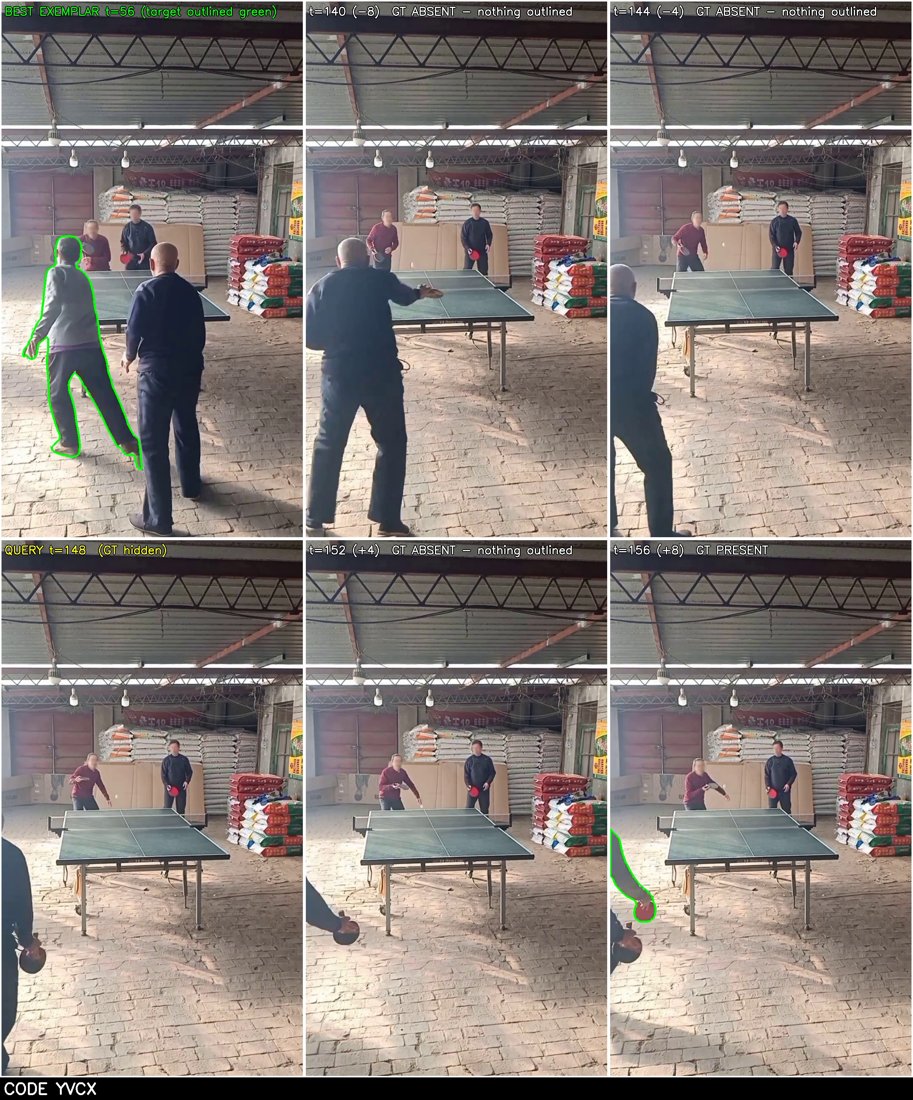}\\[2pt]\raggedright\scriptsize \textbf{(c) Question 1: removed}\\Annotation: absent. Answer: \emph{visible} (disagrees)\\``Its paddle and sleeve are still in the query frame at lower left, matching the t=156 annotation.''\end{minipage}\hfill
\begin{minipage}[t]{0.24\textwidth}\centering\includegraphics[width=\linewidth]{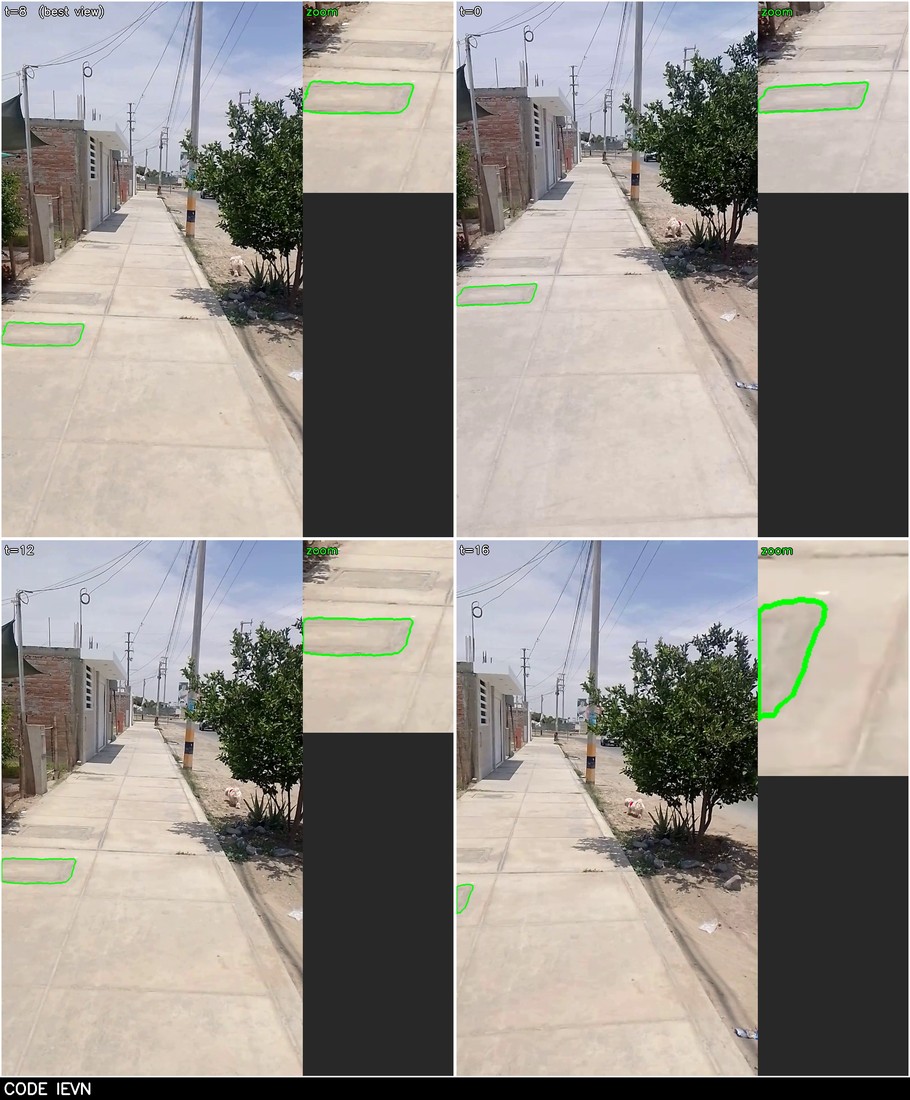}\\[2pt]\raggedright\scriptsize \textbf{(d) Question 2: removed}\\Answer: \emph{no}\\``The outlined target is a plain patch of concrete sidewalk surface with no distinct edges, markings, or parts.''\end{minipage}
\caption{\textbf{Examples removed by Haiku}, with an excerpt of its reason. (a,b)~Question~1: the query frame (bottom left) is too blurred or too dark to show the target. (c)~Question~1: the annotation misses a visible target. (d)~Question~2: the target is a featureless patch of ground. The separate unmarked query image is not shown.}
\label{fig:haiku}
\end{figure*}

\subsubsection{Can the object be recognised?}
A label can only be learned if the input shows it. In SA-V, this fails in two ways: (i) the query frame does not show whether the target is present, because it is too dark or motion-blurred, or only a tiny part of the target is visible; (ii) the target cannot be identified from its views at all, as for one without distinctive features, such as a patch of ground (\cref{fig:haiku}d). Training on such examples would teach the corrector to answer without visual evidence and would make its verdict unreliable.

We remove them with an external VLM, Claude Haiku (\texttt{claude-haiku-5-5}), which answers one question per case: Question 1 checks that the visibility label can be read from the query frame, and Question 2 checks that the target can be identified from its annotated views. Each question has its own input.

\paragraph{Question 1: is the target visible in the query frame?}
\begin{itemize}
 \item \emph{Input} (\cref{fig:haiku}a--c): a sheet with the exemplar and the four annotated frames nearest to the query, at frame offsets $-8$, $-4$, $+4$ and $+8$ relative to it, each with the annotated outline or the note ``absent''; and, as a second image, the query frame by itself at the resolution of the stored SA-V frame (e.g.\ $720\times1280$ or $1080\times1920$ pixels), with no outline or label drawn on it.
 \item \emph{Answers}: visible, not visible, or cannot tell.
 \item \emph{Removed if} the answer is \emph{cannot tell}, or if it disagrees with the annotation (\emph{visible} although annotated absent, or \emph{not visible} although annotated present).
\end{itemize}

\paragraph{Question 2: could the target be found again?}
\begin{itemize}
 \item \emph{Input} (\cref{fig:haiku}d): a sheet with four annotated views of the target from across the video, each a full frame and a zoom, none within 8 frames of the query.
 \item \emph{Question}: ``if you were shown a new frame of this video, could you find this exact object again, using only what these four views show?''
 \item \emph{Answers}: \emph{yes} or \emph{no}. \emph{Removed if} the answer is \emph{no}.
\end{itemize}

\paragraph{Result.} Haiku judged 123,959 candidate training examples and removed 32,284 of them ($26\%$).

\subsubsection{Identity contrasts}
\label{sec:identity_contrasts}
An identity switch, in which SAM~3 moves from the target to a similar-looking object, is the hardest failure to detect: the query frame alone does not reveal it, and only the exemplar identifies the target. Such switches are rare in natural data. We therefore add \emph{identity contrasts}, i.e.\ training examples that can be solved only by comparing the objects in view with the exemplar and context. In total, 7,441 training examples ($8.2\%$ of the corpus) involve an identity confusion: 5,036 constructed look-alike examples and 2,405 natural identity switches, in which SAM~3's mask at the query lies on another annotated object.

\paragraph{Finding look-alikes.}
An external multimodal model (TypeSafe Jev~1.14) is shown the target outlined in green and another annotated object of the same video outlined in blue, and is asked ``Could the object in the BLUE outline be mistaken for the target object (GREEN outline)?''. Two objects are look-alikes if probability of ``yes'' is at least $0.5$.

\paragraph{Example kinds.}
Let $T$ be the target and $L$ its look-alike, and let a \emph{touch} be a frame in which their annotated boxes intersect. The kinds are designed in pairs, so that within a pair a single input decides the answer: 

\begin{itemize}\itemsep0pt
 \item \emph{Keep} (answer \textsc{keep}): SAM~3 follows the target correctly from a touch to the query. These are real tracks, of $T$ with $T$'s exemplar (keep~$T$) and of $L$ with $L$'s exemplar (keep~$L$).
 \item \emph{Switch} (answer \textsc{reseed} on $T$): the frames of a keep~$L$ example with $T$'s exemplar, so SAM~3 follows $L$ while target is $T$. Keep~$L$ and switch share all frames and differ only in exemplar: in \cref{fig:train_examples}c,d, SAM~3 follows the same woman, and the answer is \textsc{reseed} with other woman's exemplar (c) and \textsc{keep} with this woman's exemplar (d).
 
 \item \emph{Missed target} (answer \textsc{reseed} on $T$): SAM~3 shows no mask on the last earlier frames and the query, although $T$ is visible. Besides natural cases, we construct missed targets at a query frame where SAM~3 follows both objects correctly and both are visible: in one copy $T$ is the target and SAM~3's mask of $T$ is removed from the $k$-th last earlier frame on, including the query and the later frames; a second copy does the same with $L$ as the target. Here $k$ is drawn from the lengths of real empty runs. The two copies share the query frame, which shows no mask, and differ only in the exemplar and the answer box.
 
 \item \emph{Absent target} (answer \textsc{keep}): SAM~3 correctly shows no mask because $T$ has left the view, while $L$ is visible. Each absent target is matched to a missed target with the same number of empty frames ($\pm1$), so the two kinds differ only in whether $T$ is in view.
 \item \emph{Natural switch} (answer \textsc{reseed} or \textsc{erase}): SAM~3's own mask at the query lies on another annotated object.
\end{itemize}

\paragraph{Shortcut controls.}
All examples built from one pair are used together, and the object closer to the exemplar in position and size is the target as often as it is the look-alike, so these cues do not reveal the answer. Pairs where one object contains the other, crowded scenes and objects annotated in several pieces are removed.

\subsubsection{Training}
The model is trained with the standard next-token loss on the answer only.
LoRA covers all linear projections of the language model (attention, linear attention and MLP) and the attention and MLP layers of the vision encoder, and the visual merger is trained fully.
We use the final checkpoint.

\subsection{Stage 2: crop dialogues}
\label{sec:zoom_training_data}

\paragraph{Motivation.}
Stage~1 trains a single-turn task: the corrector answers directly from its input. At test time, the corrector first requests one or two crops of the query frame, which are returned at source resolution, with magnification capped at 6$x$. Stage~2 is a short supervised stage that teaches this multi-turn protocol: the corrector learns to request a crop before answering, to request at most one more, and to answer correctly whether or not a crop is informative.

\begin{table}[t]
\centering
\caption{\textbf{Stage-2 dialogues and stage-3 queries} per situation and object size.}
\label{tab:zoom_training_cells}
\label{tab:rl_cells}
\small
\setlength{\tabcolsep}{2.6pt}
\resizebox{\linewidth}{!}{%
\begin{tabular}{@{}l rrrr rrrr@{}}
\toprule
& \multicolumn{4}{c}{Stage-2 dialogues} & \multicolumn{4}{c}{Stage-3 queries} \\
\cmidrule(lr){2-5}\cmidrule(l){6-9}
Situation & All & Small & Med. & Large & All & Small & Med. & Large \\
\midrule
Correct track   & 1,900 & 764 & 533 & 603 & 450 & 122 & 153 & 175 \\
Incorrect mask  & 1,452 & 325 & 586 & 541 & 475 & 137 & 102 & 236 \\
Missed target   & 1,618 & 164 & 630 & 824 & 500 & 143 & 139 & 218 \\
Spurious mask   &   637 & -- & -- & -- & 450 & -- & -- & -- \\
Correct absence & 1,975 & -- & -- & -- & 500 & -- & -- & -- \\
\midrule
Total           & 7,582 & 1,253 & 1,749 & 1,968 & 2,375 & 402 & 394 & 629 \\
\bottomrule
\end{tabular}}
\end{table}

\subsubsection{Crop dialogues}
\begin{figure*}[t]
\centering
\includegraphics[width=\linewidth]{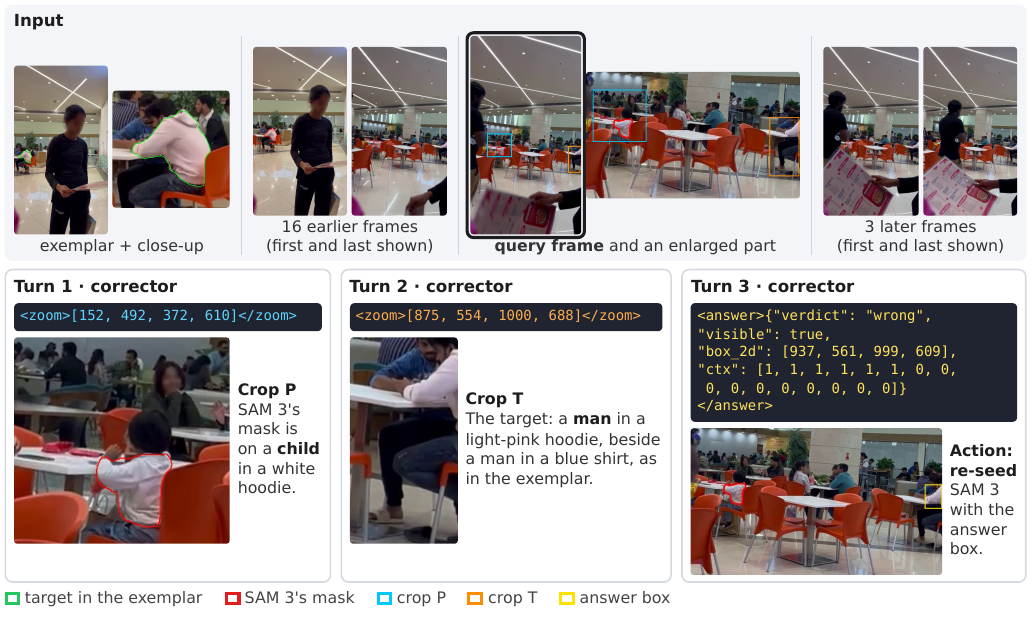}
\caption{\textbf{A crop dialogue from the stage-2 training data}. The target is a man in a light-pink hoodie at a food-court table; its box covers 0.3\% of the query frame. In the full query frame, SAM~3's object and the target are too small to tell apart. Crop P shows that SAM~3's mask is on a child in a white hoodie. Crop T shows the target beside a man in a blue shirt, as in the exemplar. The corrector answers \texttt{wrong} with the target's box, and SAM~3 is re-seeded with this box. All three turns are trained.}
\label{fig:crop_dialogue}
\end{figure*}

A crop dialogue is a sequence
\begin{equation}
 D=\bigl(x,\; b_1, c_1,\; \dots,\; b_K, c_K,\; y\bigr),\qquad K\in\{1,2\},
 \label{eq:crop_dialogue}
\end{equation}
with the following parts (\cref{fig:crop_dialogue}).
\begin{itemize}
 \item $x$, the \emph{input}: the stage-1 input without the close-up of SAM~3's query mask, which the crops replace.
 \item $b_k$, the $k$-th \emph{crop request}, written by the model: a box $(x_1,y_1,x_2,y_2)$ in query-frame coordinates normalized to $[0,1000]$.
 \item $c_k$, the $k$-th \emph{crop}, returned to the model: region $b_k$ cut from the full-resolution query frame and resized as at inference (\cref{sec:supp_inference}).
 \item $y$, the \emph{final answer}: the same outputs as in stage~1 (verdict, visibility, box and history vector $h$).
\end{itemize}
The first crop is required, as at test time, and the second is optional. Of the 7,582 dialogues, 4,257 have one crop and 3,325 have two.
All five situations are included (\cref{tab:zoom_training_cells}). A correct track and a correct absence also start with a crop, so asking for a crop never implies that SAM~3 is wrong.

\subsubsection{How the crops are chosen}
In the training dialogues, the crop requests $b_k$ are scripted. There are five kinds of crop. Two show what the answer depends on: P, a crop of SAM~3's mask, and T, a crop of the true (annotated) object. Three are uninformative: L, a crop of a look-alike or another object; R, a random region; and O, a region next to the true object that misses it.

\paragraph{Which crops a dialogue contains.}
Each situation has a \emph{base sequence} of informative crops, i.e.\ the regions its answer depends on (\cref{tab:zoom_crop_seq}).
\begin{itemize}
 \item If SAM~3 shows a mask, the first crop checks it (P).
 \item If that mask is on the wrong object or region and the object is present (incorrect mask), the second crop locates the object for \textsc{reseed} (P$\to$T).
 \item If SAM~3 has lost the object (missed target), the crop goes to the object (T).
 \item If the object is correctly absent, there is nothing to crop, so the crop is L or R.
\end{itemize}
In part of the dialogues, an uninformative crop is added or replaces T (\cref{tab:zoom_crop_seq}, right), so that the model also learns to answer correctly after an uninformative crop.
An uninformative crop does not reveal the answer: the dialogues with an L crop end in \textsc{keep} ($63.2\%$), \textsc{reseed} ($33.0\%$) or \textsc{erase} ($3.8\%$), and those with an R crop in \textsc{keep} ($75.5\%$), \textsc{reseed} ($21.8\%$) or \textsc{erase} ($2.6\%$). O crops occur only in missed targets, always followed by the T crop.

\begin{table}[t]
 \centering
 \caption{\textbf{Crop sequences} per situation: dialogues with only informative crops (base) and with an uninformative crop. Letters as in \cref{sec:zoom_training_data}.}
 \label{tab:zoom_crop_seq}
 \small
 \setlength{\tabcolsep}{3pt}
 \resizebox{\linewidth}{!}{%
 \begin{tabular}{@{}l l r l r@{}}
  \toprule
  & \multicolumn{2}{c}{Base} & \multicolumn{2}{c}{With an uninformative crop} \\
  \cmidrule(lr){2-3}\cmidrule(l){4-5}
  Situation & Sequence & Dialogues & Sequences & Dialogues \\
  \midrule
  Correct track   & P            & 1,519 & P$\to$L, P$\to$R            & 381 ($20\%$) \\
  Incorrect mask  & P$\to$T      &   813 & P$\to$L, P$\to$R            & 639 ($44\%$) \\
  Missed target   & T            &   649 & L$\to$T, R$\to$T, O$\to$T   & 969 ($60\%$) \\
  Spurious mask   & P            &   509 & P$\to$L, P$\to$R            & 128 ($20\%$) \\
  Correct absence & L or R       & 1,580 & two of L, R                 & 395 ($20\%$) \\
  \midrule
  Total           &              & 5,070 &                             & 2,512 \\
  \bottomrule
 \end{tabular}}
\end{table}

\paragraph{Where each crop is placed.}
P and T crops are loose and off-center, because a box that fits the object exactly would reveal where the correct object is. For an object box of width $w_b$ and height $h_b$, the crop has size $(r\,w_b,r\,h_b)$ with a random factor $r\in[1.5,3]$, and its center is moved from the box center by at most $0.45(r-1)\,w_b$ horizontally and $0.45(r-1)\,h_b$ vertically; crops that extend beyond the frame are clipped to it. The crop always contains the whole object.

The uninformative crops (L, R and O) are drawn from the following sources, in order of preference: look-alikes mined with JEV (\cref{sec:identity_contrasts}), other annotated objects, and wrong boxes from rollouts of the base corrector; about a quarter are random regions. Each overlaps both the annotated box and SAM~3's box by at most $10\%$ of their area, so that it does not reveal the answer.

\subsubsection{Which training examples are used}
\paragraph{Sources.}
The dialogues come from SA-V training shards \texttt{sav\_000}--\texttt{sav\_054}. They include 2,250 query frames from shards \texttt{sav\_040}--\texttt{sav\_054}, which stage~1 does not use, and 1,888 look-alike examples built as in stage~1 from separately mined pairs (306 keep, 271 switch, 583 synthetic missed target, 728 absent target).

\paragraph{Object size.}
Crops help most for small objects, so all three sizes are well represented (\cref{tab:zoom_training_cells}). Small means an annotated box below $0.5\%$ of the frame, medium $0.5$--$2\%$, and large above $2\%$.

\subsubsection{Training}
\label{sec:zoom_optimization}
Training continues from the stage-1 model with the same settings (\cref{tab:supp_train}).
The model should learn to answer after uninformative crops without learning to request them. We therefore use the standard next-token loss with a weight per model turn:
\begin{equation}
 \mathcal{L}_{\text{crop}} = -\sum_{t} w_t \log p_\theta\!\left(a_t \mid x, a_{<t}, c_{<t}\right),
 \label{eq:crop_sft}
\end{equation}
where $a_t$ is the model's $t$-th turn (a crop request $b_k$ or the final answer $y$) and $c_{<t}$ are the crops returned before it.
A request for an L, R or O crop has $w_t=0$: it stays in the context, together with the crop it returns, but the model is not trained to produce it. All other turns, i.e.\ P and T requests and the final answer, have $w_t=1$.
In total, 6,420 of the 10,907 crop requests are trained and 4,487 are not; in 3,490 dialogues every turn is trained.

\subsection{Stage 3: Crop policy refinement}
\label{sec:data_refinement}

Stage~2 imitates scripted crop placements, which need not be the regions most informative to the model. Stage~3 therefore optimizes the full dialogue with reinforcement learning (GRPO~\cite{shao2024deepseekmath}). For each query, the corrector samples complete dialogues, i.e.\ its own crop requests and its final answer, and each dialogue is rewarded by the accuracy of its final answer (\cref{eq:stage3_reward}). Crop placement thus receives credit only through the answers it enables, without supervision toward scripted boxes. A separate cross-entropy term on the verdict keeps the verdict confidence calibrated (\cref{sec:stage3_objective}).

\subsubsection{Queries and reward}
A training query is the input $x$ of a crop dialogue (\cref{eq:crop_dialogue}). Given $x$, the corrector generates the complete dialogue $D=(x,b_1,c_1,\dots,b_K,c_K,y)$; the crops $c_k$ are returned by the environment exactly as at test time (\cref{sec:supp_inference}).
Each query carries three labels:
\begin{itemize}
 \item $v^\star\in\{0,1\}$: whether the target is visible in the query frame;
 \item $b^\star$: the target's annotated box, when it is visible;
 \item $d^\star$: the correct verdict, \texttt{correct} in the two keep situations and \texttt{wrong} otherwise.
\end{itemize}
The reward compares the visibility $v$ and the box $b$ of the final answer $y$ with these labels (\cref{eq:zoom_reward}):
\begin{equation}
 R(D)=\begin{cases}
 \operatorname{IoU}(b,b^\star), & v=v^\star=1,\\
 1, & v=v^\star=0,\\
 0, & \text{otherwise.}
 \end{cases}
 \label{eq:stage3_reward}
\end{equation}
An invalid dialogue (malformed output or a repeated crop) receives $R=0$.
The reward scores only the final answer: crops are not rewarded directly but learned through the answers they lead to. It also ignores the verdict, which is supervised separately with $d^\star$ (\cref{sec:stage3_objective}).
The training set contains 2,375 queries from SA-V training videos, none of whose query frames are used in stage~1 or~2.

\subsubsection{Query selection}
\label{sec:stage3_choice}
GRPO learns only from differences between the eight dialogues it samples for a query: if all eight receive the same reward, every advantage is zero and the query contributes no policy gradient.
We therefore keep a query only if eight rollouts of the base corrector (the stage-1 model) on it received different rewards. For a spurious mask, where every erase scores 1 and every other answer scores 0, this means at least one erase and at least one other answer.
Among such queries, the training set is composed as follows:
\begin{itemize}
 \item \emph{all five situations}, with 450--500 queries each (\cref{tab:rl_cells}), so that no single action dominates the reward;
 \item \emph{object size}: all three sizes where the object is visible (\cref{tab:rl_cells});
 \item \emph{look-alikes}: at least $30\%$ of each situation, 807 in total;
 \item \emph{detector calls}: 1,546 queries ($65.1\%$) are frames on which our failure detector fires;
 \item \emph{variety}: at most three queries per object and six per video.
\end{itemize}

\subsubsection{Training objective}
\label{sec:stage3_objective}
For each query, $G=8$ dialogues $D_1,\dots,D_G$ are sampled from the current corrector $\pi_\theta$ at temperature~1, and each receives the group-centered advantage
\begin{equation}
 A_i = R(D_i)-\frac{1}{G}\sum_{j=1}^{G}R(D_j).
 \label{eq:stage3_adv}
\end{equation}
A dialogue with an above-average reward has $A_i>0$ and is made more likely; one with a below-average reward has $A_i<0$ and is made less likely. Unlike the original GRPO~\cite{shao2024deepseekmath}, we do not divide by the standard deviation of the group's rewards~\cite{liu2025understanding}: in a group with nearly equal rewards, this division would turn small reward differences into large advantages.

The model is trained with
\begin{equation}
 \mathcal{L}=\mathcal{L}_{\mathrm{PG}}^{\neg d}+\beta\,\mathcal{L}_{\mathrm{KL}}^{\neg d}+\lambda_{\mathrm v}\,\mathcal{L}_{\mathrm{CE}}(d^\star),
 \label{eq:stage3_loss}
\end{equation}
which is \cref{eq:zoom_refinement} of the main paper. Let $T_i$ be the tokens that the corrector generates in dialogue $D_i$, i.e.\ its crop requests and its final answer, without the verdict token, and let $h_{i,t}$ be the dialogue before token $y_{i,t}$. Crops and prompt turns are not generated by the corrector and carry no loss. The three terms are
\begin{align}
 \mathcal{L}_{\mathrm{PG}}^{\neg d}&=-\frac{1}{G}\sum_{i=1}^{G}\frac{A_i}{|T_i|}\sum_{t\in T_i}\log\pi_\theta(y_{i,t}\mid h_{i,t}),\label{eq:stage3_pg}\\
 \mathcal{L}_{\mathrm{KL}}^{\neg d}&=\frac{1}{G}\sum_{i=1}^{G}\frac{1}{|T_i|}\sum_{t\in T_i}\bigl(e^{r_{i,t}}-r_{i,t}-1\bigr),\label{eq:stage3_kl}\\
 \mathcal{L}_{\mathrm{CE}}(d^\star)&=-\frac{1}{G}\sum_{i=1}^{G}\log\pi_\theta(d^\star\mid h_{i,\mathrm v}),\label{eq:stage3_ce}
\end{align}
averaged over the queries of a training step.
\begin{itemize}
 \item $\mathcal{L}_{\mathrm{PG}}^{\neg d}$ is the policy-gradient term. Each batch of dialogues is used for exactly one gradient step, so the policy that sampled the dialogues is the policy being updated; the importance ratio of the clipped GRPO objective is then~1, and the objective reduces to the advantage-weighted log-likelihood of \cref{eq:stage3_pg}.
 \item $\mathcal{L}_{\mathrm{KL}}^{\neg d}$ keeps the corrector close to the stage-2 model $\pi_{\mathrm{ref}}$, so that it retains the protocol learned in stage~2. It is the per-token KL estimate of GRPO~\cite{shao2024deepseekmath}, with $r_{i,t}=\log\pi_{\mathrm{ref}}(y_{i,t}\mid h_{i,t})-\log\pi_\theta(y_{i,t}\mid h_{i,t})$, and $\beta=0.01$.
 \item $\mathcal{L}_{\mathrm{CE}}(d^\star)$ is the cross-entropy of the verdict token, where $h_{i,\mathrm v}$ is the dialogue before the verdict token of $D_i$; $\lambda_{\mathrm v}=0.02$.
\end{itemize}
\paragraph{Why the verdict is supervised separately.} The reward does not depend on the verdict. If the verdict token were trained by the policy gradient, its probability would follow the reward of the box rather than the correctness of the verdict, and the verdict confidence would no longer indicate whether SAM~3 is wrong. The verdict token is therefore excluded from $\mathcal{L}_{\mathrm{PG}}$ and $\mathcal{L}_{\mathrm{KL}}$ ($\neg d$) and trained only by $\mathcal{L}_{\mathrm{CE}}$ toward $d^\star$. Because this term is computed inside each sampled dialogue, the verdict is learned conditioned on the crops the corrector itself chose, as at test time, where its confidence decides whether a correction is applied (\cref{sec:supp_inference}).

\subsubsection{Optimization details}
The stage-2 LoRA adapters are merged into the model, the vision encoder and the merger are frozen, and new LoRA adapters are trained on the language model (\cref{tab:supp_train}). The merged stage-2 model is the fixed reference of the KL term.
Each step uses 8 queries with 8 dialogues each. Rollouts are generated by a vLLM engine colocated with the trainer; a group whose eight rewards are all equal is resampled, at most three times.
We train for 130 steps.

\subsection{Shortcut checks}
\label{sec:shortcuts}
Each training example combines images with a text prompt. The corrector should learn to answer from the images; if the text alone predicts the label, it can instead learn to answer from the text. We measure how much text reveals with a text-only test, use it to design the prompt, and apply to the final corpus.

\paragraph{Text-only test.}
For each label, a classifier is trained on the prompt text alone and predicts the label. It is evaluated with 5-fold cross-validation grouped by video, so that no video appears in both training and test folds. An AUROC of $0.5$ means that the text carries no information about the label.

\paragraph{Prompt design.}
SAM~3's prediction can be given to the corrector in two ways: as text, by listing SAM~3's box on every earlier frame (a \emph{box trail}), or as images, by drawing SAM~3's mask as an outline on each frame. In an earlier version of our corpus that used a box trail, the text-only test predicted the verdict with AUROC $0.77$--$0.79$: the box coordinates alone were enough to tell many failures apart. Our prompt therefore contains no boxes, and SAM~3's prediction is shown only as outlines on the frames. The remaining text consists of fixed sentence templates: the number of images, the time of each frame relative to the query, and the number of later frames on which SAM~3 predicts the object as not visible.

\paragraph{Results on the final corpus} (\cref{tab:shortcuts}).
\begin{itemize}
 \item \emph{Verdict.} Without the box trail, the text-only test reaches AUROC $0.570$ on the final corpus, close to chance: whether SAM~3 is wrong must be read from the images.
 \item \emph{Visibility.} The text is more informative ($0.710$) because it reports on how many later frames SAM~3 predicts the object as not visible. This is part of SAM~3's output and is given to the corrector at test time in the same form.
 \item \emph{Identity contrasts.} A keep~$L$ example and its switch example have identical text ($0.500$), so the target can be identified only from the exemplar. Constructed missed and absent targets are separated with AUROC $0.594$, at the level of natural missed targets and correct absences ($0.616$), so the construction adds no text cue.
\end{itemize}

\begin{table}[!ht]
 \centering
 \caption{\textbf{Text-only shortcut test}: AUROC of a classifier on the prompt text alone ($0.5$: no information).}
 \label{tab:shortcuts}
 \small
 \setlength{\tabcolsep}{3pt}
 \begin{tabular}{@{}lrc@{}}
  \toprule
  Label predicted & Examples & AUROC \\
  \midrule
  Verdict (wrong vs correct) & 90,379 & $0.570$ \\
  Absent vs present & 90,379 & $0.710$ \\
  Switch vs keep~$L$ (contrasts) & 2,288 & $0.500$ \\
  Missed vs absent (contrasts) & 2,748 & $0.594$ \\
  \bottomrule
 \end{tabular}
\end{table}

\section{Implementation details}
\label{sec:method_details}

This section gives the details needed to reproduce the two components of the system: the failure detector, which decides when to call the corrector (\cref{sec:supp_detector}), and one call of the corrector, from its input to the correction of the tracker (\cref{sec:supp_inference}).

\subsection{Failure detector}
\label{sec:supp_detector}
\paragraph{Overview.}
The detector decides when to call the corrector. It is training-free: it reads quantities that SAM~3 computes at every processed frame and uses no annotation beyond the initial mask. It has two channels: the mask channel C1 operates on frames where SAM~3 predicts a mask, and the presence channel A2 on runs of frames where it predicts none; the detector that combines them is C1A2.

\subsubsection{Constants}
\Cref{tab:supp_detector_constants} lists all constants; they are the same on every benchmark and tracker.
A processed frame is a frame on which SAM~3 runs: every frame on SA-V (24\,fps) and 6 frames per second on LVOS and SeCVOS. All counts are in processed frames; only the presence poll $P$ is set in seconds (0.5\,s). Thus $W=24$ frames span 1\,s on SA-V and 4\,s on LVOS and SeCVOS.

\begin{table}[t]
\centering
\small
\caption{\textbf{Detector constants.}}
\label{tab:supp_detector_constants}
\setlength{\tabcolsep}{4pt}
\begin{tabular}{@{}llr@{}}
\toprule
Symbol & Role & Value \\
\midrule
$\theta_{\mathrm{open}}$ & episode opens if $s_{1,t}\ge\theta_{\mathrm{open}}$ & 0.06 \\
$\kappa$ & offset in the accumulator & 0.15 \\
$H$ & accumulated evidence to confirm & 3 \\
$\eta$ & area change $|\Delta\log|\widehat M_t||$ & 0.2 \\
$W$ & quiet frames before closing & 24 \\
-- & empty predictions that clear the state & 24 \\
$\tau_{\mathrm{call}}$ & call gate on $s_{1,t}$ (C1) & 0.249 \\
$\tau_{\mathrm{abs}}$ & objectness-logit gate (A2) & $-2.859$ \\
$P$ & presence poll, 0.5\,s & 12 / 3 \\
\bottomrule
\end{tabular}
\par\smallskip
{\footnotesize $P$: 12 frames on SA-V, 3 on LVOS and SeCVOS.\par}
\end{table}

\subsubsection{Signals}
The detector uses seven signals per frame, all from SAM~3.

\paragraph{Propagation-state evidence $\mathcal{V}_{1,t}$.} Four of them SAM~3 computes during propagation anyway, so they cost nothing:
\begin{itemize}\itemsep0pt
 \item \emph{Presence probability} $p_t^{\mathrm{obj}}=\sigma(o_t)$, from SAM~3's objectness logit $o_t$: does SAM~3 think object is in the frame?
 \item \emph{Mask confidence} $u_t^{\mathrm{eff}}$ which is SAM~3's own predicted IoU of its best mask.
 \item \emph{Memory confidence} $m_t$: how confident SAM~3 was on the recent frames it is relying on. Precisely, the mean of the confidences $u^{\mathrm{eff}}$ of the non-conditioning memory entries that SAM~3's memory selector uses for frame $t$.
 \item \emph{Identity continuity} $c_t$: does the masked region still look like the object did at the start? Precisely, the cosine similarity between the memory features average-pooled under the current mask and those pooled under the first non-empty mask of the object. It drops when the mask moves to a different-looking object.
\end{itemize}
\paragraph{Target-consistency evidence $\mathcal{V}_{2,t}$.} Three more checks whether the current mask is consistent with what SAM~3 knows about the target, each as a mask IoU (two empty masks count as agreeing):
\begin{itemize}
 \item \emph{Memory check} $r_t^{\mathrm{cond}}$: would SAM~3 draw the same mask if it trusted only the frames we know are right? Precisely, an extra decoder pass on the current frame that attends only to the conditioning memory; $r_t^{\mathrm{cond}}$ is its mask IoU with the current mask.
 \item \emph{Backward check} $r_t^{\mathrm{back}}$: if SAM~3 is still on the right object, its current memory should find the object again in the first frame. Precisely, an extra decoder pass on the \emph{initial} frame that uses the current frame's memory entry as its only memory; $r_t^{\mathrm{back}}$ is its mask IoU with the given initial mask.
 \item \emph{Hypothesis agreement} $r_t^{\mathrm{hyp}}$: do SAM~3's alternative masks agree? Precisely, the mean pairwise mask IoU of SAM~3's three candidate masks.
\end{itemize}

\paragraph{Scores.} Each group is summarized by its mean discrepancy, $s_{k,t}=\frac{1}{|\mathcal V_{k,t}|}\sum_{v\in\mathcal V_{k,t}}(1-v)$ for $k\in\{1,2\}$: $s_{1,t}$ is propagation-state score and $s_{2,t}$ target-consistency score.

\subsubsection{Mask channel (C1): when SAM~3 shows a mask}
C1 calls the corrector only when there is lasting evidence of a problem \emph{and} the mask visibly changes.
It is updated only on frames with a non-empty mask, in four steps:
\begin{enumerate}
 \item \emph{Open.} A suspicious stretch (an \emph{episode}) opens when the propagation-state score reaches $s_{1,t}\ge\theta_{\mathrm{open}}$.
 \item \emph{Accumulate.} While the episode is open, each non-empty frame adds the amount by which its strongest evidence exceeds a margin, $\delta_t[\max(s_{1,t},s_{2,t})-\kappa]_+$, to a running total $A_t$, where $\delta_t$ is the number of processed frames since the previous non-empty frame. A visible change of the mask, a log-area change $|\Delta\log|\widehat M_t||\ge\eta$ relative to the previous non-empty frame, sets a flag $J_t=1$.
 \item \emph{Confirm and call.} The episode is confirmed at the first frame where enough evidence has accumulated and the mask has changed ($A_t\ge H$ and $J_t=1$). The corrector is called only if the current frame is itself still suspicious ($s_{1,t}\ge\tau_{\mathrm{call}}$).
 \item \emph{Reset.} 
 Every confirmation, whether or not it led to a call, closes the episode and clears its state, so each confirmation yields at most one call; a new episode opens at the next non-empty frame with $s_{1,t}\ge\theta_{\mathrm{open}}$.
 An unconfirmed episode closes at the first non-empty frame with $s_{1,t}<\theta_{\mathrm{open}}$ that is at least $W$ frames after its last frame with $s_{1,t}\ge\theta_{\mathrm{open}}$. The whole state is cleared after 24 consecutive empty masks.
\end{enumerate}

\subsubsection{Presence channel (A2): when SAM~3 has no mask}
When SAM~3 predicts no mask, C1 receives no evidence. During a run of empty masks, A2 tests every $P$ frames (0.5\,s) whether SAM~3's objectness logit still indicates the object, and calls the corrector if $o_t\ge\tau_{\mathrm{abs}}$.
The run ends only at a non-empty mask or an applied correction, so a long empty run is checked repeatedly.

\subsubsection{Constant selection}
All constants are set on SA-V validation and then frozen.
$\theta_{\mathrm{open}}$, $H$, $\eta$ and $\kappa$ come from a grid search on SA-V validation videos over $\theta_{\mathrm{open}}\in\{0.06,0.07,0.08,0.09,0.10,0.12\}$, $H\in\{3,4,5\}$, $\eta\in\{0.2,0.3,0.4\}$ and $\kappa\in\{0.10,0.15,0.20\}$, maximizing the recall of failure episodes.
The two thresholds are quantiles on SA-V validation: $\tau_{\mathrm{call}}$ is the value of $s_{1,t}$ above which $60\%$ of the confirmations lie, and $\tau_{\mathrm{abs}}$ the value of $o_t$ above which $40\%$ of the presence checks lie.

\subsection{Corrector inference}
\label{sec:supp_inference}
This section describes a single corrector call: its input, its output, the crop protocol, and how it updates the tracker.

\subsubsection{Input}
Each call shows the corrector at most 22 images in time order, followed by the crops it requests:
\begin{itemize}\itemsep0pt
 \item the \emph{exemplar}: the first frame with the given initial mask, plus a zoomed view of the object;
 \item up to 16 \emph{earlier frames}: $U=4$ frames sampled uniformly over the earlier video, which show the object's earlier appearance, and $R=12$ recent frames, which show how a failure developed: up to three in the dense window $[t-\Delta,t)$ with $\Delta=0.5$\,s, the rest at intervals of about 1\,s;
 \item the full \emph{query frame};
 \item three \emph{later frames} in the dense window $(t,t+\Delta]$, about $1/6$, $1/3$ and $1/2$\,s after the query, produced by a separate copy of the tracker that does not affect the real one. They show whether a failure persists.
\end{itemize}
Every image shows SAM~3's predicted mask as an outline, not filled, so the object itself stays visible.
Each frame is resized to a long side of 1,000 pixels and fed as a separate image with a budget of 602,112 pixels per image.
Each frame is labeled with a time in seconds starting at a random offset. The text of the prompt lists the images in order with their times, describes the role of each image, states the crop protocol and the answer format, and contains  no box trails.

\subsubsection{Output}
The final answer is one JSON object inside an \texttt{<answer>} tag: the verdict (\texttt{correct} or \texttt{wrong}), whether the object is visible (\texttt{visible}), its box (\texttt{box\_2d}, coordinates in $[0,1000]$, zero when not visible), and a right/wrong flag for each earlier frame (\texttt{ctx}).
\texttt{correct} maps to \textsc{keep}; \texttt{wrong} with a visible object maps to \textsc{reseed} with the box; \texttt{wrong} with no visible object maps to \textsc{erase}.

\subsubsection{Crops}
Crops let the corrector inspect small or ambiguous objects at source resolution.
\begin{itemize}
 \item \emph{Request.} Before answering, the corrector must ask for one crop of the query frame and may ask for a second, written \texttt{<zoom>[x1,y1,x2,y2]</zoom>}.
 \item \emph{Returned image.} Exactly that region, cut from the source-resolution query frame with SAM~3's outline, and resized so that its long side is $\max(640,\min(L,1000))$ pixels, where $L$ is its source long side, with at most $6\times$ magnification. The returned turn also states the region in full-frame coordinates.
 \item \emph{Invalid requests.} A region whose shorter side is at most 30 pixels in the 1,000-pixel frame is refused, and a second region must have IoU at most $0.9$ with the first.
\end{itemize}
\paragraph{Skipped calls.} If SAM~3's mask is below the model's visual resolution, the call is skipped, since such a mask cannot be verified; precisely, if the shorter side of SAM~3's predicted box is below one 32-pixel visual token after the query frame is resized for the model (long side 1,000 pixels, then the pixel budget). Calls with an empty mask are never skipped.

\subsubsection{Commitment}
Decoding is greedy (bf16, no thinking mode).
A correction is committed only if the corrector is confident that SAM~3 is wrong: the verdict confidence $\ell_t$ is the mean log-probability of the tokens of the verdict value (\texttt{wrong}) in the final turn, and a correction requires $p=\exp(\ell_t)\ge\tau=0.9$.
\textsc{reseed} additionally requires a valid box with IoU below $0.5$ to SAM~3's current predicted box (taken as 0 if the mask is empty), since a closer box would change the track little; \textsc{erase} requires a non-empty current mask.
Each call makes at most one correction.

\subsubsection{Tracker update}
\textsc{reseed} passes the box to SAM~3 as a box prompt; \textsc{erase} passes an empty mask.
The prompted frame becomes a conditioning frame of SAM~3, like the initial frame; the cached prediction for that frame is discarded, all earlier memory is kept, and propagation resumes from that frame.
Because of the later frames, the output for a frame is final 0.5\,s later: the system runs online with a fixed 0.5\,s delay.

\section{Failure-detection analysis}
\label{sec:trigger_analysis}

This section evaluates the failure detector C1A2 (\cref{sec:supp_detector}), which decides when to call the corrector. It has two channels. While SAM~3 shows a mask, the mask channel C1 accumulates evidence that the mask is wrong and calls the corrector once the evidence is strong and the mask visibly changes. While SAM~3 shows no mask, the presence channel A2 checks regularly whether SAM~3 still sees a sign of the object. The detector's constants were fitted once on SA-V val and are kept unchanged for every result below.

We evaluate the detector by three criteria: whether its calls coincide with failures (precision), how many failures it reaches (recall), and how early it reaches them. Beyond the placement of calls, we analyze how early the detector fires (\cref{sec:trigger_timing}), how it compares with simpler detectors (\cref{sec:trigger_baselines}), what each component contributes (\cref{sec:trigger_ablation}), how sensitive it is to its constants (\cref{sec:trigger_sensitivity}), whether it transfers to other trackers without refitting (\cref{sec:tracker_transfer}), and how its calls perform in closed loop (\cref{sec:oracle_trigger_details}).

\subsection{Evaluation protocol}
\label{sec:measure}
Except in \cref{sec:oracle_trigger_details}, we replay SAM~3's stored predictions without any correction, so that every detector is scored on exactly the same predictions, and compare the detector's calls with the annotation.

\paragraph{Failures.}
Let $M_t$ be the annotated mask and $\widehat M_t$ SAM~3's mask on an annotated frame $t$. SAM~3 \emph{fails} at $t$ if the object is present and $\mathrm{IoU}(\widehat M_t,M_t)<0.5$, or if the object is absent and $\widehat M_t\neq\emptyset$.
A \emph{failure episode} is a maximal run of consecutive failing annotated frames. A \emph{persistent failure} is an episode lasting at least 1\,s; we focus on these because shorter episodes often resolve by themselves, while persistent ones are the failures a correction must fix.

\paragraph{Scores.}
Since only annotated frames can be checked, a call is scored at the first annotated frame at or after it, i.e.\ the frame that a correction would change.
\begin{itemize}\itemsep0pt
\item \emph{Precision}: fraction of calls made while SAM~3 is failing.
\item \emph{Recall}: the fraction of persistent failures that receive at least one call.
\item \emph{Recall within 1\,s}: the fraction of persistent failures that receive a call within 1\,s of their start.
\item \emph{Delay}: the time from the start of a persistent failure to its first call.
\end{itemize}

\paragraph{Comparison with random calls.}
A detector that calls more often finds more failures, even if its calls are placed no better than chance. We therefore compare every detector with random calls: for each object, the same number of calls placed at random frames, averaged over 1,000 random draws. A detector places its calls well if it beats random calls of the same number.

\paragraph{Dual-Agent trigger.}
For comparison we replay the trigger of Dual-Agent~\cite{bundele2026trackers} as released: an occlusion check every 12 frames, a reappearance check, and a rule on the variance of SAM~3's predicted IoU.

\subsection{Detection timing}
\label{sec:trigger_timing}
\emph{C1A2 usually calls the corrector within a fraction of a second of failure onset.}

A call early in a failure is worth more than a late one: the longer SAM~3 tracks the wrong object, the more of the wrong appearance it stores in its memory, and the more frames are already lost. \cref{tab:trigger_timing} reports the recall within 1\,s of the failure's start and the delay of the first call. The main paper shows the full curves.

\begin{table}[t]
\centering
\caption{\textbf{Detection timing} on persistent failures: recall within 1\,s of onset, for the detector and for as many random calls, and quantiles of the delay from onset to the first call (a negative delay means that the call came before the first failing annotated frame).}
\label{tab:trigger_timing}
\small
\setlength{\tabcolsep}{3pt}
\resizebox{\linewidth}{!}{%
\begin{tabular}{@{}ll cc ccc@{}}
\toprule
& & \multicolumn{2}{c}{Recall within 1\,s (\%)} & \multicolumn{3}{c}{Delay quantile (s)} \\
\cmidrule(lr){3-4}\cmidrule(l){5-7}
Benchmark & Detector & detector & random & 25\% & 50\% & 75\% \\
\midrule
SA-V test & C1A2       & 84.5 & 57.2 & $-$0.04 & 0.04 & 0.21 \\
          & Dual-Agent & 73.6 & 60.6 & 0.00 & 0.13 & 0.29 \\
\midrule
LVOS v2   & C1A2       & 67.5 & 33.7 & 0.00 & 0.33 & 0.67 \\
          & Dual-Agent & 50.3 & 23.2 & 0.33 & 0.67 & 1.17 \\
\midrule
SeCVOS    & C1A2       & 65.0 & 44.1 & 0.00 & 0.33 & 0.67 \\
          & Dual-Agent & 55.0 & 33.1 & 0.33 & 0.33 & 1.33 \\
\bottomrule
\end{tabular}}
\end{table}

\paragraph{Findings.} On every benchmark, C1A2 calls the corrector within 1\,s of the start of most persistent failures, 21--34 points more than random calls of the same number. On SA-V and LVOS, its recall within 1\,s even exceeds the recall that random calls reach over the whole failure (\cref{tab:trigger_transfer}). The median delay is one processed frame on SA-V and two on LVOS and SeCVOS, so the corrector usually sees a failure before SAM~3 has stored much of the wrong object in its memory.
On SeCVOS, C1A2 and the Dual-Agent trigger reach about the same recall over the whole failure, but C1A2 reaches it earlier: it finds more failures within 1\,s, and at the 75th percentile its first call comes at half the delay.

\subsection{Comparison with simpler detectors}
\label{sec:trigger_baselines}
\emph{At the same number of calls, C1 finds more failures than any simpler rule, and C1A2 is more precise than periodic calls.}

\begin{figure*}[t]
\centering
\includegraphics[width=\linewidth]{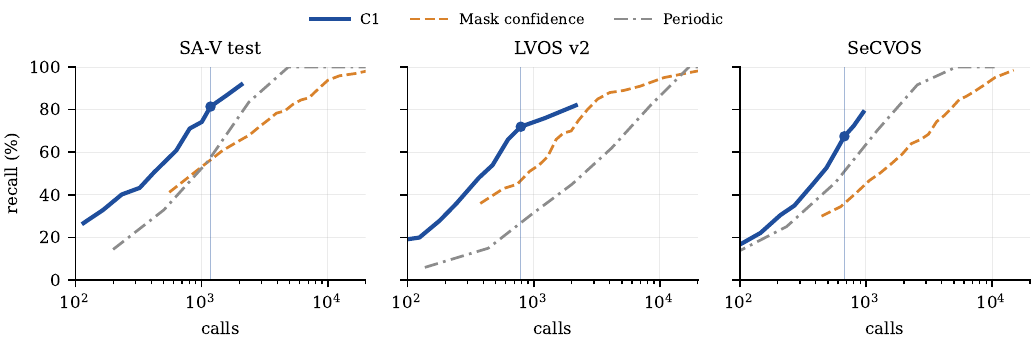}
\caption{\textbf{Recall against number of calls} for persistent failures with a non-empty mask. Each curve sweeps one rule's threshold or period; dot and vertical line: C1 as deployed.}
\label{fig:trigger_baselines}
\end{figure*}

We compare C1A2 with two simpler rules that read the same SAM~3 outputs. Each rule has one setting (a threshold or a period) that controls how often it calls. We vary this setting over its whole range (\cref{fig:trigger_baselines}) and also read it at exactly the number of calls that C1A2 makes on each benchmark, interpolating linearly between the two nearest settings (\cref{tab:trigger_baselines}).
\begin{itemize}\itemsep0pt
\item \emph{Periodic}: a call every $k$ frames, with or without a mask.
\item \emph{Mask confidence}: a call on every frame with $u_t^{\mathrm{eff}}$ below a threshold.
\end{itemize}
Mask confidence, like C1, can only act on frames where SAM~3 shows a mask, so we compare the two on the persistent failures that have a mask (\cref{fig:trigger_baselines}).

\paragraph{C1 finds the most failures.}
Over the whole range of calls, C1's curve lies above every simpler rule on all three benchmarks (\cref{fig:trigger_baselines}), and the best of them needs 1.7 to 2.7 times as many calls to reach C1's recall. Two properties of C1 explain this. First, C1 reads more evidence than SAM~3's confidence: its target-consistency checks catch failures in which SAM~3 is confident but follows the wrong object. Even its propagation-state score $s_1$, which includes mask confidence, separates failing from correct masks better than mask confidence alone on every benchmark and tracker (\cref{tab:signal_auroc_transfer}). Second, C1 accumulates evidence over frames, so a failure whose signals are only moderately raised, but stay raised, is still confirmed.

\paragraph{C1A2 against periodic calls.}
At the same number of calls, C1A2 is more precise than periodic calls on all three benchmarks, several times more on SA-V and LVOS, where it also finds more persistent failures (\cref{tab:trigger_baselines}). Periodic calls do not look at SAM~3's state: they land on failing and correct frames alike, so most of them land on frames where SAM~3 is correct. C1A2 calls only when there is evidence of a failure: C1 when evidence has accumulated over an episode and the mask has visibly changed, and A2 when SAM~3 shows no mask although its objectness indicates that the object is still in view. Precise calls matter in closed loop, because every call on a correct frame gives the corrector a chance to damage a correct track.

\begin{table}[t]
\centering
\caption{\textbf{Periodic calls vs.\ C1A2} at the number of calls of C1A2: precision and recall (\%) of persistent failures. Bold: best.}
\label{tab:trigger_baselines}
\small
\setlength{\tabcolsep}{3pt}
\resizebox{0.7\linewidth}{!}{%
\begin{tabular}{@{}l rr rr rr@{}}
\toprule
 & \multicolumn{2}{c}{SA-V test} & \multicolumn{2}{c}{LVOS v2} & \multicolumn{2}{c}{SeCVOS} \\
\cmidrule(lr){2-3}\cmidrule(lr){4-5}\cmidrule(l){6-7}
Detector & P & R & P & R & P & R \\
\midrule
Periodic & 13 & 74 &  9 & 48 & 25 & \textbf{91} \\
C1A2     & \textbf{44} & \textbf{86} & \textbf{58} & \textbf{79} & \textbf{34} & 76 \\
\bottomrule
\end{tabular}}
\end{table}

\subsection{Component ablation}
\label{sec:trigger_ablation}
\emph{Both evidence groups help, but in different ways: target-consistency evidence $s_2$ mainly adds coverage, propagation-state evidence $s_1$ makes the calls precise, and the presence channel finds failures that the mask channel cannot see.}

C1 combines two groups of evidence (\cref{sec:supp_detector}): \emph{propagation-state} evidence $s_1$, built from SAM~3's own confidence in its current prediction, and \emph{target-consistency} evidence $s_2$, built from extra checks of whether the current mask agrees with what SAM~3 knows about the target. It also requires a visible change of the mask area before it calls. We remove one of these parts, or the presence channel A2, refit the thresholds of the rest on SA-V val as for the full detector, letting the number of calls change, and report the change in the number of calls, in recall and in precision, pooled over SA-V test, LVOS and SeCVOS (\cref{tab:signal_ablation_full}).

\begin{table}[t]
\centering
\caption{\textbf{Component ablation}, pooled over the three benchmarks (refitted on SA-V val).}
\label{tab:signal_ablation_full}
\small
\setlength{\tabcolsep}{4pt}
\begin{tabular}{@{}l rrr@{}}
\toprule
Removed & $\Delta$Calls (\%) & $\Delta$R & $\Delta$P \\
\midrule
Target consistency $s_2$ & $-20.6$ & $-13.97$ & $-1.09$ \\
Propagation state $s_1$  & $-0.4$  & $-7.76$  & $-3.44$ \\
Area change              & $+39.8$ & $+7.07$  & $-2.61$ \\
Presence channel A2      & $-61.2$ & $-24.48$ & $+9.87$ \\
\bottomrule
\end{tabular}
\end{table}

\paragraph{Findings.}
Target-consistency evidence $s_2$ mainly adds coverage: without it, the detector calls less often and finds many fewer persistent failures. Propagation-state evidence $s_1$ makes the calls precise: without it, the detector makes as many calls, but fewer of them land on failures and it finds fewer persistent failures. This matches its role as the final check before a call, which requires the current frame itself to look suspicious ($s_{1,t}\ge\tau_{\mathrm{call}}$). The area-change rule saves calls and keeps the detector precise: without it, the extra failures the detector finds cost 39.8\% more calls at lower precision. The presence channel A2 covers the failures where SAM~3 has lost the object and shows no mask, which C1 cannot measure.

\subsection{Sensitivity to the constants}
\label{sec:trigger_sensitivity}
\emph{C1A2's results do not depend on the exact values of $\theta_{\mathrm{open}}$, $\kappa$ and $W$; $\eta$ balances precision and recall.}

The episode constants control when an episode opens ($\theta_{\mathrm{open}}$), how much evidence a frame adds ($\kappa$), when an unconfirmed episode closes ($W$), and how large a mask change must be to count ($\eta$). We vary one of them at a time and keep the others at their values in \cref{tab:supp_detector_constants}. Each variant is refitted on SA-V val like the deployed detector, so that it makes the same number of confirmations (\cref{tab:trigger_sensitivity}); unlike the component ablation (\cref{sec:trigger_ablation}), where the number of calls is free, the variants therefore differ only in which episodes they confirm.

Over a wide range of $\theta_{\mathrm{open}}$, $\kappa$ and $W$, precision and recall stay close to those of the deployed detector on all three benchmarks. The area-change constant $\eta$ sets how much the mask area must change before an episode can be confirmed. Requiring a change makes the calls more precise: compared with the rule switched off ($\eta=0$), the deployed value $\eta=0.2$ gains 6 points of precision on SA-V and LVOS. Among all tested values, $\eta=0.2$ gives the highest precision on SA-V and the highest recall on LVOS and SeCVOS, and no other value matches it in both precision and recall on all three benchmarks. The value fitted on SA-V val is therefore also the best balance on the test sets.

\begin{table}[t]
\centering
\caption{\textbf{Episode constants} varied one at a time: range of precision and recall (\%).}
\label{tab:trigger_sensitivity}
\small
\setlength{\tabcolsep}{3pt}
\resizebox{\linewidth}{!}{%
\begin{tabular}{@{}l rr rr rr@{}}
\toprule
 & \multicolumn{2}{c}{SA-V test} & \multicolumn{2}{c}{LVOS v2} & \multicolumn{2}{c}{SeCVOS} \\
\cmidrule(lr){2-3}\cmidrule(lr){4-5}\cmidrule(l){6-7}
Setting & prec. & rec. & prec. & rec. & prec. & rec. \\
\midrule
Deployed & 44 & 86 & 58 & 79 & 34 & 76 \\
\midrule
$\theta_{\mathrm{open}}\in[0.03,0.12]$ & 43--44 & 84--88 & 57--59 & 76--80 & 34--35 & 74--76 \\
$\kappa\in[0.05,0.30]$ & 41--45 & 84--88 & 55--60 & 77--80 & 34--35 & 70--78 \\
$W\in[6,72]$ & 43--44 & 84--87 & 57--58 & 76--79 & 34 & 74--77 \\
$\eta=0$ (rule off) & 38 & 90 & 52 & 78 & 34 & 75 \\
$\eta\in[0.1,0.3]$ & 41--44 & 83--90 & 55--60 & 78--79 & 34--35 & 74--76 \\
$\eta\in[0.4,0.5]$ & 37--39 & 74--81 & 59 & 70--75 & 35 & 72--73 \\
\bottomrule
\end{tabular}}
\end{table}

\subsection{Transfer to SAM~2.1 and SeC}
\label{sec:tracker_transfer}
\emph{Without refitting, C1A2 beats random calls on two other trackers, although the most informative single signal changes from tracker to tracker.}

\begin{figure}[t]
\centering
\includegraphics[width=\linewidth]{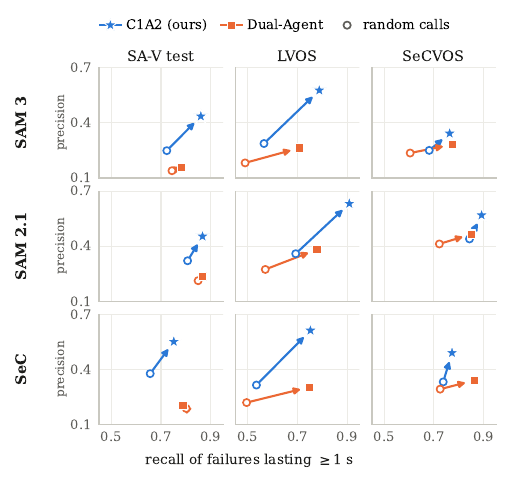}
\caption{\textbf{Failure localization and tracker transfer.} Precision against recall of persistent failures. Arrows connect each detector's random baseline with the same number of calls (circle) to its precision and recall. C1A2 uses unchanged parameters.}
\label{fig:trigger_transfer_app}
\end{figure}

C1A2 reads only signals that every SAM~2-family tracker computes. We therefore apply it, with SAM~3's constants and thresholds unchanged, to the predictions of SAM~2.1 (Hiera-L) and SeC on the same three test sets (\cref{tab:trigger_transfer,fig:trigger_transfer_app}). SeC uses SAM~2.1's prediction heads with a retrained memory.

\begin{table*}[t]
\centering
\caption{\textbf{Transfer to other trackers} with SAM~3's constants: precision and recall (\%). Gray: random calls of the same number. Bold: C1A2 above random.}
\label{tab:trigger_transfer}
\small
\setlength{\tabcolsep}{5pt}
\begin{tabular}{@{}l cc cc cc@{}}
\toprule
& \multicolumn{2}{c}{SA-V test} & \multicolumn{2}{c}{LVOS v2} & \multicolumn{2}{c}{SeCVOS} \\
\cmidrule(lr){2-3}\cmidrule(lr){4-5}\cmidrule(l){6-7}
Tracker & Precision & Recall & Precision & Recall & Precision & Recall \\
\midrule
SAM~3     & \textbf{44} \textcolor{gray}{(25)} & \textbf{86} \textcolor{gray}{(72)} & \textbf{58} \textcolor{gray}{(29)} & \textbf{79} \textcolor{gray}{(57)} & \textbf{34} \textcolor{gray}{(25)} & \textbf{76} \textcolor{gray}{(68)} \\
SAM~2.1   & \textbf{45} \textcolor{gray}{(32)} & \textbf{87} \textcolor{gray}{(81)} & \textbf{63} \textcolor{gray}{(36)} & \textbf{91} \textcolor{gray}{(69)} & \textbf{57} \textcolor{gray}{(44)} & \textbf{89} \textcolor{gray}{(84)} \\
SeC       & \textbf{55} \textcolor{gray}{(38)} & \textbf{75} \textcolor{gray}{(66)} & \textbf{61} \textcolor{gray}{(32)} & \textbf{75} \textcolor{gray}{(53)} & \textbf{49} \textcolor{gray}{(33)} & \textbf{77} \textcolor{gray}{(74)} \\
\bottomrule
\end{tabular}
\end{table*}

\paragraph{Findings.}
On every tracker and benchmark, C1A2 has higher precision and recall than random calls of the same number, with the largest margins on LVOS. Its precision also lies above the 97.5th percentile of the 1,000 random draws in every case.

\paragraph{Signals that carry the transfer.}
\Cref{tab:signal_auroc_transfer} shows how well each signal alone separates failing from correct predictions (failure AUROC), for every tracker, on annotated frames with a non-empty prediction. The strongest single signal differs between trackers. For SAM~3 and SAM~2.1, the propagation-state signals are most informative, led by the presence probability on SA-V and LVOS. 

For SeC, whose memory is retrained, the consistency checks work better, and the memory check is the strongest signal on SA-V and LVOS. The scores $s_1$ and $s_2$, which combine the signals of each group, outperform most single signals, and on SeC the target-consistency score $s_2$ is clearly stronger than $s_1$ on SA-V and LVOS. Their maximum, which C1 accumulates, outperforms every single signal in seven of the nine columns. Because C1A2 combines several signals and both evidence groups, it does not depend on any single tracker-specific signal, which is why it transfers well.

\begin{table*}[t]
\centering
\caption{\textbf{Failure AUROC of each signal.} Bold: best per column; underlined: best single signal; $s_1$, $s_2$: evidence scores; $\max(s_1,s_2)$: evidence accumulated by C1.}
\label{tab:signal_auroc_transfer}
\small
\setlength{\tabcolsep}{4pt}
\begin{tabular}{@{}l ccc ccc ccc@{}}
\toprule
& \multicolumn{3}{c}{SAM~3} & \multicolumn{3}{c}{SAM~2.1} & \multicolumn{3}{c}{SeC} \\
\cmidrule(lr){2-4}\cmidrule(lr){5-7}\cmidrule(l){8-10}
Signal & SA-V & LVOS & SeCVOS & SA-V & LVOS & SeCVOS & SA-V & LVOS & SeCVOS \\
\midrule
\multicolumn{10}{@{}l}{\emph{Propagation-state evidence $\mathcal V_1$}} \\
\quad Mask confidence $u_t^{\mathrm{eff}}$          & 0.835 & 0.851 & 0.763 & 0.833 & 0.776 & 0.746 & 0.752 & 0.714 & 0.746 \\
\quad Presence probability $p_t^{\mathrm{obj}}$ & \underline{0.841} & \underline{0.856} & 0.762 & \underline{0.837} & \textbf{\underline{0.853}} & 0.749 & 0.794 & 0.756 & 0.714 \\
\quad Identity continuity $c_t$                & 0.826 & 0.828 & \underline{0.780} & 0.802 & 0.793 & \underline{0.771} & 0.816 & 0.747 & 0.731 \\
\quad Memory confidence $m_t$                  & 0.826 & 0.833 & 0.728 & 0.823 & 0.768 & 0.714 & 0.745 & 0.710 & 0.713 \\
\addlinespace
\multicolumn{10}{@{}l}{\emph{Target-consistency evidence $\mathcal V_2$}} \\
\quad Memory check $r_t^{\mathrm{cond}}$ & 0.824 & 0.816 & 0.714 & 0.816 & 0.757 & 0.734 & \underline{0.844} & \underline{0.780} & 0.661 \\
\quad Backward check $r_t^{\mathrm{back}}$ & 0.764 & 0.740 & 0.665 & 0.778 & 0.726 & 0.707 & 0.802 & 0.737 & 0.701 \\
\quad Hypothesis agreement $r_t^{\mathrm{hyp}}$ & 0.820 & 0.832 & 0.753 & 0.829 & 0.838 & 0.763 & 0.748 & 0.754 & \underline{0.746} \\
\midrule
$s_1$                                           & 0.858 & \textbf{0.873} & 0.783 & 0.852 & 0.820 & 0.766 & 0.817 & 0.764 & \textbf{0.750} \\
$s_2$                                           & 0.863 & 0.860 & 0.790 & 0.858 & 0.839 & 0.787 & 0.875 & 0.830 & 0.741 \\
$\max(s_1,s_2)$                                 & \textbf{0.868} & 0.868 & \textbf{0.799} & \textbf{0.860} & 0.844 & \textbf{0.788} & \textbf{0.878} & \textbf{0.833} & 0.745 \\
\bottomrule
\end{tabular}
\end{table*}

\subsection{Closed-loop evaluation with oracle corrector}
\label{sec:oracle_trigger_details}
\emph{When corrections change the tracker's memory, C1A2 still sends most of its updates to failing predictions and corrects more long failures than random calls.}

\paragraph{Setup.}
So far, SAM~3's predictions were replayed without corrections. In the real system, each correction changes SAM~3's memory and therefore all later predictions, including the signals the detector reads. To test the detector in this closed loop independently of the corrector's quality, we answer every call with an oracle corrector that has access to the annotation and feed its correction back into SAM~3: if the object is present, it writes the annotated mask (a \emph{mask write}, also when SAM~3 was already right); if the object is absent and SAM~3 shows a mask, it erases it; otherwise nothing happens (a \emph{no-op}). A mask write or an erasure is an \emph{update}.
We compare C1A2 with random calls (the same number per object) under the same oracle (\cref{tab:oracle_closed_loop}).

\begin{table}[t]
\centering
\caption{\textbf{Closed loop with an oracle corrector}: updates on a failing prediction, precision (\%).}
\label{tab:oracle_closed_loop}
\small
\setlength{\tabcolsep}{4pt}
\begin{tabular}{@{}l ccc@{}}
\toprule
Calls from & SA-V test & LVOS v2 & SeCVOS \\
\midrule
C1A2       & \textbf{50} & \textbf{53} & \textbf{60} \\
Random     & 22 & 12 & 16 \\
\bottomrule
\end{tabular}
\end{table}

\begin{figure*}[t]
\centering
\includegraphics[width=0.85\linewidth]{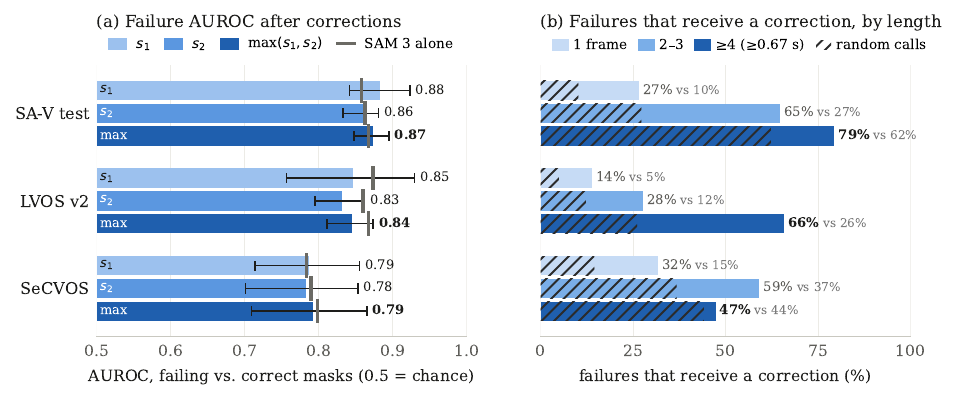}
\caption{\textbf{Diagnosis on corrected trajectories} (C1A2 with the oracle). Left: failure AUROC of $s_1$, $s_2$ and $\max(s_1,s_2)$ on the corrected run (95\% intervals); gray tick: SAM~3 alone. Right: fraction of failures that receive a correction, by length; hatched: random calls.}
\label{fig:closed_loop_state}
\end{figure*}

\paragraph{Findings.}
On every benchmark, 50--60\% of C1A2's updates are made on a failing prediction, more than twice the fraction for random calls (\cref{tab:oracle_closed_loop}). C1A2 also corrects more failures than random calls at every length, and longer failures more often (\cref{fig:closed_loop_state}, right).

The evidence scores keep working after corrections: on the corrected runs, $s_1$, $s_2$ and their maximum, which C1 accumulates, separate failing from correct masks about as well as on SAM~3 alone (\cref{fig:closed_loop_state}, left). With the frames on which a correction is applied left out, the AUROC is 0.78--0.86 for $s_1$, 0.78--0.85 for $s_2$ and 0.78--0.86 for their maximum.

\section{Additional recovery results}
\label{sec:recovery_results}
This section analyzes the complete system with the final corrector (\cref{sec:corrector}). When the failure detector requests an inspection, the corrector verifies SAM~3's current prediction, requests one or two zoomed crops of the query frame, and proposes one action: \textsc{keep}, \textsc{reseed} with a corrected target box (a \emph{re-detection}), or \textsc{erase}. Self-gated commitment applies a corrective action only if the verdict confidence $p$ reaches the commit threshold, $p\ge\tau=0.9$.

We report the system's cost (\cref{sec:cost}), how the gate selects corrections (\cref{sec:gate_calibration}), what each part of the corrector's training contributes (\cref{sec:supp_ablation}), how active zoom contributes (\cref{sec:errors}), how the corrector generalizes (\cref{sec:transfer_lvos}), and qualitative examples (\cref{sec:qual}). All J\&F values come from the official scorers.

\subsection{Efficiency}
\label{sec:cost}
\emph{Our system is faster than Dual-Agent, uses fewer FLOPs and is more accurate.}

\paragraph{Setup.}
We run SAM~3 alone, Dual-Agent and our system on the full SA-V test set, each on one RTX PRO 6000 (Blackwell), and measure the wall-clock time per video, and the nominal FLOPs of all models per video (\cref{tab:cost}). Dual-Agent checks SAM~3 with an 8B reasoning verifier and corrects it with a 4B model; our system uses a single 4B corrector. Our corrector and Dual-Agent's verifier are both served by vLLM in bf16.

\paragraph{Findings.}
Our system is 3.3 times faster than Dual-Agent while being more accurate. The difference lies in where the computation goes. Our corrector reads up to 22 images and two crops per call (\cref{sec:supp_inference}); the frames are encoded in parallel, each crop adds one short turn, and the answer is a short structured output. Dual-Agent's verifier generates a long reasoning chain, and autoregressive decoding processes one token at a time. It also needs 2.6 times less GPU memory (18.4 vs 47.1\,GiB) and fits on a 24\,GB card. The failure detector itself adds only 18\,ms per frame.

\begin{table}[t]
\centering
\small
\caption{\textbf{System cost} on SA-V test, one GPU per system. TFLOP: nominal, per video. GPU GiB: peak memory with all models loaded in PyTorch, without a serving engine. Bold: better of the two self-correcting systems.}
\label{tab:cost}
\begin{tabular}{@{}l rrrr@{}}
\toprule
System & s/video & TFLOP & GPU GiB & J\&F \\
\midrule
SAM~3 alone & 78.0 & 5,410 & 6.2 & 84.4 \\
\midrule
Dual-Agent~\cite{bundele2026trackers} & 410.1 & 8,641 & 47.1 & 81.7 \\
Ours & \textbf{122.7} & \textbf{7,146} & \textbf{18.4} & \textbf{85.6} \\
\bottomrule
\end{tabular}
\end{table}

\subsection{Self-gated commitment}
\label{sec:gate_calibration}
\emph{The corrector's confidence predicts whether a correction will help, so the gate $p\ge0.9$ keeps mainly helpful corrections.}

\paragraph{Setup.}
To relate the confidence $p$ to the effect of a correction, we also need the corrections that the gate would reject. We therefore run the final model \emph{ungated}, applying every correction it proposes, on SA-V test and LVOS v2 val. For each correction we measure the change in J of the target around the corrected frame: on SA-V from the last annotated frame before the correction to the next one, and on LVOS from the frame before to the corrected frame. A correction \emph{helps} if J rises by more than $0.05$ and \emph{hurts} if J falls by more than $0.05$.

\begin{figure}[t]
\centering
\includegraphics[width=\linewidth]{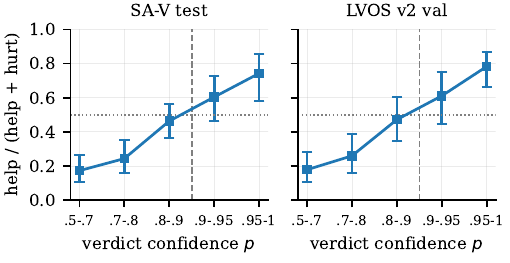}
\caption{\textbf{Fraction of helpful corrections by verdict confidence}: helping corrections among helping and hurting ones in the ungated runs, with Wilson 95\% intervals. Dashed: threshold $p=0.9$.}
\label{fig:gate_help_share}
\end{figure}

\paragraph{Findings.}
The more confident the corrector is, the more likely its correction helps: the share of helping corrections grows steadily with $p$ on both benchmarks, and most corrections with $p>0.95$ help (\cref{fig:gate_help_share}). The AUROC, i.e.\ the probability that a random helping correction has a higher confidence than a random hurting one, is 0.73 on SA-V and 0.77 on LVOS (\cref{tab:calibration}), and the confidence separates the two for re-detections and erasures alike. The gate therefore keeps mainly helpful corrections: among the corrections with $p\ge0.9$, those that help outnumber those that hurt two to one on SA-V and 2.6 to one on LVOS.

\subsection{Training ablations}
\label{sec:supp_ablation}
\label{sec:calibration}
\emph{The verdict loss keeps the confidence calibrated, which makes the gate effective; identity contrasts teach the corrector to recognize the target by its appearance.}

\paragraph{Setup.}
The final corrector is trained in three stages (\cref{sec:supp_training}): stage~1 teaches the verdict and the actions, stage~2 the crop requests, and stage~3 refines the whole dialogue with a reward on the final answer. We train three variants, each with one component removed:
\begin{itemize}\itemsep0pt
 \item \emph{Without verdict loss.} The stage-3 reward scores only the visibility and box, so the final model is also trained with a cross-entropy loss on its verdict (the \emph{verdict loss}), which ties its confidence to how often the verdict is correct. This variant drops that loss ($\lambda_{\mathrm v}=0$ in \cref{eq:stage3_loss}).
 \item \emph{Without zoom.} Stages~2 and~3 use the same examples with all crop turns removed; the corrector answers from the full frames only.
 \item \emph{Without identity contrasts.} Stage~1 is trained without the identity contrasts (\cref{sec:identity_contrasts}), which are replaced by other examples at equal corpus size; this variant is compared with the stage-1 model.
\end{itemize}

We evaluate every model in closed loop with gate (\cref{tab:supp_ablation,tab:stage1_ablation}). We also measure how well its confidence is \emph{calibrated}, i.e.\ how closely $p$ matches the fraction of correct verdicts (\cref{tab:calibration}). A verdict is correct if it says \texttt{wrong} exactly when SAM~3 is failing; ECE is the average gap between confidence and accuracy over ten confidence bins. We also report how well confidence separates helping from hurting corrections (Help, the AUROC of \cref{sec:gate_calibration}), which is what the gate relies on. We measure both on the ungated runs, so that the gate does not filter the answers.

\begin{table}[t]
\centering
\small
\setlength{\tabcolsep}{6pt}
\caption{\textbf{Ablations of the final model} (J\&F, gated at $p\ge0.9$ unless stated).}
\label{tab:supp_ablation}
\begin{tabular}{@{}l cc@{}}
\toprule
Corrector & SA-V test & LVOS v2 val \\
\midrule
Final model             & \textbf{85.6} & \textbf{89.8} \\
\quad without the gate  & 81.3 & 87.7 \\
Without verdict loss    & 84.8 & 89.2 \\
Without zoom            & 84.8 & 89.7 \\
\midrule
SAM~3 alone & 84.4 & 88.2 \\
\bottomrule
\end{tabular}
\end{table}

\begin{table}[t]
\centering
\small
\setlength{\tabcolsep}{6pt}
\caption{\textbf{Identity contrasts} in stage~1 (J\&F, gated at $p\ge0.9$).}
\label{tab:stage1_ablation}
\begin{tabular}{@{}l cc@{}}
\toprule
Stage-1 corrector & SA-V test & LVOS v2 val \\
\midrule
With identity contrasts    & \textbf{85.1} & \textbf{89.3} \\
Without identity contrasts & 84.9 & 88.4 \\
\bottomrule
\end{tabular}
\end{table}

\begin{table}[t]
\centering
\small
\setlength{\tabcolsep}{3pt}
\caption{\textbf{Confidence quality} on the ungated runs. ECE: calibration error of the verdict confidence; Help: AUROC of the confidence for helping vs.\ hurting corrections.}
\label{tab:calibration}
\resizebox{0.8\linewidth}{!}{%
\begin{tabular}{@{}l cc cc@{}}
\toprule
& \multicolumn{2}{c}{SA-V test} & \multicolumn{2}{c}{LVOS v2 val} \\
\cmidrule(lr){2-3}\cmidrule(l){4-5}
Corrector & ECE$\downarrow$ & Help$\uparrow$ & ECE$\downarrow$ & Help$\uparrow$ \\
\midrule
Final model & 0.053 & \textbf{0.727} & 0.107 & \textbf{0.774} \\
Without verdict loss & 0.243 & 0.621 & 0.409 & 0.481 \\
Without zoom & 0.061 & 0.666 & 0.090 & 0.677 \\
\bottomrule
\end{tabular}}
\end{table}

\paragraph{The gate needs a calibrated confidence.}
The gate applies a correction only when $p$ is high, which is effective only if a high $p$ indicates a likely correct verdict. The final model uses its confidence this way: it also proposes corrections in doubtful cases but assigns them a low $p$, leaving the decision to the gate. Without the gate, 452 answers on SA-V become corrections. The model assigns $p\ge0.9$ to 129 of them. With the gate, the closed loop applies 80 corrections, fewer than 129 because later calls change once fewer corrections are applied. The gate raises J\&F by 4.3 on SA-V and 2.1 on LVOS compared with applying every correction (\cref{tab:supp_ablation}). With the gate, the final model outperforms SAM~3 alone on both benchmarks.

\paragraph{The verdict loss keeps the confidence calibrated.}
When stage~3 uses the reward alone, the corrector becomes overconfident: 82\% of its answers on SA-V have $p\ge0.9$, against 26\% for the final model, and its ECE is 3.8--4.6 times higher (\cref{tab:calibration}). Its confidence also loses the information the gate relies on: it separates helping from hurting corrections less well, and on LVOS no better than chance. The verdict loss preserves both properties, so the gate can select the helpful corrections, and the final model outperforms the variant without the verdict loss on both benchmarks (\cref{tab:supp_ablation}).

\begin{figure*}[t]
\centering
\includegraphics[width=0.8\textwidth]{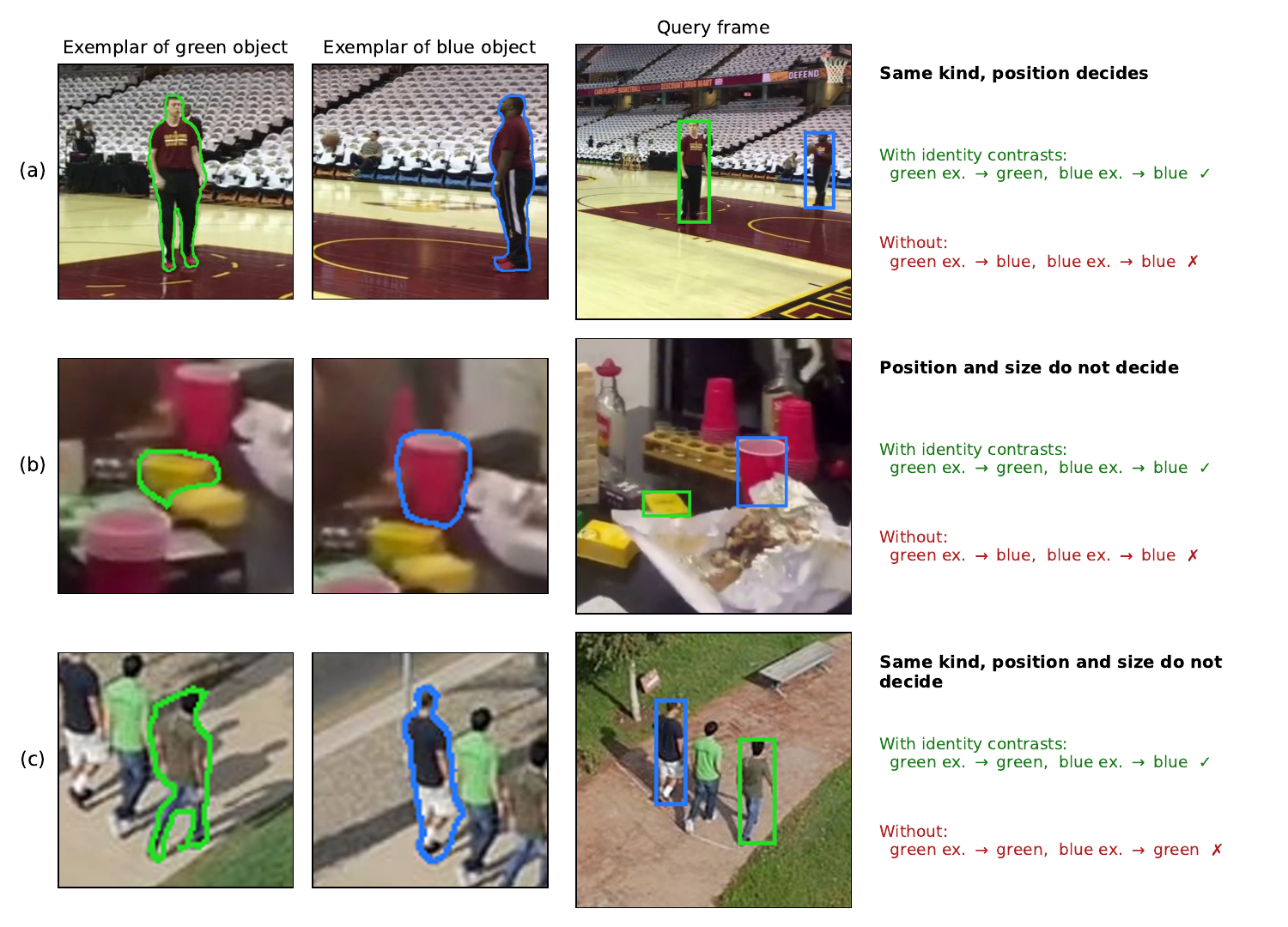}
\caption{\textbf{Identity tests.} The corrector sees one object's exemplar and the bare query frame; it passes if its choice follows the exemplar. Without identity contrasts, it picks the same object for both exemplars.}
\label{fig:identity_test}
\end{figure*}

\paragraph{Identity contrasts teach recognition by appearance.}
In an identity switch, SAM~3 moves from the target to a similar-looking object. The frames alone cannot reveal the error; only the \emph{exemplar}, the annotated view of the target given at initialization, shows which of the two objects is the target. Identity contrasts are stage-1 training examples that force the corrector to use the exemplar: the same query frame appears twice, once with the target's exemplar and once with the look-alike's, and the two copies have opposite answers (\cref{fig:train_examples}c,d). In closed loop, identity contrasts improve J\&F on both benchmarks, most on LVOS (\cref{tab:stage1_ablation}).

Since J\&F mixes identity errors with all other failures, we also test this ability directly. The test uses annotated frames from LVOS v2 val, SA-V val or SA-V test (125 videos) in which the target and a second annotated object are both visible. The corrector receives the exemplar of one of the two objects and, as at deployment, the earlier frames, the query frame and the later frames; no mask is drawn on any frame, so the tracker's output cannot reveal the answer. The frames are the same for both exemplars, so only the exemplar can tell the corrector which object is the target. 

We compare the log-probabilities that the corrector assigns to the two annotated boxes and take the higher one as its choice. A test is passed if the corrector chooses the first object when given its exemplar and the second object when given the other exemplar, i.e.\ if its choice follows the exemplar (\cref{fig:identity_test}).
Identity contrasts raise the pass rate from 68.1\% to 74.6\% overall (\cref{tab:identity_test}). The gain is largest, from 46.8\% to 58.4\%, on the hardest tests: the two objects are of the same kind (same object category), and choosing the object closer to the exemplar in position or size would give the wrong answer, so only appearance can decide. Without identity contrasts, the corrector often chooses the same object for both exemplars.

\begin{table}[t]
\centering
\small
\caption{\textbf{Identity test}: tests passed (\%) by the stage-1 corrector with and without identity contrasts.}
\label{tab:identity_test}
\begin{tabular}{@{}l r cc@{}}
\toprule
Tests & N & with & without \\
\midrule
All & 894 & \textbf{74.6} & 68.1 \\
Same kind & 486 & \textbf{63.1} & 55.7 \\
Position and size do not decide & 245 & \textbf{69.0} & 58.5 \\
\quad and same kind & 158 & \textbf{58.4} & 46.8 \\
\bottomrule
\end{tabular}
\end{table}

\subsection{Active zoom}
\label{sec:errors}
\label{sec:act_situation}
\emph{Zoom gives the corrector the resolution that small objects require: it places its crops on the target, the corrections it makes from them are reliable, and the gate becomes more selective.}

\paragraph{Setup.}
The corrector receives every frame downscaled to a fixed pixel budget (\cref{sec:supp_inference}). For an object that covers a small part of the frame, this leaves too few pixels to verify its identity or to localize it. Zoom lets the corrector request crops of the query frame at source resolution, enlarged up to $6\times$. We assess zoom by ablation (\cref{sec:supp_ablation}) and by analyzing every correction that the gated final model applies, labeled by its action and by whether the requested crop covers at least half of the annotated target.

\begin{figure*}[t]
\centering
\includegraphics[width=\textwidth]{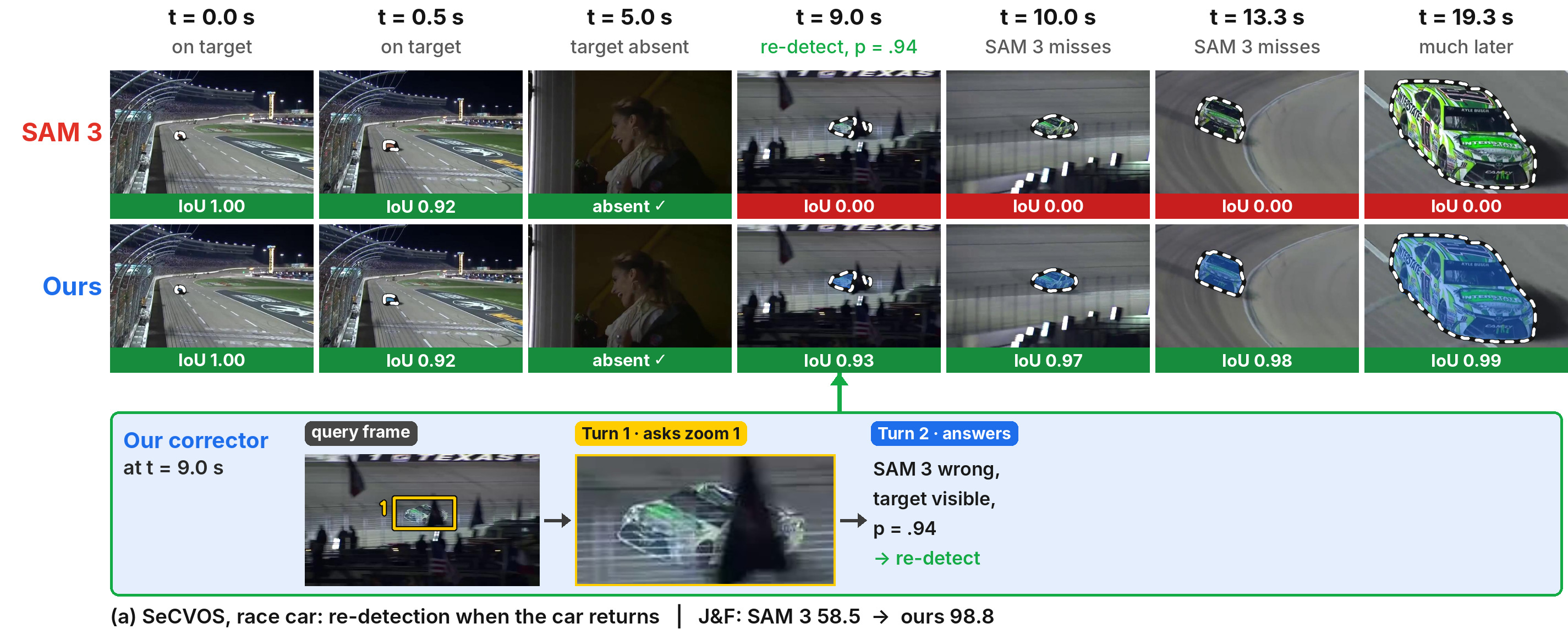}\\[6pt]
\includegraphics[width=\textwidth]{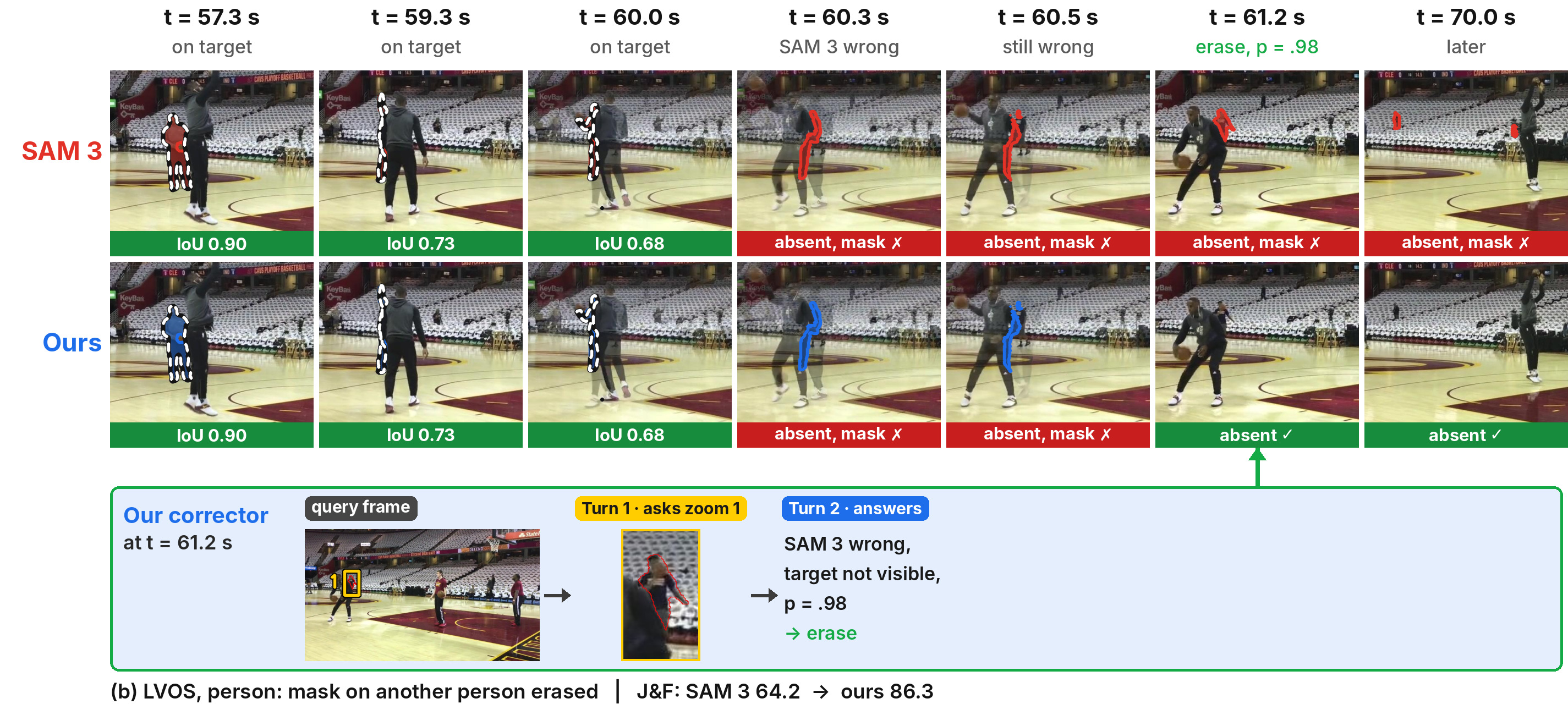}
\caption{\textbf{SAM~3 alone (top row, red) vs.\ our system (bottom row, blue), with our corrector's zoom.} Dashed white line: annotated target. Bar under each tile: mask IoU (green $\ge0.5$, red $<0.5$), or, for a target annotated absent, whether a mask is shown. Green column header and arrow: the frame at which our corrector's act is applied, with the action and the verdict confidence $p$ (re-detect and new box: \textsc{reseed}; erase: \textsc{erase}). Blue strip: our corrector's turns at the act: the zoom regions it asked for (yellow, numbered), the crops it received, and its answer. Below each example: dataset, object and per-object J\&F. (a)~Re-detection of a small car after a scene cut. (b)~Erasure of a small mask on another person after the target has left the view.}
\label{fig:qual}
\end{figure*}

\begin{figure*}[t]
\centering
\includegraphics[width=\textwidth]{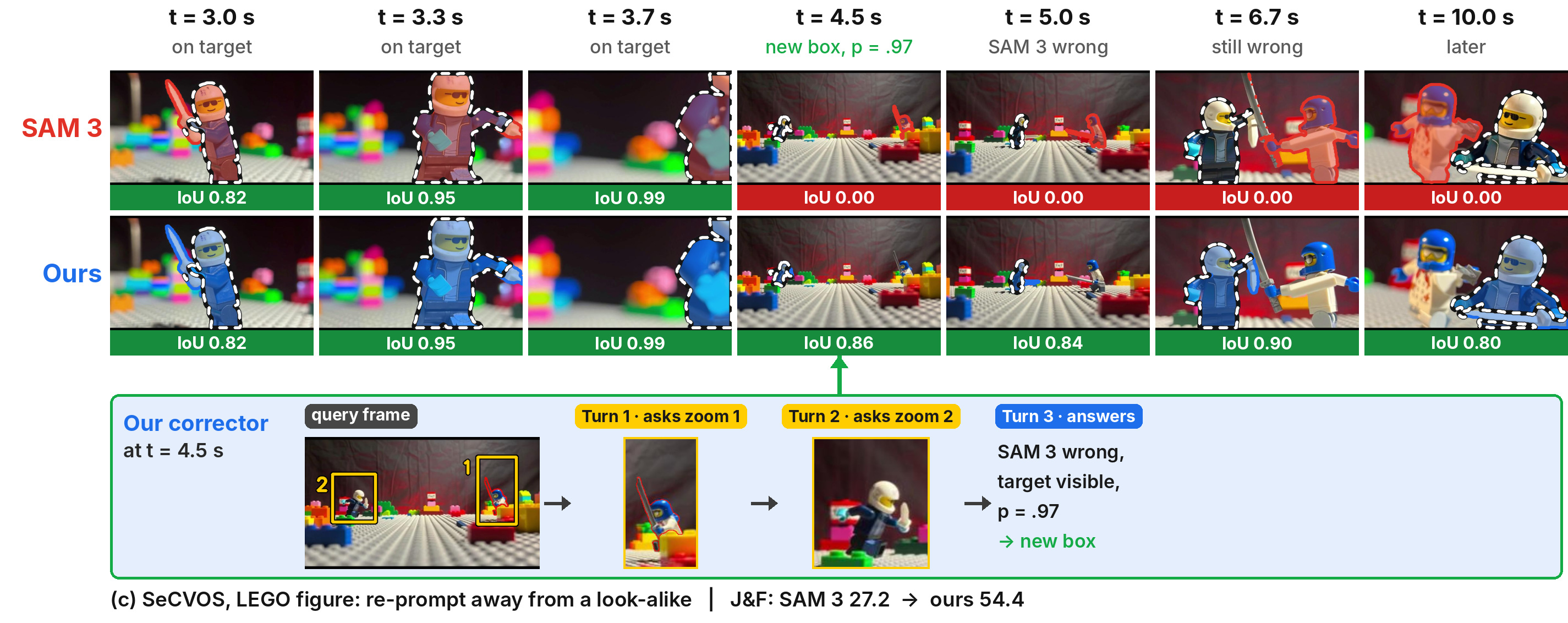}\\[6pt]
\includegraphics[width=\textwidth]{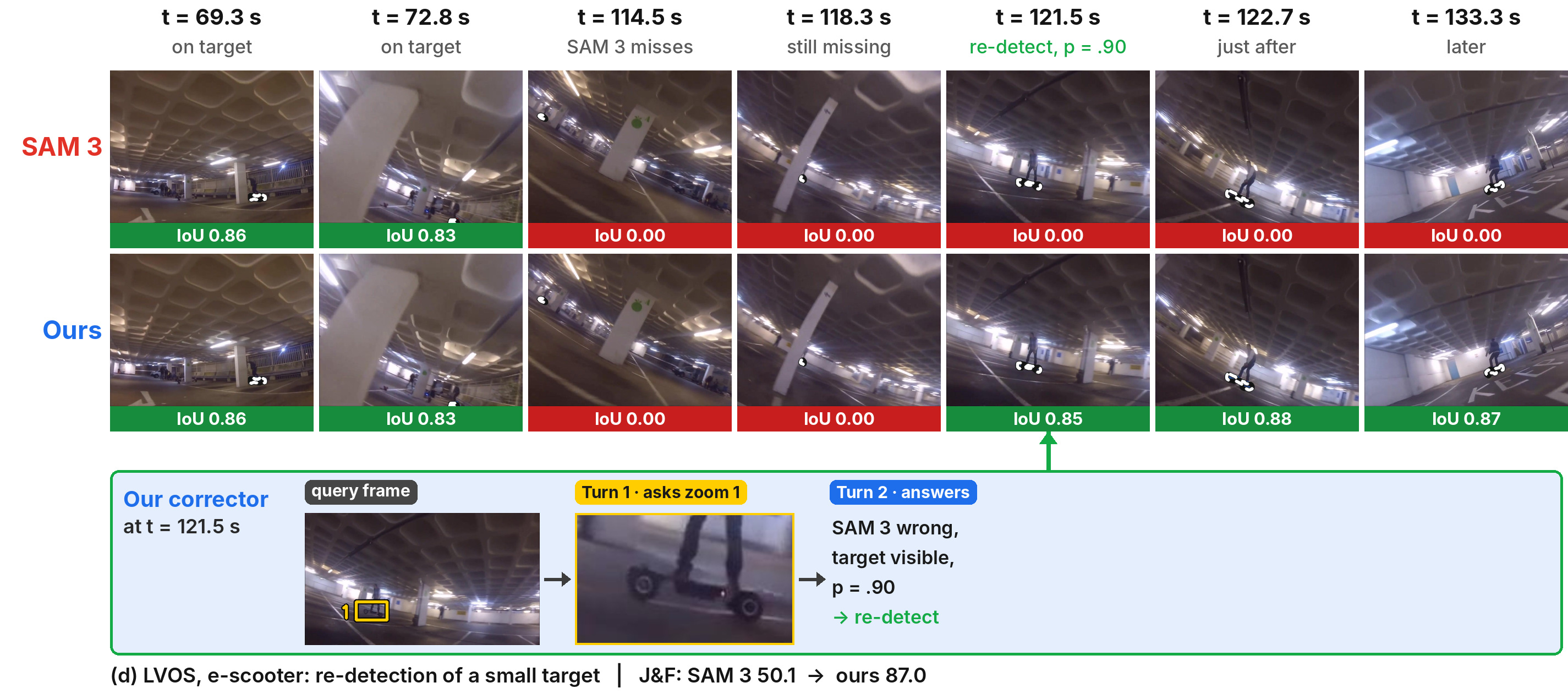}
\caption{\textbf{More examples, as in \cref{fig:qual}.} (c)~After a switch to a look-alike, two crops separate the target from the look-alike, and the corrector re-seeds on the target. (d)~Re-detection of a target covering 0.2\% of the frame.}
\label{fig:qual2}
\end{figure*}

\paragraph{Crops on the target yield reliable corrections.}
A crop helps only if it shows the object to be judged. On LVOS, the crop covers at least half of the annotated target in 79\% of the calls in which the target is present. The corrections made from these crops are reliable. When the crop shows the target, the corrector re-detects it, and 26 of 30 re-detections help on LVOS and 13 of 15 on SA-V. When the target has left the view, the corrector erases SAM~3's mask, and these erasures help or leave J unchanged in all 27 cases on LVOS and in 15 of 18 on SA-V. \Cref{fig:qual,fig:qual2} show examples: in each, the crop enlarges an object that is too small to judge in the full frame, either the target or the wrong object that SAM~3 follows.

\paragraph{The gains carry over to the full system.}
In closed loop, zoom adds 0.8 J\&F on SA-V (\cref{tab:supp_ablation}). Zoom also makes the confidence more useful to the gate. Both models are similarly well calibrated, but with zoom the confidence separates helping from hurting corrections better, with AUROC 0.73 instead of 0.67 on SA-V and 0.77 instead of 0.68 on LVOS (\cref{tab:calibration}). The gate therefore admits a more reliable set of corrections.

\subsection{Generalization}
\label{sec:transfer_lvos}

\begin{table}[t]
\centering
\small
\setlength{\tabcolsep}{5pt}
\caption{\textbf{Closed-loop recovery on other trackers} (J\&F).}
\label{tab:transfer_lvos}
\begin{tabular}{@{}l cc@{}}
\toprule
Tracker & SeCVOS & LVOS v2 val \\
\midrule
SAM~2.1 & 58.2 & 84.1 \\
\quad + ours & \textbf{73.6} & \textbf{86.3} \\
\midrule
SeC & 70.0 & 86.5 \\
\quad + ours & \textbf{76.5} & \textbf{87.4} \\
\bottomrule
\end{tabular}
\end{table}

\emph{Without retraining or refitting, our detector and corrector improve SAM~2.1 and SeC on both benchmarks.}

\paragraph{Setup.}
The corrector is trained exclusively on SAM~3 trajectories, and the detector's constants are fitted on SAM~3. We apply the detector, the corrector and self-gated commitment unchanged to SAM~2.1 (Hiera-L) and to SeC, which pairs SAM~2.1's prediction heads with a retrained memory.

\paragraph{Findings.}
Both trackers improve on SeCVOS and LVOS v2 val (\cref{tab:transfer_lvos}). The largest gain is for SAM~2.1 on SeCVOS ($+15.4$ J\&F), so that SAM~2.1 with our system surpasses SeC alone, despite an 11.8-point gap between the two trackers. SeC gains 6.5 J\&F on SeCVOS, although its memory already incorporates vision-language reasoning.

\paragraph{Why the system transfers.}
Both components interface with the tracker only through quantities that every SAM~2-family tracker exposes. The detector reads the objectness logit, the predicted mask IoU and the memory features (\cref{sec:supp_detector,sec:tracker_transfer}). The corrector receives the predicted mask, rendered as an outline on the frames, and acts through a box prompt or an empty-mask prompt. Neither component uses SAM~3-specific features, so the tracker can be replaced without retraining.

\begin{figure*}[t]
\centering
\includegraphics[width=\textwidth]{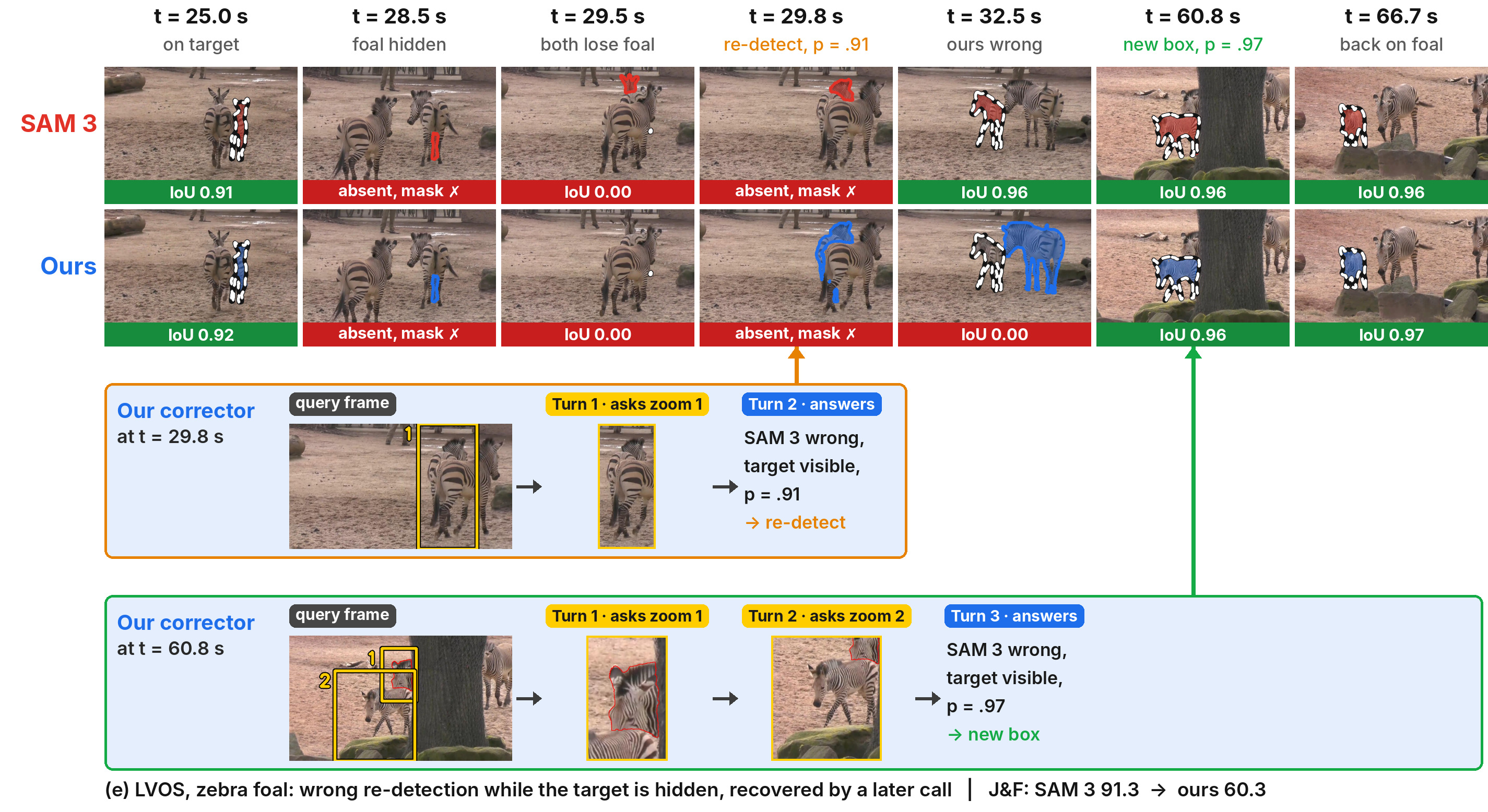}
\caption{\textbf{A hard case, as in \cref{fig:qual}.} (e)~While the target (a zebra foal) is occluded, the crop shows only the adults and the corrector re-seeds on an adult zebra (orange). A later call, with two crops, places a new box on the foal (green), and the track follows it again.}
\label{fig:qual3}
\end{figure*}

\subsection{Qualitative examples}
\label{sec:qual}

\emph{The examples illustrate the mechanism of \cref{sec:errors}: in each success, one or two crops resolve a target or a wrong object that is too small to judge in the downscaled frame.}

\Cref{fig:qual,fig:qual2,fig:qual3} compare SAM~3 alone with our system on five objects from LVOS and SeCVOS; the caption of \cref{fig:qual} describes the layout. Examples (a)--(d) cover the main correction types: re-detection after a scene cut, erasure of a spurious mask, recovery from a switch to a look-alike, and re-detection of a very small target. Example (e) shows a case of occlusion by a look-alike.
\begin{itemize}
 \item \emph{(a) Re-detection after a cut.} The car reappears after a scene cut, covering 1.5\% of the frame, and SAM~3 does not re-acquire it (IoU 0.00). The corrector crops the car and issues \textsc{reseed}; the track recovers (IoU 0.93, then 0.98). J\&F: 58.5 $\to$ 98.8.
 \item \emph{(b) Erasure of a spurious mask.} After the target leaves the view, SAM~3 places a small mask on a spectator. The crop shows a different person, and the corrector issues \textsc{erase}. J\&F: 64.2 $\to$ 86.3.
 \item \emph{(c) Recovery from a look-alike switch.} After a cut, SAM~3 follows the other LEGO figure (IoU 0.00). The corrector crops both figures and issues \textsc{reseed} with a box on the target, which the track then follows (IoU 0.80--0.90). J\&F: 27.2 $\to$ 54.4.
 \item \emph{(d) Re-detection of a very small target.} The e-scooter covers 0.2\% of the frame and is lost by both systems. The crop resolves it, and the corrector issues \textsc{reseed} (IoU 0.85, then 0.88). J\&F: 50.1 $\to$ 87.0.
 \item \emph{(e) Occlusion by a look-alike.} The foal is occluded by an adult zebra and annotated absent; the crop shows only the adults, and the corrector re-seeds on an adult (orange frame). After the foal reappears, a later corrector call returns the track to it, 31\,s later. J\&F: 91.3 $\to$ 60.3.
 
\end{itemize}

\raggedbottom %
\FloatBarrier %

\clearpage
\appendix

\end{document}